\documentclass[11pt, a4paper, logo, copyright, nonumbering]{astribot}
\usepackage[authoryear, sort&compress, round]{natbib}
\usepackage{dblfloatfix}
\usepackage{ulem}
\usepackage{caption}
\usepackage{dramatist}
\usepackage{xspace}
\usepackage{pifont} 
\usepackage{multirow}
\usepackage{tcolorbox}
\usepackage{xltabular}
\usepackage{longtable}
\usepackage{hyperref}
\usepackage{amsfonts}
\usepackage{amsmath}
\usepackage{amssymb}
\usepackage{lineno}
\usepackage{multirow}
\usepackage{adjustbox}
\usepackage{siunitx}
\usepackage{mathtools}
\usepackage{multirow}
\usepackage[bottom]{footmisc}
\usepackage{fontawesome5}
\usepackage{subcaption}
\usepackage{setspace}
\usepackage{lipsum} 
\usepackage{multicol}
\usepackage{annotate-equations}
\usepackage{pdfpages}
\usepackage{threeparttable}
\usepackage{makecell}
\usepackage{xcolor}

\tcbset{textmarker/.style={
        halign=center,parbox=false,boxrule=0mm,boxsep=0mm,arc=0mm,
        outer arc=0mm,left=1mm,right=1mm,top=7pt,bottom=7pt,
        toptitle=1mm,bottomtitle=1mm,oversize,}}
\newtcolorbox{hintBox}{textmarker,
    colback=yellow!10!white}
\newtcolorbox{importantBox}{textmarker,
    colback=red!10!white}
\newtcolorbox{noteBox}{textmarker,
    colback=green!10!white}
\newcommand{\note}[1]{\begin{noteBox} \textbf{} #1 \end{noteBox}}
\newcommand{\warning}[1]{\begin{hintBox} \textbf{} #1 \end{hintBox}}
\newcommand{\important}[1]{\begin{importantBox} \textbf{} #1 \end{importantBox}}

\definecolor{shadeblue}{RGB}{74, 111, 255}
\definecolor{shadered}{RGB}{255, 107, 107}
\definecolor{shadegreen}{RGB}{76, 175, 80}

\makeatletter
\def\@BTrule[#1]{%
  \ifx\longtable\undefined
    \let\@BTswitch\@BTnormal
  \else\ifx\hline\LT@hline
    \nobreak
    \let\@BTswitch\@BLTrule
  \else
     \let\@BTswitch\@BTnormal
  \fi\fi
  \global\@thisrulewidth=#1\relax
  \ifnum\@thisruleclass=\tw@\vskip\@aboverulesep\else
  \ifnum\@lastruleclass=\z@\vskip\@aboverulesep\else
  \ifnum\@lastruleclass=\@ne\vskip\doublerulesep\fi\fi\fi
  \@BTswitch}
\makeatother

\addto\extrasenglish{
}

 {\begin{list}{}%
         {\setlength{\leftmargin}{#1}}%
         \item[]%
 }
 {\end{list}}
 
\renewcommand{\today}{}
\newcommand{\model}[0]{\mbox{Lumo-1}\xspace}

\title{Mind to Hand: Purposeful Robotic Control \\via Embodied Reasoning}
\author[*]{
Astribot Team
\\
\small
\texttt{research@astribot.com}
\\
\vspace{2em}
\small
Project Page: 
\url{www.astribot.com/research/Lumo1}
\\
\vspace{1em}
\small
Author List in \hyperref[sec:contribution]{Contributions}
}

\begin{abstract}
Humans act with context and intention, with reasoning playing a central role. While internet-scale data has enabled broad reasoning capabilities in AI systems, grounding these abilities in physical action remains a major challenge. We introduce \model, a generalist vision-language-action (VLA) model that unifies robot reasoning (``mind'') with robot action (``hand''). 
Our approach builds upon the general multi-modal reasoning capabilities of pre-trained vision-language models (VLMs), progressively extending them to embodied reasoning and action prediction, and ultimately towards structured reasoning and reasoning–action alignment. This results in a three-stage pre-training pipeline: (1) \textbf{Continued VLM pre-training} on curated vision-language data to enhance embodied reasoning skills such as planning, spatial understanding, and trajectory prediction; (2) \textbf{Co-training} on cross-embodiment robot data alongside vision-language data; and (3) \textbf{Action training with reasoning process} on trajectories collected on Astribot S1, a bimanual mobile manipulator with human-like dexterity and agility. Finally, we integrate \textbf{reinforcement learning} to further refine reasoning–action consistency and close the loop between semantic inference and motor control.
Extensive experiments demonstrate that \model achieves significant performance improvements in embodied vision-language reasoning, a critical component for generalist robotic control. Real-world evaluations further show that \model surpasses strong baselines ($\pi$0, $\pi$0.5) across a wide range of challenging robotic tasks, with strong generalization to novel objects and environments, excelling particularly in long-horizon tasks and responding to human-natural instructions that require reasoning over strategy, concepts and space.
\end{abstract}

\begin{document}
\maketitle

\section{Introduction}
The long-standing vision of robotics is to build intelligent agents capable of operating in human environments - perceiving the world as people do, reasoning about the course and consequences of their actions, and ultimately co-existing with humans to support daily life. Despite decades of progress, this vision remains largely unrealized~\citep{gupta2021embodied,team2021creating}. A key challenge lies in the immense diversity of the real world, which requires robot policies to generalize across a wide spectrum of tasks and environments. Human interaction further compounds this difficulty: intent is often expressed through natural, flexible instructions such as, ``I’m thirsty, please bring me something low in calories on the kitchen table.''  Executing such commands demands more than action planning based on the current state - it requires reasoning over abstract concepts, spatial relations, and contextual cues.

Recent advances in Vision-Language-Action (VLA) models have opened promising directions for building intelligent generalist robot policies. These models typically build upon pre-trained Vision-Language Models (VLMs) and extend them with action prediction capabilities~\citep{zitkovich2023rt, black2024pi_0, team2025gemini, bjorck2025gr00t, intelligence2025pi_, cheang2025gr3}. This integration allows robots to interpret natural language instructions and perform various tasks. However, current VLAs remain limited in generalization, robustness, and interpretability, particularly compared to their vision and language foundations. In practice, they provide little transparency into why one action is chosen over another. These limitations arise not only from data scarcity but also from insufficient reasoning - a fundamental requirement for purposeful action. In contrast, humans implicitly evaluate context and intention before acting, transforming perception into coherent, adaptive behavior. For robots to achieve comparable rationality, actions must go beyond direct mappings from observations to control signals, instead emerging as the product of structured reasoning.

In this report, we introduce \model, a Vision-Language-Action (VLA) model for end-to-end robotic control. Given natural language instructions, robot onboard sensor inputs, and the robot state, \model generates actions to control a whole-body bimanual robot. \model builds upon the pre-trained vision-language model Qwen2.5-VL-7B~\citep{bai2025qwen2}. To enable robot action prediction capabilities, we first train \model with next-token prediction objective over discrete actions. This strategy preserves general language understanding, accelerates and stabilizes policy learning, and enables natural co-training with large-scale vision–language data. To achieve a compact discrete representation of actions, we introduce a \textbf{spatial action tokenizer} that provides a controllable compression rate according to action resolution requirements, with a more compact representation than both the FAST tokenizer~\citep{pertsch2025fast} and binning-based discretization. 
Using this representation, \model generates variable-length action tokens that decode into short-horizon robot trajectories (up to 1.33 seconds), with shorter prediction horizons for more complex and dexterous motions.
For fine-tuning, we add an action expert trained with flow matching~\citep{lipman2022flow} to improve inference efficiency. To further accelerate the fine-tuning process and to enhance the action expert's generalizability across different tasks, we introduce a pre-training stage for the action expert to learn the unconditional distribution of actions. During fine-tuning, this distribution is transformed into conditional action generation, resulting in an efficient and effective training pipeline.

We conduct in-depth studies on architecture design,  action tokenization, and compute scaling, identifying key design choices that are critical for reasoning and instruction following. To advance embodied and action-centric reasoning, we propose a systematic three-stage training pipeline: (1) \textbf{Continued VLM pre-training} on curated vision-language data to strengthen embodied reasoning; (2) \textbf{Co-training} on cross-embodiment robot data alongside vision-language data to enable action prediction capability while preserving general knowledge; (3) \textbf{Action training with reasoning process} to promote structured reasoning toward purposeful and successful action execution.
Finally, we leverage \textbf{Reinforcement Learning (RL)} to refine embodied reasoning and strengthen the alignment between high-level reasoning and low-level control.
This training pipeline enables \model to generalize beyond robot data, handling novel objects, environments, and concepts such as size, spatial relations, and commonsense knowledge. Its reasoning traces provide reliable cues and transparent insights into decision-making. 
We validate \model in extensive real-world experiments on three types of challenging tasks: (1) generalizable pick-and-place, (2) long-horizon tasks, and (3) dexterous manipulation. Across all task categories, \model consistently outperforms state-of-the-art baseline $\pi_0$ and $\pi_{0.5}$, demonstrating strong generalization to novel objects, environments, and complex semantics requiring reasoning over strategy, concepts and space.

\section{The \model Model}

\subsection{Preliminaries}
\paragraph{Vision-Language Models.}
To equip action models with visual and linguistic world knowledge acquired from web-scale data, we leverage vision–language models (VLMs), which typically comprise three key components: (1) a visual encoder that maps images into patch-level embeddings, (2) a large language model (LLM) backbone, and (3) a projection module that aligns visual features with the language model's input space. VLMs are generally trained via next-token prediction on paired or interleaved image-text data. In this work, we build upon Qwen2.5-VL-7B~\citep{bai2025qwen2}, which adheres to this canonical architecture.

\paragraph{Vision-Language-Action Models.}
Vision-language-action (VLA) models, denoted as $\pi_\theta$, are typically optimized via imitation learning on large-scale robot demonstration datasets~$\mathcal{D}$. For each timestep $t$ with observation~$\mathbf{o}_t$ and natural-language instruction~$\ell$, the training objective is to maximize the probability of generating the ground truth action~$\mathbf{a}_t$, or more generally, an action chunk~$\mathbf{a}_{t:t+H}$ over a horizon of $H$ timesteps:  
\begin{equation}
    \max_\theta \, \mathbb{E}_{(\mathbf{a}_{t:t+H}, \mathbf{o}_t, \ell) \sim \mathcal{D}} 
    \; \log \big( \pi_\theta(\mathbf{a}_{t:t+H} \mid \mathbf{o}_t, \ell) \big).
\end{equation}

The observation ~$\mathbf{o}_t$ typically comprises multi-view visual inputs and the proprioceptive state of the robot.  
Architecturally, VLA models extend the design principles of large-scale language and vision–language modelling. They employ dedicated tokenizers for each modality to transform inputs and outputs into either discrete or continuous token sequences, which are then processed by a unified autoregressive transformer backbone which parameters are commonly initialized from a pre-trained vision–language foundation model. With both policy inputs and outputs represented in tokenized form, imitation learning can be formulated as a next-token prediction problem over the concatenated sequence of observation, instruction, and action tokens.

\paragraph{Reasoning and Reinforcement Learning for VLA Models}
Step-by-step reasoning prior to producing an output - commonly referred to as chain-of-thought (CoT) reasoning - has become a key paradigm for enhancing large language model (LLM) performance. Extending this idea to vision-language-action (VLA) models, we enrich the demonstration dataset $\mathcal{D}$ with reasoning traces, allowing the model $\pi_\theta$ to jointly optimize over both reasoning $\mu$ and action chunk $\mathbf{a}_{t:t+H}$:
\begin{equation}
    \max_\theta \, \mathbb{E}_{(\mathbf{a}_{t:t+H}, \mathbf{o}_t, \ell) \sim \mathcal{D}} 
    \; \log \big( \pi_\theta(\eqnmarkbox[blue]{Psi2}{\mu}, \mathbf{a}_{t:t+H} \mid \mathbf{o}_t, \ell) \big).
\label{eq:optimization_objective}
\end{equation}

To further encourage correct reasoning and its alignment with action generation, we employ a reinforcement learning stage using Group Relative Policy Optimization (GRPO)~\citep{shao2024deepseekmath}. GRPO operates over a group of $G$ sampled results $\{\mathbf{z}_1, \mathbf{z}_2 \ldots, \mathbf{z}_G \}$ from the current policy $\pi_\theta$, where each response $\mathbf{z}_i$ comprises both the reasoning sequence $\mu$ and the predicted action $\mathbf{a}_{t:t+H}$. Each response is assigned a reward $r_i$ reflecting its overall quality, and the optimization objective is defined as: 
\begin{equation}
    \mathcal{J}_\text{GRPO}(\theta) = \frac{1}{G} \sum_{i=1}^{G}(\frac{\pi_{\theta}(\mathbf{z}_i|\mathbf{o}_t,\ell)}{\pi_{\theta_\text{old}}(\mathbf{z}_i|\mathbf{o}_t,\ell)}A_i - \beta D_{KL}(\pi_{\theta}(\mathbf{z}_i|\mathbf{o}_t,\ell)\parallel \pi_{\theta_\text{old}}(\mathbf{z}_i|\mathbf{o}_t,\ell))),
\end{equation}
\begin{equation*}
     \\ \text{where} \quad A_i=\frac{r_i - \text{mean}(\{ r_1, \ldots, r_G \})}{\text{std}(\{ r_1, \ldots, r_G \})}.
\end{equation*}
Here, $A_i$ measures the relative advantage of the $i$-th response within the sampled group. The KL regularization term, weighted by $\beta$, constrains policy updates to remain close to the previous model $\pi_{\theta_\text{old}}$, thereby ensuring stable and conservative policy improvement.

\begin{figure}[t]
  \centering
  \includegraphics[width=\linewidth]{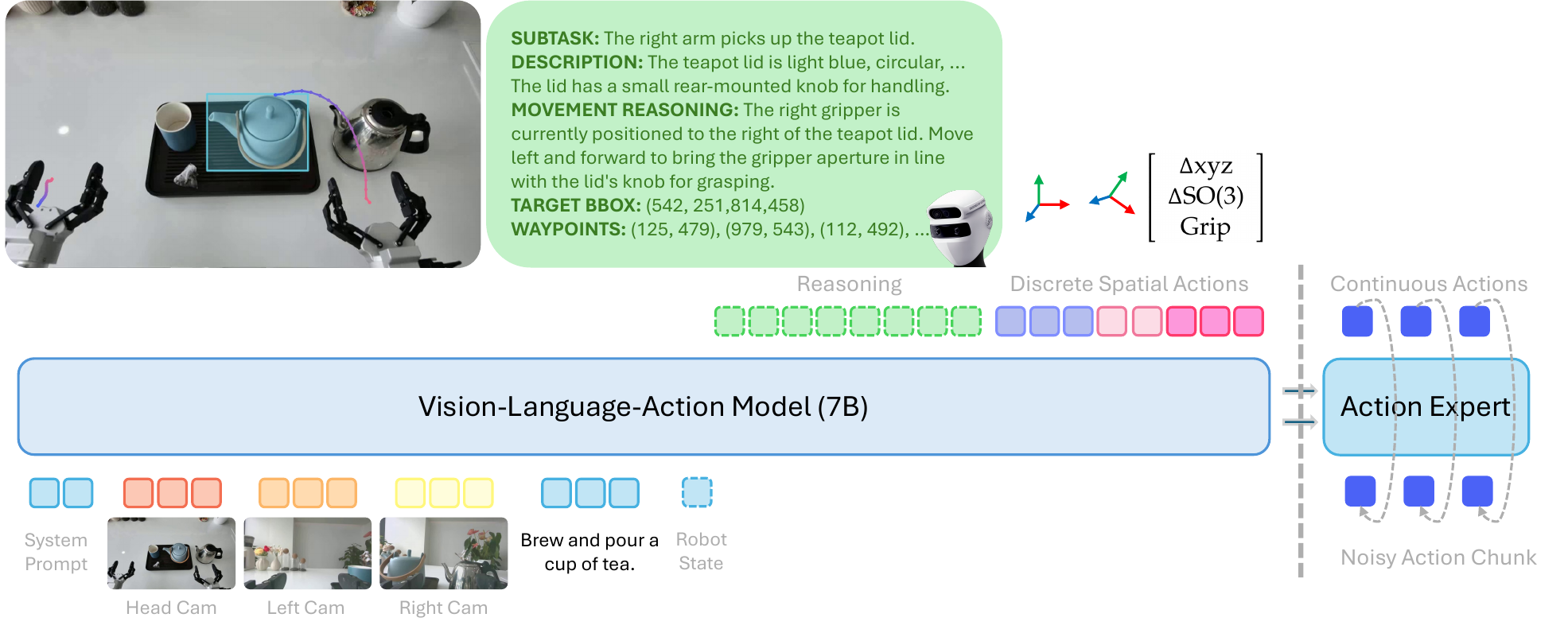}
   \caption{\textbf{Model Architecture Illustration.} \model supports next-token prediction for both vision-language and action data, as well as flow-matching  for modeling continuous actions. }
   \label{fig:model_architecture}
\end{figure}

\subsection{Model Architecture}
\paragraph{Model Overview.}  
\model is an end-to-end Vision-Language-Action (VLA) model designed to jointly model distributions over both action chunks and textual outputs, as formally defined in Eq.~\ref{eq:optimization_objective}. The text modality serves two purposes: (1) supporting pre-training and co-training on vision-language tasks, such as visual question answering, and (2) enabling structured reasoning about actions, such as subtask description, object localization and robot trajectory prediction. 
High-level textual inference is captured by $\pi_\theta(\mu \mid \mathbf{o}_t, \ell)$. For low-level action generation, it can be decomposed into the joint distribution as follows:
\begin{equation}
\pi_\theta(\mu, \mathbf{a}_{t:t+H} \mid \mathbf{o}_t, \ell) 
= \pi_\theta (\mathbf{a}_{t:t+H} \mid \mathbf{o}_t, \mu) \,
\pi_\theta(\mu \mid \mathbf{o}_t, \ell),
\end{equation}
where low-level action inference depends only on $\mu$. Both high-level and low-level distributions are parameterized within a single, unified model.  

The underlying architecture of \model is a multi-modal transformer, as illustrated in Fig.~\ref{fig:model_architecture}. Each input token corresponds to either a text token or an image patch token, which are processed by modality-specific encoders. The model outputs are composed of text tokens and discrete action tokens, the latter encoded using our proposed spatial action tokenizer as introduced in Sec.~\ref{subsec:spatail_tokenizer}. During fine-tuning, a pre-trained flow-matching action expert is integrated to generate actions more efficiently, conditioned on the key–value (KV) cache produced by the VLA backbone.

\subsection{Spatial Action Tokenization Algorithm}
\label{subsec:spatail_tokenizer}
In practice, the tokenization strategies for image and text generally follow established designs in modern vision–language models. In contrast, action tokenization remains relatively underexplored.
The most widely used approach relies on simple binning-based discretization schemes~\citep{brohan2022rt,zitkovich2023rt,kim2024openvla}, where each action dimension is quantized independently. Specifically, for a given action $\mathbf{a}_t$, the value range of each dimension is divided into $N$ uniform bins, most commonly with $N=256$. For a $D$-dimensional action chunk $\mathbf{a}_{t:t+H}$ with a time horizon of $H$, the resulting flattened token sequence is of length $D \times H$, which becomes inefficient for high-frequency trajectories or robots with high degrees of freedom, as hundreds of tokens may be required per action chunk - significantly increasing training complexity and inference latency. Recent work such as FAST~\citep{pertsch2025fast} employs compression-based tokenization using discrete cosine transform (DCT) encoding. However, it remains limited in capturing spatially structured dependencies essential for coordinated actions, and its variable-length tokenized chunks exhibit a dispersed distribution, increasing modeling complexity and susceptibility to decoding errors from incorrect predictions.

\begin{figure}[h]
  \centering
  \includegraphics[width=\linewidth]{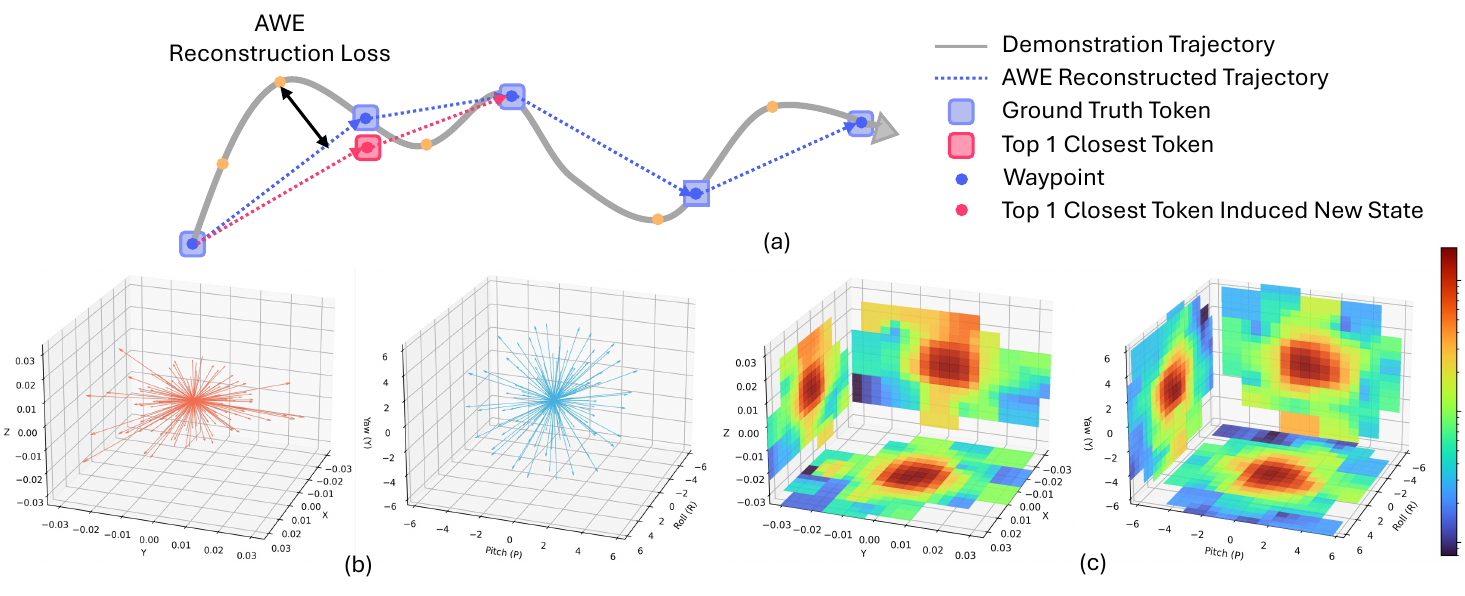}
   \caption{\textbf{Illustration of Spatial Action Tokenizer.} (a) Robot trajectories are decomposed into the shortest subsequence of states (waypoints) within an acceptable reconstruction error budget using AWE~\citep{shi2023waypoint}. (b) The \textbf{motion token library} is constructed by clustering delta actions from a large-scale, diverse dataset, with rotation and translation processed independently. During training, at each timestep, one of the top-3 closest tokens are randomly selected from the motion token library to approximate the next waypoint, the selected token then serves as the reference for determining the subsequent token. (c) shows the probability densities of delta actions derived from a diverse robot trajectory dataset, projected onto 2D planes.}
   \label{fig:action_tokenizer}
\end{figure}

To address these limitations, we introduce a \textbf{spatial action tokenizer} for representing motion sequences. Each action is expressed in the delta end-effector (EE) space, with rotations parameterized in $\mathrm{SO}(3)$~\citep{geist2024learning,gao2025towards}. Compared to joint-space representations, which depend on robot-specific configurations and joint dimensionality, EE-space deltas offer a more compact and cross-embodiment friendly formulation, being invariant to embodiment-specific kinematics and absolute positions or states. For example, an action such as opening a cabinet remains transferable regardless of which robot is performing the task and where the cabinet is located in space. 

We then focus on tokenizing delta end-effector motions, as illustrated in Fig.~\ref{fig:action_tokenizer}. To extract meaningful action deltas, we employ AWE~\citep{shi2023waypoint} to decompose each robot trajectory into a minimal set of waypoints whose linear interpolation approximates the original trajectory within a specified error threshold. Position and rotation are treated separately, using distinct distance metrics: point-to-line distance for position and rotational distance (after slerp interpolation) for rotation. The reconstruction thresholds are determined based on trajectory replays and human priors that reflect general action resolution requirements. The deltas between consecutive waypoints are modeled via k-means clustering, where each cluster centroid defines a motion primitive that is incorporated into the \textbf{motion token library}. 
Fig.~\ref{fig:action_tokenizer} (b) visualizes the resulting motion token library for left-arm motion, constructed with 150 clusters fitted on a large-scale, diverse dataset.
This tokenization approach preserves the spatial semantics of actions while mitigating irrelevant variability in data collection. In teleoperation, differences in operator proficiency and personal preferences often lead to variations in motion speed and micro-movements. The waypoint decomposition abstracts away such temporal and micro-motion discrepancies, while k-means clustering further suppresses residual micro-movement noise, thereby simplifying modeling and enhancing representational consistency. 

During training, we set a fixed maximum action horizon of 40 frames (equivalent to 1.33 seconds at a 30 Hz observation rate). The AWE~\citep{shi2023waypoint} algorithm is employed to extract waypoints and determine action deltas. To ensure consistent modeling horizons for $\Delta$xyz and $\Delta$SO(3), waypoint selection is synchronized such that a new waypoint is triggered whenever either translation or rotation exceeds its respective threshold. Note that different threshold choices define the resulting action compression rate. The same synchronization is enforced across the end-effectors of the left arm, right arm, and torso. For token assignment, we adopt a greedy selection strategy: for each state, the corresponding ground truth token is defined as the one whose application results in a state closest to the next waypoint. To enhance robustness against suboptimal inference-time predictions, we introduce a top-3 token sampling strategy: where the robot randomly selects among the three most relevant motion primitives. To prevent error accumulation, the robot state is updated after each token execution. Finally, we cap the number of tokens per end-effector delta translation/rotation at 5, with more complex and dexterous motions naturally corresponding to shorter prediction horizons. Note that in our spatial action tokenizer, each action token directly corresponds to a valid motion, making the system inherently more robust to prediction errors compared with FAST~\citep{pertsch2025fast}, which may produce invalid decoding when incorrect tokens are predicted. Furthermore, the motion token library explicitly defines valid action deltas, making it more resilient to data collection errors. For instance, an erroneous trajectory containing abrupt large movements will be approximated by a sequence of valid small-motion tokens.

\subsection{Combining Discrete and Continuous Action Representation}
Building on recent advances in generative modeling, VLA models have explored representing action distributions through diffusion~\citep{liu2024rdt, ze20243d, chi2023diffusion} or flow matching~\citep{black2024pi_0}, offering more expressive formulations for continuous value action chunks. However, as previously noted in~\citep{driess2025knowledge}, fine-tuning VLMs with continuous outputs often results in unstable training dynamics, as the learning signal must be propagated through continuous adapters (e.g. diffusion heads). This can degrade both the model's ability to interpret language instructions and the overall performance of the resulting VLA policy. To address this, we first train the VLM backbone on discretized actions during pre-training and then introduce an action expert to model the continuous action vector field through flow matching during fine-tuning. To further improve generalization, as well as sample and training efficiency of the action expert, we propose a novel pre-training stage for the action expert. In this stage, the action expert is trained to capture the broad unconditional distribution of $\mathbf{a}_{t:t+H}$ from a large-scale, diverse robot dataset, and is subsequently transformed into a conditional model by incorporating task-specific context during fine-tuning. 

\section{Training Recipe}

\begin{figure}[ht!]
  \centering
  \includegraphics[width=\linewidth]{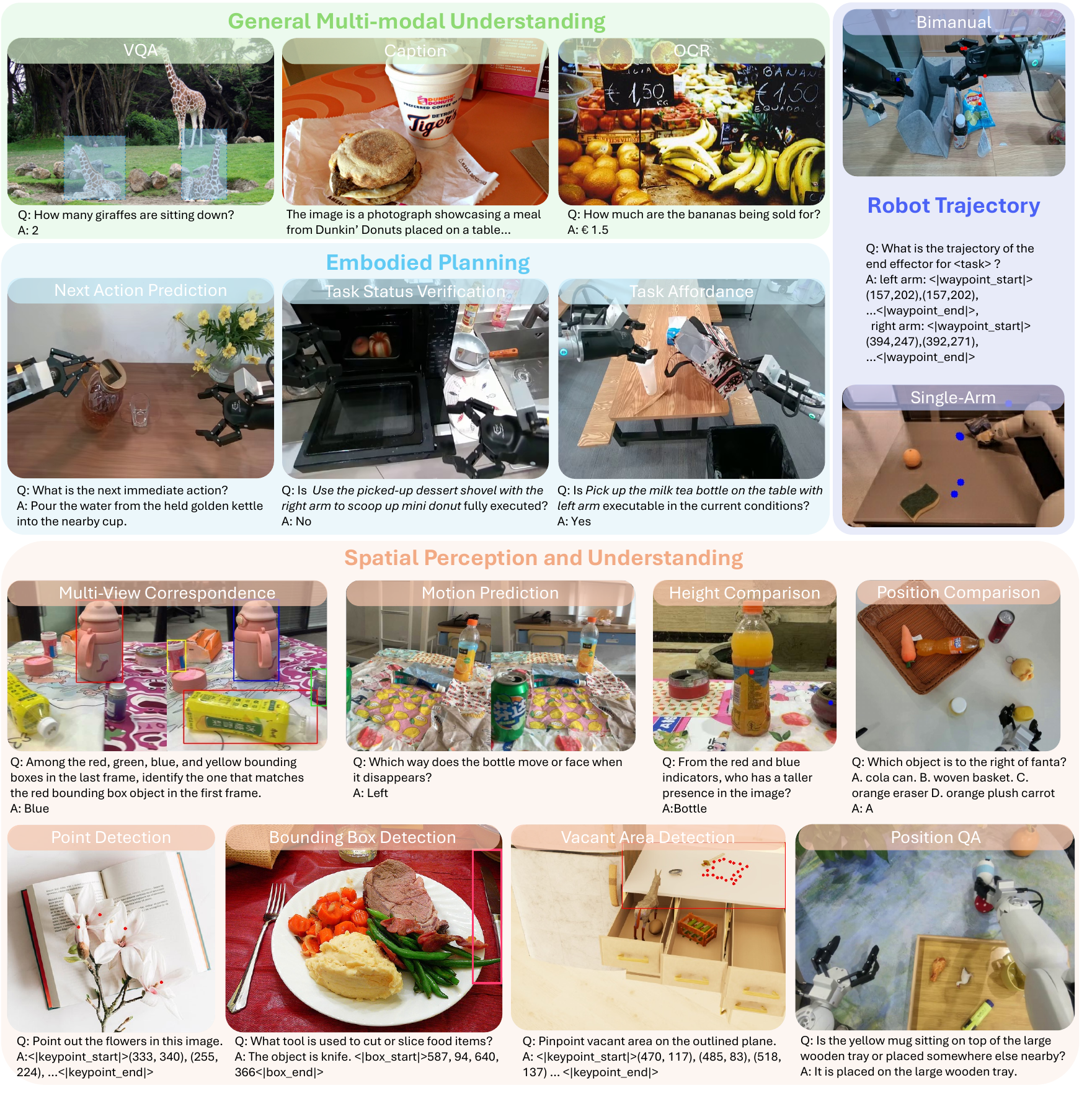}
   \caption{\textbf{Overview of Curated Vision-Language Data}. The curated dataset is designed to enhance core embodied reasoning abilities while preserving the general multi-modal understanding and reasoning capabilities of the pre-trained VLM. }
   \label{fig:vlm_data}
\end{figure}

We build \model by continuously training Qwen2.5-VL-7B~\citep{bai2025qwen2} on approximately 407 billion tokens through a three-stage training pipeline. First, we perform continued VLM pre-training to strengthen the model's embodied reasoning capabilities (Sec.~\ref{sec:stage1}). Second, we co-train on broad cross-embodiment robot data alongside vision-language data to instill awareness of robotic actions and enable action prediction (Sec.~\ref{sec:stage2}). Finally, we train on structured reasoning-action data to enable the model to systematically perceive, plan, and control (Sec.~\ref{sec:stage3}). For dynamic resolution, we specify only the minimum and maximum pixels to 3{,}136 and 230{,}400, respectively. This allows the number of image tokens to be primarily determined by each image's native resolution.

\begin{figure}[t!]
  \centering
  \includegraphics[width=\linewidth]{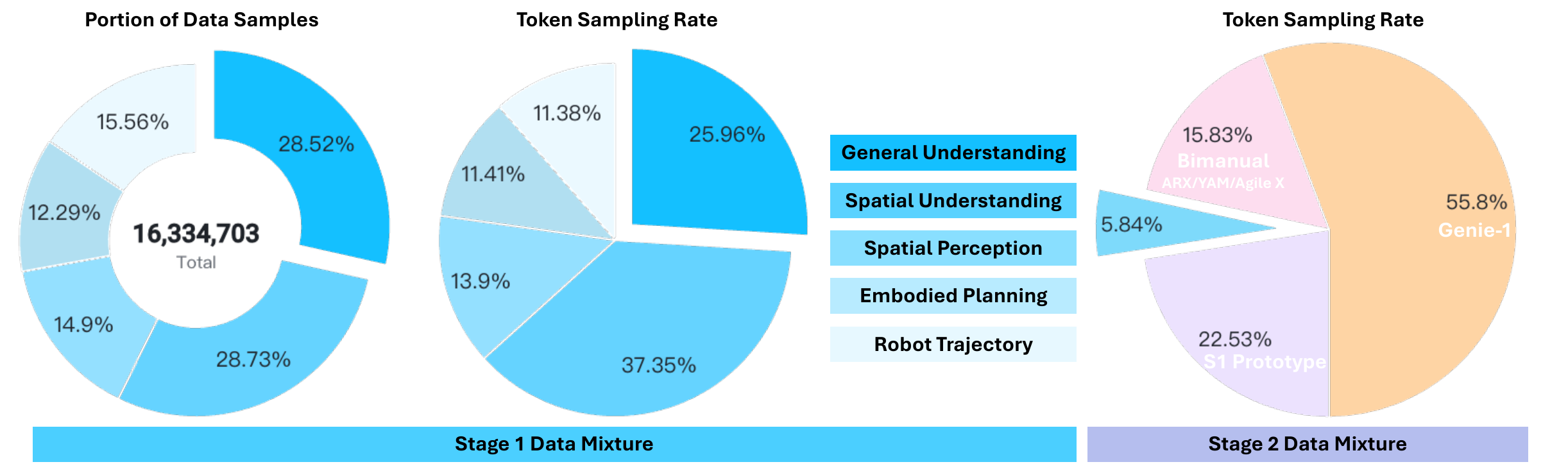}
   \caption{\textbf{Distribution of Data Mixture}: (Left) We curate a VLM dataset comprising roughly 16.3M samples that extend general multi-modal understanding with an emphasis on spatial perception, spatial reasoning, embodied planning, and robot trajectory prediction. During Stage1 continued VLM pre-training, we further prioritize spatial understanding as it forms the foundation of embodied reasoning. (Right) Stage2 co-trains on diverse cross-embodiment bimanual trajectories from Genie-1, Astribot S1 prototype, and bimanual ARX/YAM/Agile X, along with VLM data down-sampled to contribute $5.84\%$ of total training tokens. }
   \label{fig:data_mixture}
\end{figure}

\subsection{Stage 1: Continued VLM Pre-training}
\label{sec:stage1}
A key requirement for generalist robot policies is \textbf{embodied reasoning} - the ability to ground objects, relations, and knowledge in the physical world and to translate these grounded representations into effective action execution. Our goal is to develop a Vision-Language-Action model with embodied reasoning as its foundation, capable of generalizing across diverse scenarios while maintaining strong foundational multi-modal understanding. To achieve this, we construct a large-scale pre-training dataset comprising high-quality vision-language data, designed to enhance both robot-centric and general-purpose multi-modal capabilities, as illustrated in Fig.~\ref{fig:vlm_data}. This dataset emphasizes embodied planning, spatial perception, spatial reasoning, and trajectory generation, while also including data for general multi-modal understanding; the distribution of data mixture is illustrated in Fig.~\ref{fig:data_mixture} (Left). Below, we detail the task categories and data sources that form the basis of our continued VLM pre-training. 

\paragraph{General Multi-modal Understanding.}
To support general multi-modal understanding, we curate a large-scale dataset from open-source resources, including Cambrian-10M~\citep{tong2024cambrian}, LLaVA-665K~\citep{liu2025llava}, Pixmo Caption~\citep{deitke2025molmo}, Robo2VLM~\citep{chen2025robo2vlm} and Whatsup~\citep{kamath2023s}. The dataset spans key vision–language tasks such as VQA (covering perception, spatial reasoning, counting, math, and factual knowledge), captioning (dense scene descriptions), and OCR (scene text, handwritten text, documents, tables, and charts). Collectively, these resources establish a comprehensive foundation for training general-purpose vision–language models, which we leverage to preserve and enhance the broad multi-modal understanding capabilities of Qwen2.5-VL-7B~\citep{bai2025qwen2}.

\paragraph{Embodied Planning.}
To directly enhance the model's task planning capabilities in embodied scenarios, we construct a large-scale embodied planning dataset. We leverage open-source datasets, including EgoPlan~\citep{chen2023egoplan} and ShareRobot~\citep{ji2025robobrain}, and further generate planning tasks based on AGIBot~\citep{bu2025agibot} and Galaxea~\citep{jiang2025galaxea}. Task construction follows the ten task types defined in ShareRobot~\citep{ji2025robobrain}, focusing on three core task reasoning capabilities: \textbf{Next Action Prediction} (identifying the most plausible next action), \textbf{Task Status Verification} (determining whether a task is completed), \textbf{Action Affordance} (evaluating action feasibility), while excluding the \textit{Past Description} task which requires historical visual context. For affordance reasoning, we define two objectives - Feasibility and Achievability - to capture the reasoning variations required across different phases of task execution. Each task formulation uses ten GPT-generated prompt templates, with one randomly selected per sample. Images are sampled from subtask trajectories within specified temporal intervals according to the target objective. To ensure diverse subtask coverage, we employ a \textbf{frequency-based sampling} strategy: for each high-level task, subtask occurrence frequencies are computed, and questions are sampled proportionally, balancing both common and rare subtasks.

\paragraph{Spatial Perception.}
To enhance the model's spatial perception capabilities, we construct a suite of tasks centered on prediction of bounding boxes, keypoints, object parts and attributes by systematically filtering and restructuring open-source datasets. Specifically, PixmoPoint~\citep{deitke2025molmo} is processed following the RoboBrain2~\citep{team2025robobrain} pipeline, where we filter annotations by label density, discarding samples with more than ten labels. We further employ Qwen2.5-VL-7B-Instruct to retain only indoor scenes, and generate questions using 30 ChatGPT-designed prompt templates (e.g., ``Help me find all the \{label\} in the scene''). For PACO LVIS~\citep{ramanathan2023paco}, which originally provides only bounding boxes, we convert annotations into vision–language QA pairs. Using Qwen3-30B-A3B-Instruct-2507, we first filter indoor scenes, then generate three candidate questions per label and randomly select one. Both object-level and object-part questions emphasize functional use (e.g., ``What device can be moved to control the cursor on a screen?''; ``Which part of a handbag can be grasped to carry it?''). For What'sUp~\citep{kamath2023s}, a multiple-choice dataset for spatial relation understanding, we randomize the order of answer options (A–D) to mitigate positional bias. OCID REF~\citep{wang2021ocid} is reformulated into two complementary tasks: (1) object recognition, where the model identifies an object given its bounding box, and (2) object localization, where the task is to predict the bounding box given an object label. RoboPoint~\citep{deitke2025molmo} includes target-object annotations (bounding boxes or keypoints of specified objects) and vacant-region annotations (keypoints marking empty regions). Robo2VLM~\citep{chen2025robo2vlm} is structured as a multiple-choice task, requiring the model to select the correct answer from several candidates given an image. Finally, ShareRobot Affordance~\citep{ji2025robobrain} formulates affordance prediction as identifying the target location an agent should move to, conditioned on the task description.

\paragraph{Spatial Understanding.}  
We formulate Spatial QA as spatial understanding, primarily based on RefSpatial~\citep{zhou2025roborefer} and self-collected data. RefSpatial~\citep{zhou2025roborefer} organizes tasks into three categories: 2D, 3D, and simulator-based scenarios. The \textbf{2D tasks} include multi-turn question-answering, focusing on queries over relative positions, size comparisons, and direct spatial reasoning from questions. To ensure task relevance, Qwen3-30B-A3B-Instruct-2507 is applied to filter out instances that rely on absolute coordinate predictions.  The \textbf{3D tasks} extend spatial QA into 3D, including similar tasks analogous to 2D, as well as reasoning across paired views, localizing empty regions, and selecting objects that satisfy spatial relations from multiple-choice options. \textbf{Simulator tasks} are structured as multi-turn dialogues in which the system identifies target objects or vacant regions in response to questions. Beyond RefSpatial, we further construct spatial understanding tasks using both self-collected data and open-source RGBD video datasets, thereby enriching the diversity of spatial reasoning scenarios. From our self-collected robot data, we construct \textbf{Astribot Spatial Compass}, which features diverse QA and multiple-choice tasks. For each observation frame, we extract object bounding boxes, segmentation masks and depth maps using off-the-shelf vision foundation models such as SAM~\citep{kirillov2023segment} and VGGT~\citep{kirillov2023segment}. Combining image, mask, and depth information enables us to recover 3D bounding boxes for objects. We further employ Qwen2.5-VL-7B to generate both image-level and object-level captions. Building on these multi-modal annotations, we construct tasks that cover relative position (e.g., left, on, frontmost), object size, pixel location, vacant region, and object existence. These tasks are derived through either rule-based reasoning or QwQ~\citep{yang2025qwen3}, and presented in various formats, including question answering, true/false, multiple choice, and fill-in-the-blank.
We also source part of the data from SpaceVista~\citep{Sun2025SpaceVistaAV} covering five task categories: height or width comparison, counting, existence, object matching, and cross-frame position reasoning. For each category, we construct QA pairs with varying combination of input modalities, including text, image, bounding box, mask, and point. Visual cues such as bounding boxes, masks, and points are overlaid on the input images to indicate the corresponding regions of interest.

\paragraph{Robot Trajectory.}
This task focuses on predicting the motion trajectory of the robot end-effector projected onto 2D head camera images. We leverage the open-source MolmoAct~\citep{lee2025molmoact} auxiliary trace dataset and ShareRobot~\citep{ji2025robobrain}, and additionally construct trajectory QA pairs from the curated AGIBot beta~\citep{bu2025agibot} dataset, which spans diverse scenarios such as industrial production lines, supermarket retail, and household chores. For each fine-grained instruction, the corresponding waypoints are obtained by projecting the end-effector's 3D positions onto the head-camera observation using the camera's intrinsic and extrinsic parameters. Along each trajectory, 4–10 uniformly sampled waypoints are extracted for both the left and right arms. Initially, an equal number of waypoints are selected for each arm, after which redundant or static waypoints are removed.  

\paragraph{Training Details.}
We train on the curated VLM dataset for 7{,}000 steps, totaling 13.7B tokens using 128 H100 GPUs. A cosine learning rate schedule is adopted, starting at $5 \times 10^{-5}$ and decaying to $1 \times 10^{-5}$, with a linear warm-up over the first 5\% of total steps. The maximum sequence length per data sample is set to 4096. More detailed training configuration can be found in Supp. \ref{supp:training}.

\subsection{Stage 2: Co-Training on Cross-Embodiment Robot and VLM Data}
\label{sec:stage2}
The objective of Stage2 co-training is to endow the model with action prediction capabilities while preserving its general embodied reasoning abilities. To this end, the model is trained on a diverse suite of robotic tasks, allowing it to develop broadly transferable physical awareness and skills, jointly optimized with the curated VLM dataset from Stage1. The distribution of data mixture is illustrated in Fig.~\ref{fig:data_mixture} (Right).

\paragraph{Task and Robot Diversity.} In this stage, we emphasize training on \textbf{diverse} robotic behaviors to enable general understanding of physical actions. The training data encompass multiple robot platforms, including AGIBot Genie-1 (dual 7-DoF arms), Bimanual ARX, Bimanual Agile X, Bimanual YAM (all 6-DoF arms), and the prototype Astribot S1 (dual 7-DoF arms), which slightly differs from the configuration of the target Astribot S1 used in evaluation.
All robots are equipped with two wrist-mounted cameras and a base or onboard first-person-view front camera, operating at a control frequency of 30 Hz. The datasets cover a wide range of objects and environments, comprising 145 tasks in total. Each task involves complex, temporally extended behaviors that can be naturally decomposed into multiple subtasks. Actions are encoded via our spatial action tokenizer as introduced in Sec.~\ref{subsec:spatail_tokenizer}, fitted on diverse target-embodiment (Astribot S1) trajectories. Similar to the unified action space design of RDT-1B~\citep{liu2024rdt}, we adopt a compact tokenized representation of length 8, corresponding to:
\begin{center}
[\colorbox{blue!30}{$\Delta$xyz,  $\Delta$SO(3)}, \colorbox{red!30}{$\Delta$xyz, $\Delta$SO(3), Gripper}, \colorbox{green!30}{$\Delta$xyz, $\Delta$SO(3), Gripper}],
\end{center}
where the \textcolor{blue!30}{blue}, \textcolor{red!30}{red}, and \textcolor{green!30}{green} blocks denote actions for the torso, left arm, and right arm, respectively.
During training, our model predicts token sequences in multiples of 8, centered around 40 tokens, representing variable-horizon trajectory prediction (up to 1.33 seconds). 

\paragraph{Intra-Prompt Trajectory De-duplication.} 
Robot datasets often contain a substantial number of redundant trajectories, typically involving interactions with the same object in similar configurations, which tend to cluster near the center of the overall trajectory distribution. Such repeated data effectively resample the central region of an otherwise uniform trajectory distribution, increasing training costs while reducing action diversity. 
To address this, we propose an intra-prompt de-duplication method. Trajectories annotated with the same task and subtask are first projected onto three planes: x–y, x-z, and y-z, and each plane is discretized into a grid. For each plane, each trajectory is then represented as a boolean occupancy matrix over the grid, indicating which cells are occupied. Similar trajectories are identified by comparing their occupancy matrices. For single-arm trajectories projected on a single plane, let $O_1$ and $O_2$ denote the occupancy grids of trajectories $\tau_A$ and $\tau_B$, with $n_1$ and $n_2$ occupied cells, respectively. We define the normalized difference as $\mathcal{D}(\tau_A,\tau_B) = \frac{\sum (O_1 \oplus O_2)}{\max(n_1, n_2)}$, where $\oplus$ denotes the XOR operation. If $\mathcal{D}(\tau_A,\tau_B)$ falls below a pre-defined threshold, trajectory $\tau_B$ is considered redundant and removed from training. Occupancy matrices from both arms and all projection planes are considered jointly in the comparison.

\paragraph{Robot Trajectory Mirroring.}
Due to real-world data collection constraints, robot datasets often exhibit strong asymmetry between left- and right-hand trajectories. This arises primarily from operator handedness (most are right-handed) and positional bias during data collection - operators standing on one side of the workspace tend to favor the hand with a clearer field of view, leading to imbalanced demonstrations. However, such a human-induced asymmetry should not affect the learning of robotic policies. To address this, we apply a data mirroring strategy that transfers right-hand trajectories to the left-hand domain, enabling balanced training and evaluation even without left-hand data.
For visual data, the head-mounted (first-person) camera views are horizontally flipped. For wrist cameras, we first swap left- and right-hand image streams and then apply horizontal flipping, thereby synthesizing left-hand views from right-hand recordings (and vice versa).
For action data, each action is represented by the end-effector position and rotation. Mirroring is performed in two steps. For position $(x,y,z)$, given a base coordinate system where the positive directions correspond to forward, left, and upward, the flip is achieved by negating the $y$ component. For orientations, each arm maintains its own local base frame. The orientation is first transformed from the local base frame to the world frame, mirrored in the world frame, and then mapped back to the opposite arm's base frame.

\paragraph{Training Details.} Co-training on cross-embodiment and VLM data runs for 100{,}000 steps, processing a total of 200B tokens using 128 H100 GPUs. A constant learning rate of $1\times10^{-5}$ is applied, following a linear warm-up over the first 1\% of training steps. More detailed training configuration can be found in Supp. \ref{supp:training}.

\subsection{Stage 3: Target-Embodiment Action Training with Reasoning Process}
\label{sec:stage3}
This stage focuses on cultivating a structured reasoning process that enables purposeful action generation on the target embodiment, rather than simply memorizing trajectories.

\paragraph{Robot Platform.}
Our target embodiment is Astribot S1~\citep{dai2025co, gao2025towards}, a bimanual mobile manipulator equipped with two 7-DoF arms mounted on a highly articulated 4-DoF torso. Each arm features a parallel-jaw gripper capable of handling payloads up to \SI{10}{\kilogram}, enabling the manipulation of a broad range of everyday objects. The 4-DoF torso supports waist rotation, hip flexion, and knee-like articulation, allowing the robot to transition smoothly between standing and squatting postures. This design greatly enhances mobility and expands the effective workspace. The S1 achieves a vertical reach spanning from ground level to \SI{2}{\meter}, and a horizontal reach of up to \SI{1.94}{\meter} (including grippers). Overall, the platform is engineered for high performance, robustness, and operational safety in general-purpose manipulation tasks.

\begin{figure}[t]
  \centering
  \includegraphics[width=\linewidth]{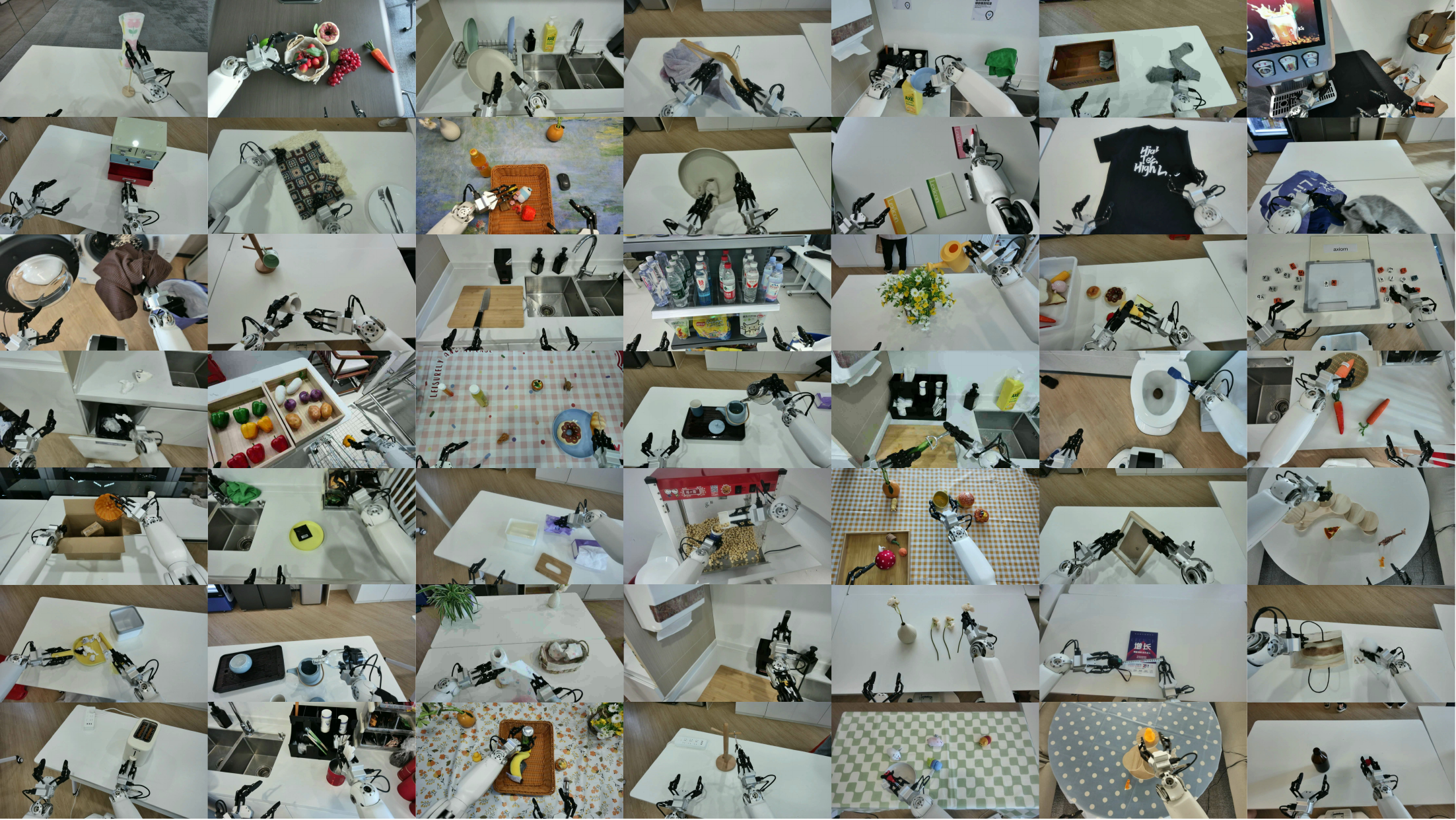}
   \caption{\textbf{Sample Tasks Collected on Astribot S1.} The tasks encompass a wide range of everyday activities, collected across diverse objects, lighting conditions, and environments.}
   \label{fig:s1_data}
\end{figure}

\paragraph{Robot Data Collection.}
\label{sec:s1_data}
As Illustrated in Fig.~\ref{fig:s1_data}, we collect large-scale and diverse Astribot S1 teleoperation trajectories~\citep{gao2025towards}. The data collection process is divided into three categories: (1) \textbf{General Pick and Place}, which represents the most common manipulation scenario in daily environments. For efficient data acquisition, each task requires the teleoperator to sequentially pick up seven items scattered on a table and place them into a container. To maximize data diversity, we employ a data collection scheduler that systematically guides teleoperators. The scheduler specifies (a) the task configuration - including object combinations, approximate positions, and background settings - and (b) the sequence of objects to be manipulated. At the start of each trajectory, a newly randomized configuration is generated,  guiding the operator to set up the environment accordingly. This mechanism enables precise control of the overall data distribution while introducing extensive variability across trajectories, resulting in a rich and well-balanced dataset. After collection, all trajectories undergo automated and manual quality validation, during which invalid or low-quality samples are excluded to maintain high dataset integrity; (2) \textbf{Specialized Pick and Place}: extends the general pick-and-place task by introducing additional challenges, such as:
(a) increasing object identification difficulty, for instance by presenting visually similar objects that differ only in color, texture, size, or position;
(b) increasing environmental variability, such as placing objects at different heights, by relative positions, or in vacant areas; 
(3) \textbf{Diversity Tasks}, which emphasizes diverse robot behaviors and language annotations. We collect a total of 50 tasks covering a broad spectrum of everyday activities. Each task involves complex, temporally extended behaviors that can be naturally decomposed into multiple subtasks, encompassing diverse motion primitives such as sweeping, peeling, pouring, brushing, folding, pressing, and rotating. The overall task semantics are provided in Supp.~\ref{supp:datasemantics}.

\paragraph{Reasoning Data Construction.} To directly enhance the model's reasoning and planning capabilities in embodied scenarios to enable purposeful action generation, we construct a large-scale embodied task-centric reasoning dataset based on self-collected Astribot S1 data. The dataset contains over 16.2 million frames, designed for structured action reasoning across multiple levels of granularity, broadly categorized into \textbf{textual reasoning} and \textbf{visual reasoning}. Details of the data construction process are provided in Supp.~\ref{supp:reasoningconstruction}.

\textbf{Textual reasoning} includes several forms of embodied cognition. \textbf{(1) Abstract concept reasoning} combines visual observations and textual instructions to infer implicit semantics. For example, as illustrated in Fig.~\ref{fig:reasoning_abstract}, the instruction ``put the high-calorie, caffeinated drink behind the yellow notebook'' induces the reasoning process ``the high-calorie, caffeinated drink refers to a regular cola can.'' \textbf{(2) Subtask reasoning} focuses on identifying the most plausible next step to achieve the intended goal. \textbf{(3) Visual observation description} emphasizes recognizing and describing key features of the scene and the target manipulable object. \textbf{(4) Movement reasoning} involves inferring the robot gripper's position and determining action directions in verbal form. For instance, as shown in Fig.~\ref{fig:model_architecture}, ``the right gripper is currently positioned to the right of the teapot lid. Move left and forward to align the gripper aperture with the lid's knob for grasping.''

\textbf{Visual reasoning} focuses on perception-grounded inference and motion prediction. Perception and grounding are represented through bounding boxes or keypoint coordinates, as illustrated in Fig.~\ref{fig:reasoning_long_horizon_part1}. The target manipulable object is localized via bounding box coordinates, while the corresponding placement or interaction region is denoted by keypoints, enabling flexible representation of vacant areas or multiple feasible target positions. In addition, action trajectory prediction is formulated as waypoint estimation corresponding to the action prediction horizon, effectively aligning 2D visual prediction with downstream action generation.

\paragraph{Reasoning Training Method.} To enable purposeful action prediction, we adopt a reasoning–action training paradigm that explicitly couples reasoning with action generation by sequencing reasoning and action tokens. In this formulation, reasoning is not treated as an auxiliary objective but serves to shape the policy representation toward purposeful and coherent actions. To ensure fast test-time deployment, the model is co-trained under two modes - determined by a system prompt - that control whether it should ``think'' before acting. These two modes are referred to as the full reasoning mode and the partial reasoning mode (which performs only subtask reasoning), respectively.
During inference, users can flexibly enable or disable the reasoning process based on task complexity. 

\paragraph{Training Details.}
We train on the curated dataset for 70{,}000 steps, totaling 193B tokens using 128 H100 GPUs. The Warmup-Stable-Decay~\citep{hu2024minicpm} strategy is applied to the learning rate schedule, consisting of a 1\% warmup phase and a 10\% final decay phase. More detailed training configuration can be found in Supp. \ref{supp:training}.

\paragraph{Reasoning for Long-Horizon Tasks.}
\label{sub:long_horizon_reasoning}
Training for long-horizon tasks presents unique challenges. When trained with flat (coarse) instructions, the model must implicitly infer which subtask to perform next, making it prone to error accumulation and out-of-distribution (OOD) behaviors. This setup also limits behavior transfer across tasks that share similar subtask-level skills.
An effective strategy, as explored in $\pi_{0.5}$~\citep{intelligence2025pi_}, is to first predict the semantic subtask - inferring the next appropriate behavior based on task structure and scene semantics - and then generate the corresponding low-level action chunk. Although conceptually simple, this approach often suffers from inconsistent subtask predictions during real-world deployment. Leveraging the inherent flexibility of our architecture, we introduce a reasoning strategy for fine-tuning long-horizon tasks that first predicts the completeness of the previously executed subtask and only generates the next subtask when the previous one is deemed complete. In this configuration, the model input consists of the flat (coarse) instruction and the last executed subtask (which can be set to None during the first inference step or as needed during execution). This additional subtask context serves as a short-term history, offering three key advantages:
(1) Faster inference: the model skips subtask generation when the previous subtask is still in progress;
(2) Reduced modeling complexity: the model avoids ambiguity in visually similar states - for instance, distinguishing between ``opening'' and ``closing'' a microwave door when it is half open - by instead judging whether the current subtask (e.g., ``open the door'') has been fully executed, a simpler decision that is less confusing and could effectively leverage pre-trained embodied VLM capabilities; and
(3) Improved behavioral consistency: when multiple valid behaviors exist under the same instruction (e.g., cleaning a table by picking up various items), the model consistently completes one subtask before initiating another. Similarly, when an object can be manipulated by either arm, maintaining the previous subtask context ensures consistent arm usage rather than oscillating between alternatives.

\section{Reinforcement Learning for Reasoning-Action Alignment}
\label{sec:rl}
We observe that after Stage3 pre-training, the model exhibits strong action execution performance under in-domain data distributions. However, certain reasoning issues persist~\citep{fang2025robix}: (1) unreasonable textual reasoning or imprecise visual reasoning, such as incorrect subtask prediction or inaccurate bounding box prediction; (2) inconsistencies among reasoning components, e.g., misalignment between textual reasoning, visual predictions, and trajectory motions. To address these issues, we leverage Reinforcement Learning (RL) to refine embodied reasoning and strengthen the alignment between high-level reasoning and low-level action.

\subsection{RL Data Selection} 
To correct erroneous reasoning outcomes and further enhance the model's capability, we adhere to the Stage3 data configuration, which includes both coarse and fine-grained instructions, as well as full reasoning and partial reasoning modes. First, candidate data are sampled, with keyframes (e.g., transitions near subtask completion) assigned higher sampling weights. Subsequently, to enhance practical applicability, we augment the training data by randomly modifying textual instructions (e.g., altering target objects or placement locations). These augmented samples are not thereafter supervised on trajectory or action due to the lack of ground truth. Finally, to maximize training efficiency, we pre-filter samples based on inference results and reward variance, retaining those that are informative and discarding samples that are either trivial or excessively challenging. This ensures that the remaining data provide sufficient gradient information during GRPO~\citep{shao2024deepseekmath} training.

\subsection{RL Reward Design} 
The design of reward supervision varies across data types. Specifically, rewards are categorized as follows:

\textbf{Visual Reward.} Due to the limitations of next-token prediction - which assigns equal weight to all output tokens and fails to account for the precision requirements of specific reasoning components - training convergence becomes slow. To address this, we implement specialized reward supervision for critical visual outputs, including IoU reward for bounding boxes, an accuracy reward for keypoints, and a distance reward for waypoints. These rewards can be computed as follows:
\begin{equation}
  r_{\text{bbox}} = \text{IoU}(b_1, b_2)
\end{equation}
Here, IoU denotes the intersection-over-union metric. The IoU reward encourages the model to generate bounding boxes that closely match the ground truth and is applied only during the ``pick'' phase.
\begin{equation}
  r_{\text{keypoint}} = \frac{\sum_{i=1}^{N} \mathbb{I}(\text{k}_i \in b)}{N},
\end{equation}
where $N$ represents the total number of keypoints, $\mathbb{I}(\text{k}_i \in b)$ yields a value of 1 if the $i$-th keypoint $k_i$ is located within the ground truth bounding box region $b$, and 0 otherwise. The accuracy reward measures the fraction of predicted keypoints falling within the ground-truth region, thereby promoting precise placement prediction, and is applied only during the ``place'' phase.
For the waypoint reward, inspired by recent work~\citep{huang2025thinkact}, we evaluate the spatial consistency between the predicted and ground truth trajectories by measuring positional deviation at key points (start/end positions) and along the full trajectory.  Trajectories for both arms are normalized to $[0,1]$, with point-wise deviations measured by Euclidean distance and full-trajectory similarity assessed by Dynamic Time Warping (DTW) \citep{senin2008dynamic} distance: 
\begin{equation}
\begin{aligned}
  r_{\text{waypoint}} &= 0.5 \ r_{\text{goal}} + 0.5 \ r_{\text{traj}} \\
  r_{\text{goal}} &= \frac{1}{2} \left[\max \left( 0, 1 - \|p_1 - \hat{p}_1 \|_2^2 \right) + \max \left( 0, 1 - \| p_n - \hat{p}_n \|_2^2 \right)\right] \\
  r_{\text{traj}} &= \max \left( 0, 1 - \text{dtw}\left(\tau, \hat{\tau}\right) \right)
\end{aligned}
\end{equation}
Here, $\tau = \{p_1, \cdots, p_n\}$ and $\hat{\tau} = \{\hat{p}_1, \cdots, \hat{p}_n\}$ denote the ground-truth and the predicted trajectories, respectively. The ground truth is obtained by projecting the 3D positions of the end-effector onto head-camera observations, ensuring consistency with the action.

\textbf{Consistency Reward.} To ensure textual correctness and coherence among reasoning components, we adopt a VLM-based evaluation scheme following ~\citep{fang2025robix}. Specifically, we use Qwen3-VL-32B-Instruct~\citep{qwen3vl} to assess the plausibility of textual predictions (text reasonableness) and their alignment with spatial inferences (text-spatial consistency). We design a specific prompt that instructs the model to make judgments based on the input image, the instruction, the ground-truth text, and the model's own parsed output; the detailed evaluation prompt template is provided in the Supp.~\ref{supp:rlevalprompt}. The corresponding reward is defined as follows:
\begin{equation}
  r_{\text{consistency}} = 0.5 \ r_{\text{text}} + 0.5 \ r_{\text{text-spatial}} 
\end{equation}

\textbf{Action Reward.} Since the ultimate purpose of reasoning is to drive correct physical execution, we introduce a dedicated action reward. Previous approaches typically rely on a critic model to evaluate task progress~\citep{ye2023reinforcement, zhai2025vision} or employ an outcome reward indicating task success~\citep{li2025simplevla} in specific benchmark environments such as LIBERO~\citep{liu2023libero}. In contrast, we directly use action prediction errors as supervisory signal. Considering that the temporal length of action prediction is adaptive, and the variability of reward magnitudes, we focus solely on the unified action space at the final timestep of each prediction. At this timestep, we separately compute errors for position, rotation, and gripper state. These error components are then aggregated into a final reward via exponential decay and a weighted sum with manually tuned hyperparameters:
\begin{equation}
  r_{\text{action}} = \sum_{i=1}^{8} \left[w_i \cdot r_{\text{action}}^{(i)}\right] ,\quad \text{where} \quad r_{\text{action}}^{(i)} = \exp\left(-k_i \cdot f_i\left(a^{(i)}, \ \hat{a}^{(i)}\right)\right)
\end{equation}
Here, $f_i$ denotes the error function for the $i$-th action component, $k_i$ is a decay coefficient, and $w_i$ is a weighting factor. $\{a^{(1)}, \cdots, a^{(8)}\}$ and $\{\hat{a}^{(1)}, \cdots, \hat{a}^{(8)}\}$ denote the ground-truth and the predicted continuous action sequences.

\textbf{Format Reward.} To ensure structural compliance across reasoning modes and subtask stages, we apply regex-based matching to enforce adherence to predefined output formats. A binary reward (1 or 0) is assigned based on whether the generated output conforms to predefined templates. This format reward is incorporated into the overall reward with a weight of 0.1.

\subsection{RL Training Details} 

During training, we observe a phenomenon consistent with~\citep{li2025simplevla}: despite the application of diverse task designs and multi-modal augmentations (in text, images, and actions), the model tends to converge to a narrow solution pattern characterized by low reward variance. To promote exploration, we adopt techniques inspired by DAPO~\citep{Yu2025DAPOAO}, such as clip-higher strategy, higher sampling temperature, while applying a small KL divergence penalty. More detailed training settings and hyperparameters can be found in Table \ref{tab:rl_configuration}.

\begin{table}[htbp]
\centering
\setlength{\tabcolsep}{10pt}
\begin{tabular}{lc||lc}
\hline
\multicolumn{2}{c||}{\textbf{Basic Training Configuration}} & \multicolumn{2}{c}{\textbf{Generation and Training Parameters}} \\
\hline
\hline
Trainable Part & Full Model & \multicolumn{2}{l}{\textbf{Generation Parameters}} \\
Per-device Batch Size & 2 & Temperature & 1.6 \\
Peak LR & $1\times 10^{-6}$ & Top\_p & 1.0 \\
Training Epoch & 1 & Top\_k & 50 \\
Optimizer & AdamW & Repetition\_penalty & 1.0 \\
Weight Decay & 0.0 & & \\
Warmup Ratio & 0.00 & \multicolumn{2}{l}{\textbf{Training Parameters}} \\
LR Schedule & Cosine & KL\_coefficient & 0.04 \\
Max Seq. Length & 2048 & Epsilon\_high & 0.28 \\
Max Compl. Length & 2048 & Epsilon\_low & 0.2 \\
Num. of Compl. & 8 & Importance\_sampling\_level & token \\
GPU Nums & $4\times 8$ & & \\
\end{tabular}
\caption{\textbf{Configuration for Reasoning-Action Reinforcement Fine-tuning}, detailing the basic training parameters (left) and the generation as well as additional training parameters (right).}
\label{tab:rl_configuration}
\end{table}

\section{Experiments}
We conduct extensive experiments to thoroughly evaluate the performance of \model, focusing on six key research questions:
\begin{itemize}
\item $\mathcal{Q}1$: Does \model effectively enhance embodied reasoning capabilities?
\item $\mathcal{Q}2$: Does training on cross-embodiment robot data facilitate learning on target robots?
\item $\mathcal{Q}3$: Does training on structured reasoning traces lead to more purposeful action generation and out-of-distribution generalization?
\item $\mathcal{Q}4$: Can reinforcement learning further enhance model capability?
\item $\mathcal{Q}5$: Is \model capable of few-shot adaptation on long-horizon and dexterous tasks?
\item $\mathcal{Q}6$: Can we derive meaningful scaling laws for training generalist robot policies?
\end{itemize}

\subsection{VLM Evaluation [$\mathcal{Q}1$]}
To evaluate \model's embodied reasoning capability, we assess the model after continued VLM pre-training (\model-Stage1), and after co-training on cross-embodiment robot action data and VLM data (\model-Stage2), across public benchmarks focused on spatial understanding and perception. We compare \model against state-of-the-art general multi-modal models, including Qwen2.5-VL-7B and Qwen2.5-VL-32B~\citep{bai2025qwen2}, as well as dedicated embodied reasoning VLMs including RoboBrain-7B-2.0~\citep{team2025robobrain} and Robix-7B~\citep{fang2025robix} on 7 benchmarks: BLINK~\citep{Fu2024BLINKML}, CV-Bench~\citep{tong2024cambrian}, EmbSpatial~\citep{duetal2024embspatial}, RefSpatial-Bench~\citep{duetal2024embspatial}, SAT~\citep{ray2024sat}, Where2Place~\citep{yuan2024robopoint} and RoboSpatial~\citep{song2025robospatial}. As shown in Table~\ref{tab:vlm}, \model\ outperforms its backbone (Qwen2.5-VL-7B-Instruct) on 6 out of 7 benchmarks and surpasses specialized embodied models RoboBrain-7B-2.0 and Robix-7B on most benchmarks after Stage1 pre-training. These results underscore \model's strong performance in object localization, spatial referencing, and fine-grained visual understanding. Furthermore, these capabilities remain largely preserved after Stage2 co-training on diverse robot trajectories, indicating that the integration of action learning does not compromise the model's core multi-modal reasoning and perception abilities.

\begin{table}[]
\centering
\resizebox{\textwidth}{!}{%
\begin{tabular}{@{}lccccccc@{}}
\toprule
\multirow{2}{*}{Models/Metrics} & 
\multicolumn{1}{c|}{CV-Bench} & 
\multicolumn{1}{c|}{EmbSpatial} & 
\multicolumn{3}{c|}{Where2Place} & 
\multicolumn{2}{c}{RoboSpatial} \\
\cmidrule(lr){2-2} \cmidrule(lr){3-3} \cmidrule(lr){4-6} \cmidrule(l){7-8} 
& All & All & Seen & Unseen & All & All (mask) & All (point) \\
\midrule
\textbf{General Baselines} & & & & & & & \\
\midrule
Qwen2.5-VL-7B-Instruct & 79.10 & 71.26 & 10.00 & 13.10 & 10.93 & 46.45 & 45.71 \\
Qwen2.5-VL-32B-Instruct & 81.78 & 74.59 & 17.98 & 30.06 & 21.61 & 51.84 & 51.14 \\
\midrule
\textbf{Embodied Baselines} & & & & & & & \\
\midrule
RoboBrain-7B-2.0 & 85.81 & \uline{75.88} & \uline{68.84} & \uline{64.87} & \uline{66.06} & \uline{61.06} & \uline{55.14} \\
Robix 7B Base* & \textbf{86.50} & \textbf{77.40} & - & - & 41.90 & - & - \\
\midrule
\textbf{Ours} & & & & & & & \\
\midrule
Lumo-1-Stage1 & \uline{86.36} & 75.60 & \textbf{70.05} & \textbf{66.75} & \textbf{69.06} & \textbf{62.57} & \textbf{57.14} \\
Lumo-1-Stage2 & 84.93 & 71.68 & 65.81 & 60.07 & 64.09 & 56.59 & 52.00 \\
\bottomrule
\toprule
\multirow{2}{*}{Models/Metrics} & 
\multicolumn{3}{c|}{BLINK} & 
\multicolumn{3}{c|}{RefSpatial-Bench} & 
\multicolumn{1}{c}{SAT} \\
\cmidrule(lr){2-4} \cmidrule(lr){5-7} \cmidrule(l){8-8} 
& Depth & Spatial & All & Location & Placement & All & All \\
\midrule
\textbf{General Baselines} & & & & & & & \\
\midrule
Qwen2.5-VL-7B-Instruct & 71.77 & \textbf{90.21} & 81.65 & 10.47 & 3.46 & 6.96 & 62.67 \\
Qwen2.5-VL-32B-Instruct & 74.19 & \uline{83.22} & 79.03 & 12.99 & 9.50 & 11.24 & 69.33 \\
\midrule
\textbf{Embodied Baselines} & & & & & & & \\
\midrule
RoboBrain-7B-2.0 & \uline{85.48} & 82.52 & \uline{83.90} & \uline{46.10} & \uline{36.03} & \uline{41.07} & \uline{72.67} \\
Robix 7B Base* & - & - & \textbf{87.60} & - & - & - & 71.10 \\
\midrule
\textbf{Ours} & & & & & & & \\
\midrule
Lumo-1-Stage1 & \textbf{87.90} & 77.60 & 82.40 & \textbf{51.99} & 50.00 & \textbf{51.00} & \textbf{74.67} \\
Lumo-1-Stage2 & 85.48 & 76.92 & 80.90 & 49.50 & \textbf{52.00} & 50.75 & 69.33 \\
\bottomrule \\
 * Reported by Robix. \\
\end{tabular}
}
\caption{\textbf{Performance of Lumo-1 on Embodied Reasoning Related Benchmarks.} The highest score within each group is highlighted in \textbf{bold}, while the second-highest score is \underline{underlined}.}
\label{tab:vlm}
\end{table}

\subsection{Generalizable Pick and Place [$\mathcal{Q}2-\mathcal{Q}3$]}
We evaluate on the task of generalizable pick and place with ``put A into/to B'' instruction, where A is the object and B is the target location. We evaluate 4 models: 
\begin{itemize}
\item \textbf{\model-Stage1-PNP}: General Pick-and-Place model trained for action prediction only, initialized from the \model-Stage1 checkpoint.
\item \textbf{\model--Stage2-PNP}: General Pick-and-Place model trained for action prediction only, initialized from the \model-Stage2 checkpoint.
\item \textbf{\model-Stage3}: Reasoning-augmented model trained jointly for reasoning generation and action prediction.
\item \textbf{$\pi_0$-PNP}: General Pick and Place model trained by fine-tuning $\pi_0$ on the same data as \textbf{\model-Stage1-PNP} and  \textbf{\model-Stage2-PNP}.
\end{itemize}

\paragraph{Evaluation Settings.} 
We evaluate \model\ under four settings similar to GR3~\citep{cheang2025gr3}: (1) \textbf{Basic}, (2) \textbf{Unseen Environments}, (3) \textbf{Unseen Instructions}, and (4) \textbf{Unseen Objects}. In Basic, we evaluate in an environment that is seen during training. We use 60 training-seen objects to access the model's basic instruction-following ability. In Unseen Environments, the same set of objects is evaluated across 3 distinct environments that are unseen during training. In Unseen Instructions, the model is prompted with instructions that demand higher-level conceptual understanding, such as spatial or semantic reasoning (e.g. ``put the [left coke] / [high-calorie drink] into the round woven basket'' ). In Unseen Objects, evaluation is performed on 105 novel objects that are absent from the training dataset, testing the model's ability to generalize to unseen items.

We evaluate model performance using two metrics: instruction-following rate (IFR) and task success rate (SR). The IFR reflects how accurately the robot identifies and approaches the object/location indicated in the instruction, while the SR measures whether the robot successfully completes the given instruction. Both metrics are reported as percentages, with higher values indicating better instruction comprehension and task execution capabilities. For reasoning-augmented models, we also examine the reasoning outputs, which offer interpretable insights into the model's action selection process.
\begin{figure}[ht!]
  \centering
  \includegraphics[width=\linewidth]{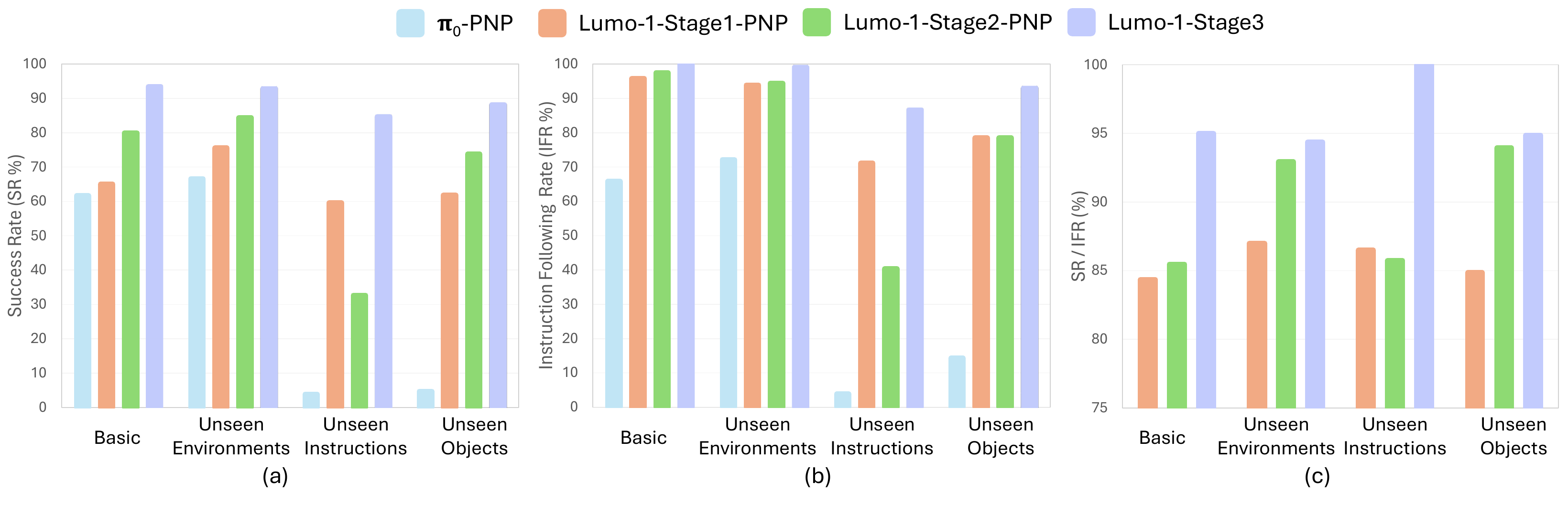}
   \caption{\textbf{Experiment Results of Generalizable Pick and Place}. As shown in (a), \model-Stage3 consistently outperforms the baseline $\pi_0$ and earlier training stages across all four evaluation categories. In (b), it demonstrates further improvement in semantic understanding and reasoning capability relative to \model-Stage1, and in (c), it exhibits enhanced action execution accuracy over \model-Stage2, as reflected by the IFR/SR metric.}
   \label{fig:generalizable_pnp_results}
\end{figure}

\textbf{Basic Instruction Following.}
For both the \textbf{Basic} and \textbf{Unseen Environments} settings, we partition the 60 training-seen objects into 12 mini-batches of 5 objects each. In every evaluation episode, the model is instructed to identify and pick a target object from the candidate set according to the given instruction, and continue this process until all objects are placed into the container. Each mini-batch is evaluated twice under two distinct layout configurations, resulting in a total of 120 ``put A into/to B'' evaluations. To ensure fair comparison across models, object layouts of each mini-batch are kept as consistent as possible throughout evaluation.

As shown in Fig.~\ref{fig:generalizable_pnp_results} (a), \model consistently outperforms $\pi_0$ across four evaluation categories, with Stage1–3 training progressively improving the success rate. All models exhibit strong robustness in unseen environments, highlighting the benefits of large-scale pre-training. Fig.~\ref{fig:generalizable_pnp_results} (b) further shows that \model (Stage1–3) demonstrates superior instruction-following capability compared to $\pi_0$, while Fig.~\ref{fig:generalizable_pnp_results} (c) illustrates that Stage2 pre-training significantly enhances action accuracy in unseen environments, improving from $86.98\%$ to $92.95\%$.

\begin{figure}[h!]
  \centering
  \includegraphics[width=\linewidth]{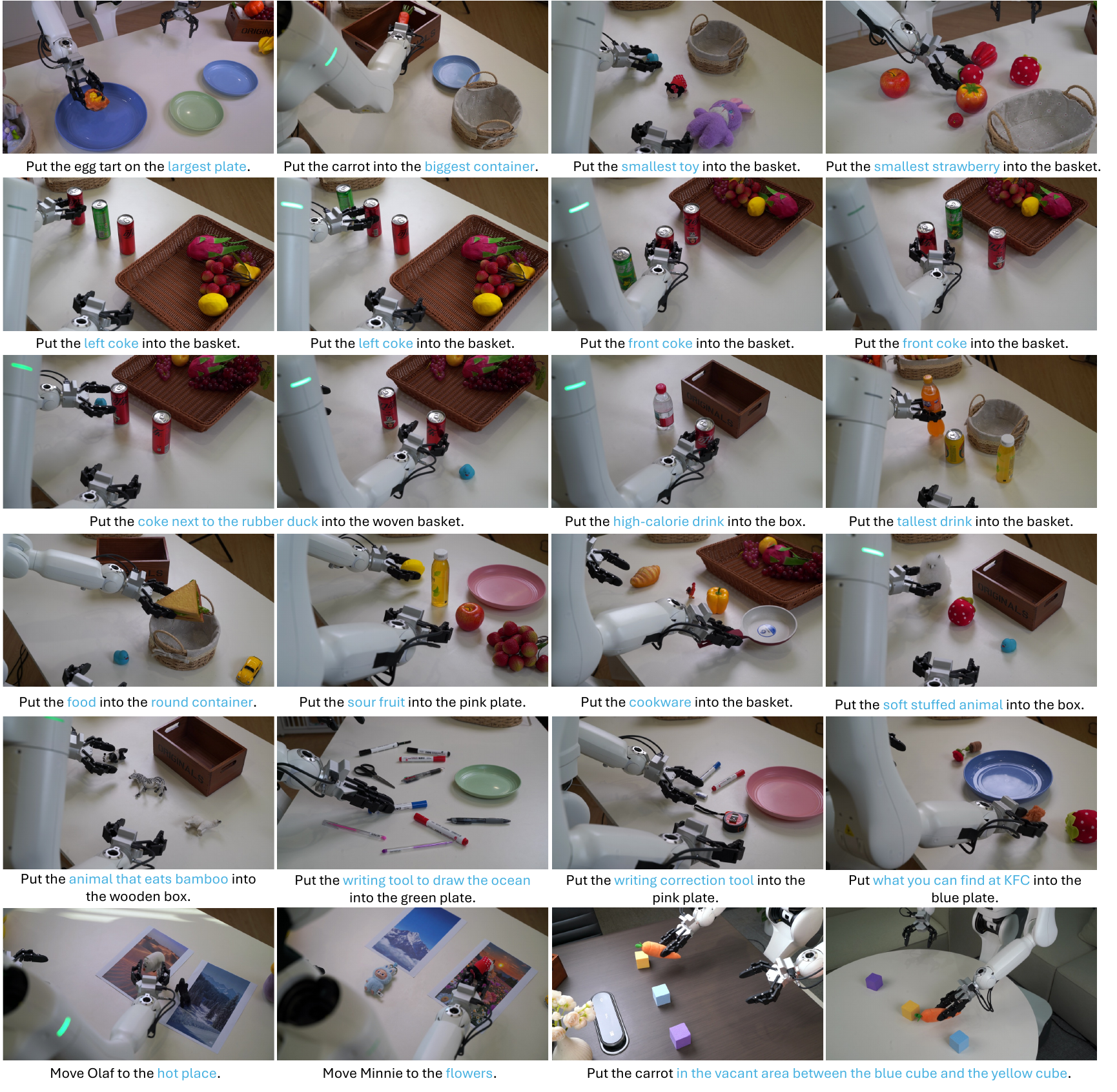}
   \caption{\textbf{Generalization and Instruction Following Capabilities.} \model demonstrates strong instruction following capability and further shows the ability to generalize to unseen, conceptually abstract prompts (marked by blue).}
   \label{fig:general_pnp}
\end{figure}

\begin{figure}[h!]
  \centering
  \includegraphics[width=\linewidth]{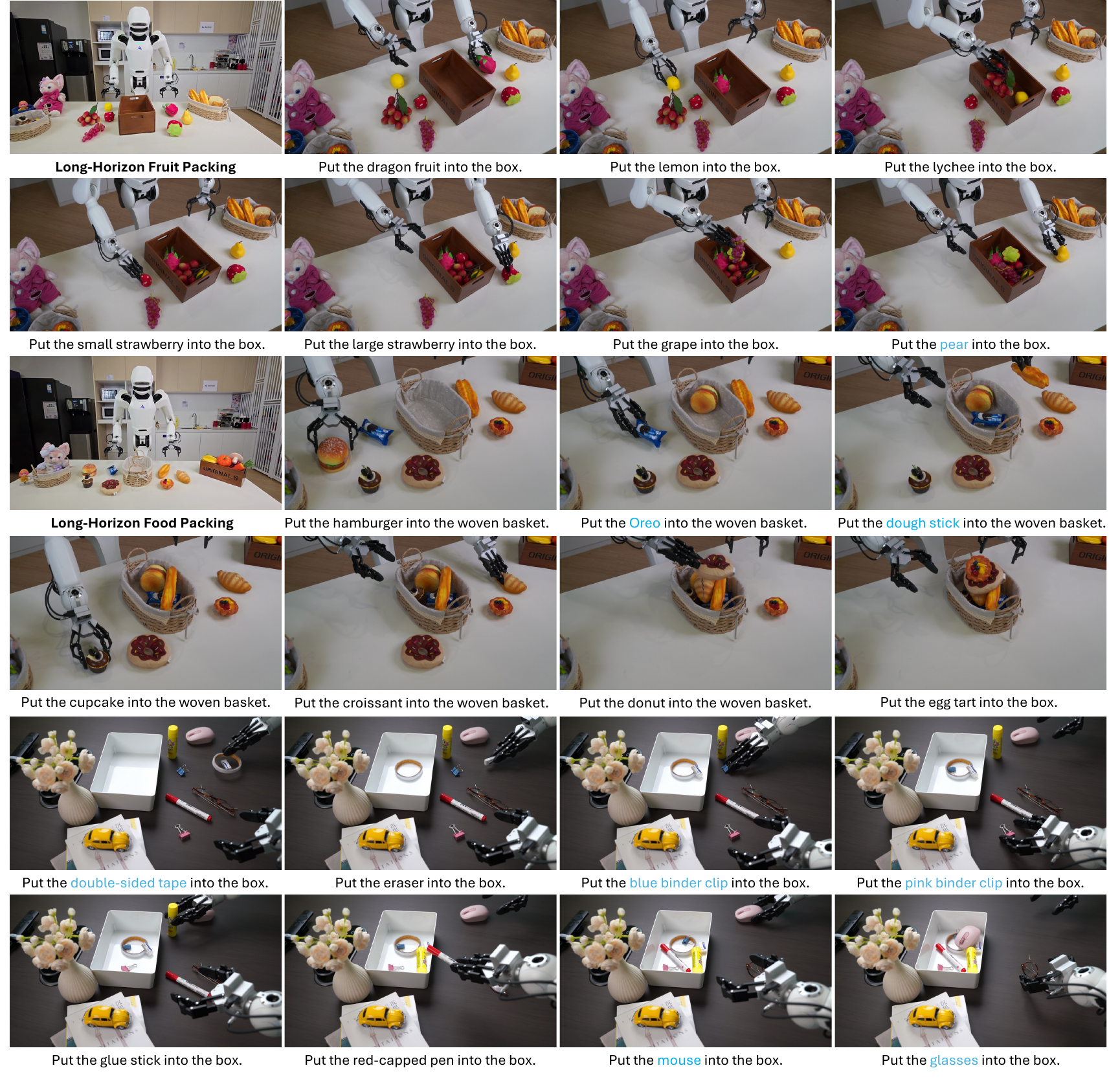}
   \caption{\textbf{Long-Horizon Packing Tasks.} \model exhibits strong instruction following capability, extends effectively to long-horizon packing tasks, and generalizes robustly to previously unseen objects (highlighted in blue).}
   \label{fig:general_pnp_long_horizon}
\end{figure}

\textbf{Generalizable Instruction Following}. The \textbf{Unseen Instructions} setting evaluates the model's ability to comprehend abstract concepts involving size, spatial relations, and common-sense reasoning. Example prompts include ``put the largest strawberry on the plate'', ``move the animal that eats bamboo to the white storage box''.  The evaluation instructions do not appear in the robot training data and require the model to interpret complex semantics. We construct a set of 26 such instructions, each evaluated twice under distinct layout configurations to reduce evaluation variance. The \textbf{Unseen Objects} setting assesses generalization to novel items. We partition 105 unseen objects into 21 mini-batches of 5 objects each, and for each trial, the model must identify and manipulate a target object. Similar to previous evaluations, we assess performance under two distinct layout configurations. 

As shown in Fig.~\ref{fig:generalizable_pnp_results} (a), $\pi_0$ performs poorly on ``hard cases'' involving unseen instructions and unseen objects, likely due to overfitting and degradation of VLM knowledge after large-scale flow-matching fine-tuning. Under the \textbf{Unseen Instructions} setting, Stage2 also under-performs Stage1, as extensive cross-embodiment action training with severely down-sampled VLM data leads to diminished semantic understanding. This degradation is mitigated by Stage3 training, which reinforces reasoning capabilities more aligned with VLM pre-training. Notably, in Fig.~\ref{fig:generalizable_pnp_results} (c), the Stage2 model outperforms Stage1 on both the \textbf{Unseen Environments} and the \textbf{Unseen Object} benchmark, exhibiting significantly higher SR/IFR scores - suggesting that cross-embodiment training across diverse scenarios enhances action accuracy and generalization to novel environments and objects. See Fig.~\ref{fig:general_pnp} and Fig.~\ref{fig:general_pnp_long_horizon} for sample model rollouts.

\begin{figure}[htbp]
  \centering
  \includegraphics[width=\linewidth]{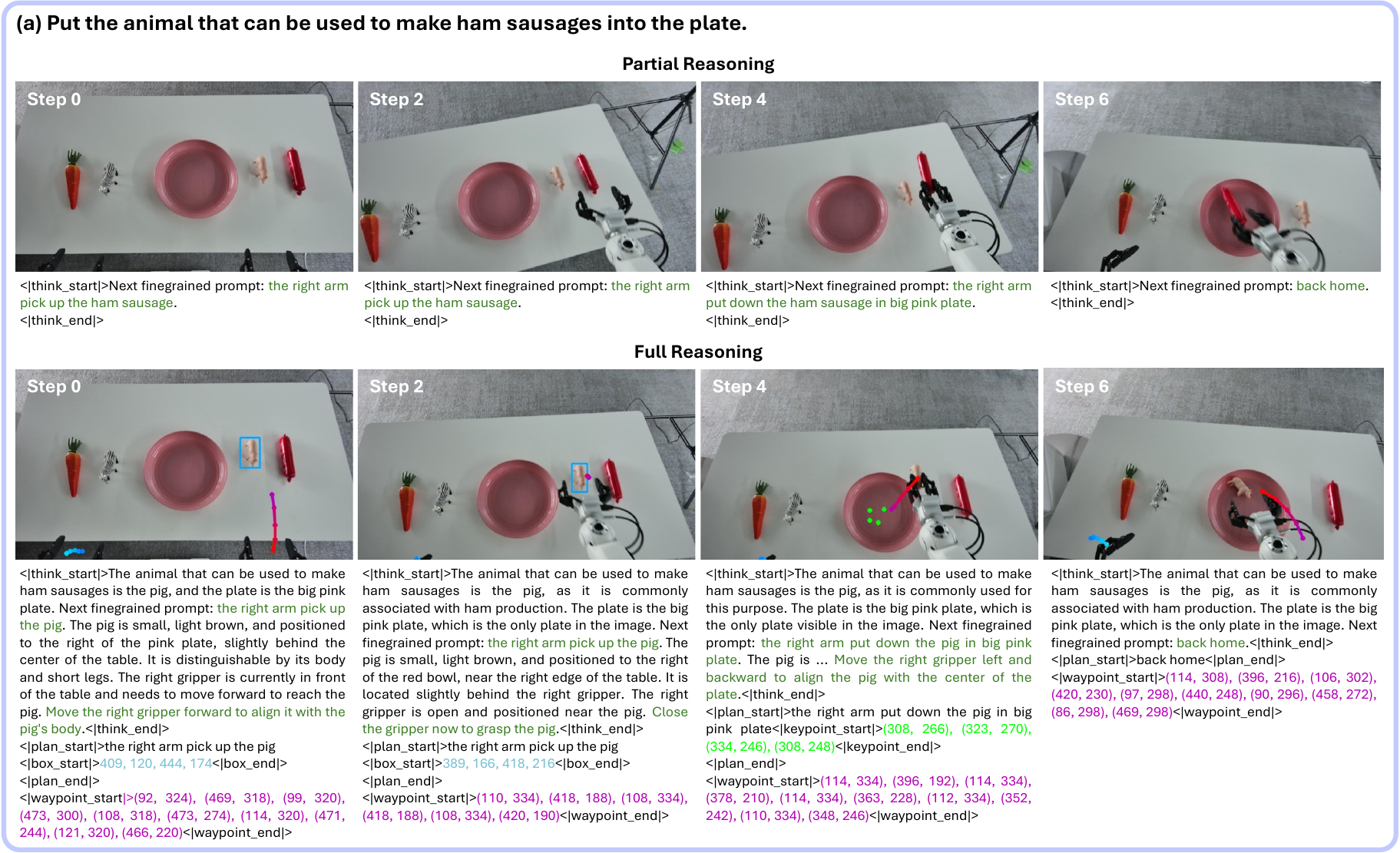}
  \includegraphics[width=\linewidth]{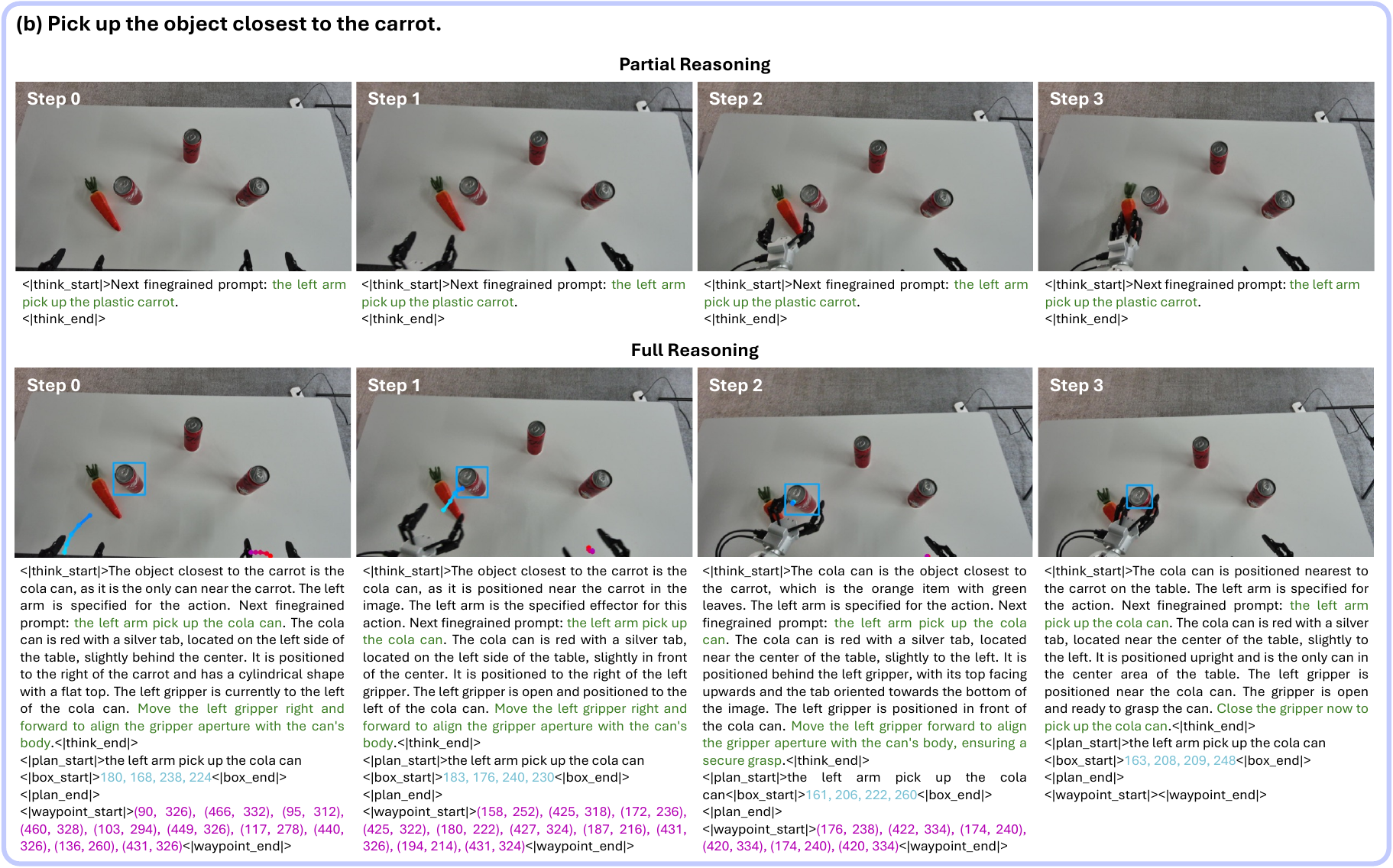}
  \caption{\textbf{Partial Reasoning vs. Full Reasoning.} The full reasoning mode of \model shows superior understanding capability than the partial reasoning mode when faced with complicated instructions. In this mode, bounding boxes are shown as blue rectangles, key points as green dots, the end effector's waypoints as gradient polylines.}
  \label{fig:full_vs_partial}
\end{figure}

The experimental results also reveal that the partial reasoning mode occasionally produces reasoning errors when encountering complex or ambiguous instructions, whereas the full reasoning mode further strengthens the comprehension of \model. As illustrated in Fig.~\ref{fig:full_vs_partial}, the next fine-grained prompt generated under partial reasoning extracts only the explicitly mentioned objects when encountering unfamiliar instructions, leading to incorrect reasoning and subsequent execution failures. In contrast, the full reasoning mode follows a chain-of-thought procedure. It first parses the intended grasp and placement targets, then generates the next fine-grained prompt along with guidance text that includes object descriptions and action refinements, and finally outputs visual aids for grounding (bounding boxes or key points) and waypoints for action execution. This chain-of-thought approach effectively handles complex instructions and progressively bridges reasoning to action, thereby improving execution performance. Nevertheless, full reasoning incurs a higher latency. Given that partial reasoning approaches the performance of full reasoning through co-training while being substantially faster, we recommend prioritizing the partial reasoning mode for practical applications.

\begin{figure}[t!]
  \centering
  \includegraphics[width=\linewidth]{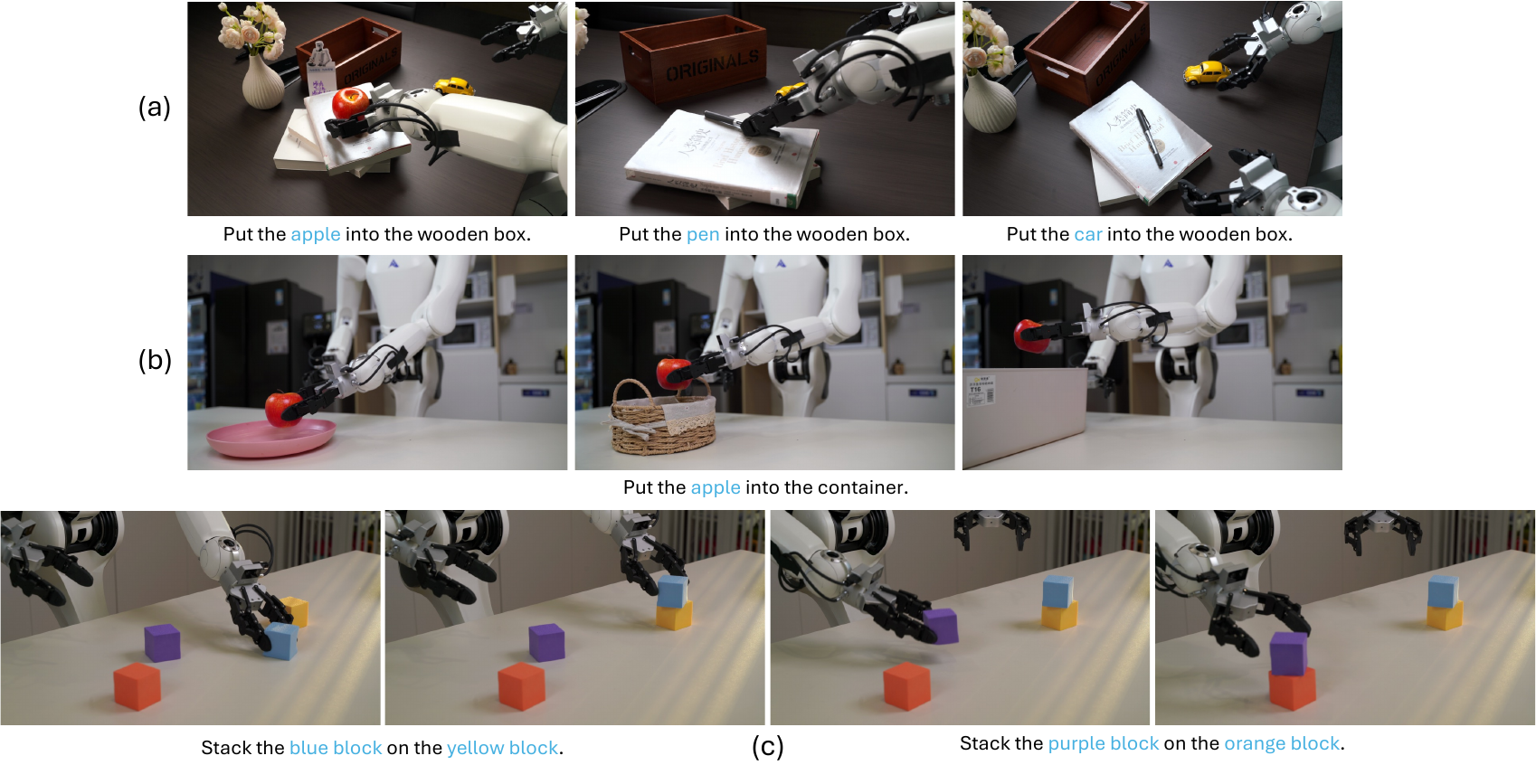}
   \caption{\textbf{\model Readily Adapts to Novel Heights.}  \model demonstrates the ability to pick and place novel objects across novel heights. Blue denotes novel objects.}
   \label{fig:general_pnp_height}
\end{figure}

\begin{figure}[h!]
  \centering
  \includegraphics[width=0.8\linewidth]{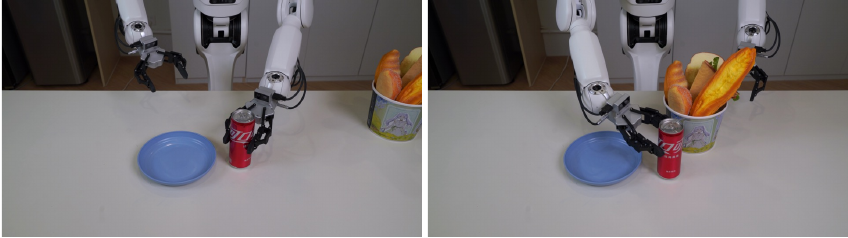}
   \caption{\textbf{\model Demonstrates Context-aware Arm Selection.} The robot chooses the right arm when the target is on its right and switches to the left arm if obstructed. }
   \label{fig:general_pnp_obstacle}
\end{figure}

In conclusion, the three-stage pre-training pipeline of \model progressively enhances its capabilities. Stage1 establishes the foundation for embodied reasoning; Stage2 strengthens precise action execution and introduces broader generalization to novel objects and environments. While these objectives may partially conflict during intermediate phases of training, Stage3 reconciles them, yielding a model that benefits from both strong reasoning and accurate action generation. We highlight the following key capabilities of \model:
    \begin{itemize}
        \item Strong instruction following performance and generalization capability, as demonstrated in Fig.~\ref{fig:general_pnp} and Fig.~\ref{fig:general_pnp_long_horizon}, including conceptually abstract prompts, novel objects, and long-horizon pick and place tasks. Furthermore, enabling full-reasoning mode yields additional performance gains, particularly on challenging cases, as shown in Fig.~\ref{fig:full_vs_partial}.
        \item Robust adaptation to varying pick-up and placement heights, as illustrated in Fig.~\ref{fig:general_pnp_height}.
        \item Context-aware arm selection based on environmental observations. As shown in Fig.~\ref{fig:general_pnp_obstacle}, the robot reaches for the coke with the right arm when it is positioned to its right side, and switches to the opposite arm when an obstacle obstructs the right arm's trajectory.
    \end{itemize}

\subsection{RL Evaluation [$\mathcal{Q}4$]}
To validate the effectiveness of the RL training phase, we sample from the \textbf{Generalizable Pick and Place} data as the validation set, resulting in approximately 950k samples. 
The validation data mirror the training set in composition, containing prompts at two granularity levels (flat-instruction and fine-grained prompt) and covering both reasoning modes, referred to as ``full reasoning`` and ``partial reasoning''. 
We use VLLM~\citep{kwon2023efficient} for greedy decoding to generate model outputs. Model performance is assessed using the reward score defined in Sec.~\ref{sec:rl} as our primary metric. While the absolute reward value may vary depending on environmental and training configurations, it provides a reliable measure for relative comparison across different models.
In addition, we introduce the Net Superiority Rate (NSR) metric, defined as the difference between the number of instances where the RL-trained model outperforms the Stage3 model ($N_{\text{RL} > \text{Stage3}}$) and the number of instances where the Stage3 model outperforms the RL-trained model ($N_{\text{RL} < \text{Stage3}}$), normalized by the total number of comparable instances ($N_{\text{total}}$):
\begin{equation}
    \text{NSR} = \frac{N_{\text{RL} > \text{Stage3}} - N_{\text{RL} < \text{Stage3}}}{N_{\text{total}}}
\end{equation}
The NSR provides an intuitive measure of relative performance between the two models. A positive NSR indicates overall superiority of the RL-trained model, while a negative value favors the Stage3 model. An NSR near zero suggests comparable performance between the models. By focusing on consistent, instance-wise performance advantages, this metric offers a clear and concise comparison.

\begin{table}[htbp]
\centering
\setlength{\tabcolsep}{10pt}
\begin{tabular}{lcccc}
\toprule
 & \multicolumn{2}{c}{full reasoning} & \multicolumn{2}{c}{partial reasoning} \\
\cmidrule(lr){2-3} \cmidrule(lr){4-5}
 & Stage3 & RL & Stage3 & RL \\
\midrule
total reward & $79.72_{\pm 25.06}$ & \makecell{$83.23_{\pm 21.31}$  \footnotesize $\textcolor{green}{\uparrow 3.51}$} & $67.42_{\pm 17.56}$ & \makecell{$71.59_{\pm 13.04}$  \footnotesize $\textcolor{green}{\uparrow 4.17}$} \\
bbox reward & $83.26_{\pm 32.17}$ & \makecell{$85.20_{\pm 29.76}$  \footnotesize $\textcolor{green}{\uparrow 1.94}$} & - & - \\
keypoint reward & $91.10_{\pm 25.62}$ & \makecell{$91.08_{\pm 26.64}$  \footnotesize $\textcolor{red}{\downarrow 0.02}$} & - & - \\
waypoint reward & $96.23_{\pm 18.33}$ & \makecell{$99.68_{\pm \ 2.60}$  \footnotesize $\textcolor{green}{\uparrow 3.45}$} & - & - \\
action reward & $64.38_{\pm 18.79}$ & \makecell{$68.82_{\pm 14.61}$  \footnotesize $\textcolor{green}{\uparrow 4.44}$} & $63.80_{\pm 19.51}$ & \makecell{$68.44_{\pm 14.49}$  \footnotesize $\textcolor{green}{\uparrow 4.64}$} \\
\bottomrule
\end{tabular}
\caption{\textbf{Comparison of Rewards: Stage3 Model vs. RL-Trained Model}. Reward values are presented in percentage. While the upper bounds may exceed $100\%$ due to the displayed standard deviations, all actual observations remain within the valid range of [0, 1].}
\label{tab:rl_reward}
\end{table}
\begin{table}[htbp]
\centering
\setlength{\tabcolsep}{10pt}
\begin{tabular}{lcc}
\toprule
& full reasoning & partial reasoning \\
\midrule
bbox NSR & $+5.05\%$ &  - \\
keypoint NSR & $+2.69\%$ & - \\
waypoint NSR & $+22.43\%$ &  -  \\
action NSR & $+23.33\%$ & $+21.03\%$ \\
\bottomrule
\end{tabular}
\caption{\textbf{Comparison of Net Superiority Rates (NSR): Stage3 Model vs. RL-Trained Model.}}
\label{tab:rl_nsr}
\end{table}

\begin{figure}[htbp]
  \centering
  \includegraphics[width=0.925\linewidth]{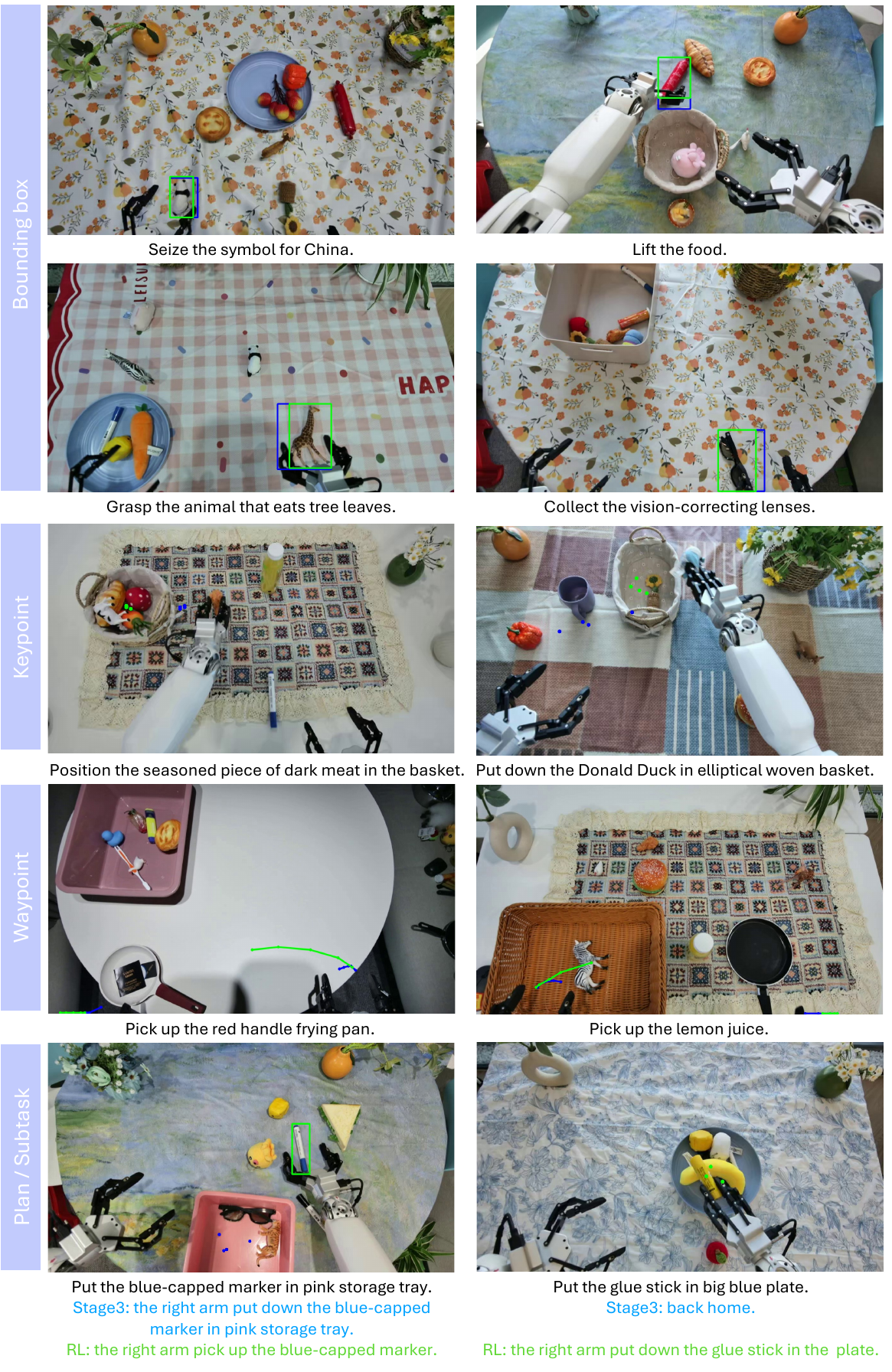}
  \caption{\textbf{Selected Examples Comparing the Stage3 Model and the RL-trained Model.} The visualization shows the model prediction results, with Stage3 in \colorbox{white}{\textcolor{shadeblue}{blue}} and RL in \colorbox{white}{\textcolor{shadegreen}{green}}.}
  \label{fig:stage3_vs_rl}
\end{figure}

The comparative results between the Stage3 and RL-trained models are summarized in Table~\ref{tab:rl_reward} and Table~\ref{tab:rl_nsr}.
As shown in Table~\ref{tab:rl_reward}, the RL-trained model consistently achieves higher reward values compared to the Stage3 model across nearly all evaluation metrics and reasoning modes. In particular, under the full reasoning mode, the RL model shows notable improvements in locating key areas such as bounding box, waypoint, and action rewards. Table~\ref{tab:rl_nsr} further confirms these findings: NSR values are consistently positive, indicating overall superiority of the RL-trained model, with the largest gains observed in waypoint and action rewards. These results demonstrate that the RL training phase effectively enhances model performance, particularly in trajectory planning and action execution. As illustrated in Fig.~\ref{fig:stage3_vs_rl}, the selected samples reveal a positive trend across various components of the reasoning process.

\subsection{Post-Training [$\mathcal{Q}5$]}
To access \model's capability as a generalist robotic foundation model, we evaluate on 6 fine-tuning tasks including long-horizon and dexterous manipulation tasks. It is well recognized that benchmarking standards for robotic policy evaluation remain lacking. Assuming that effective robot data collection emphasizes diverse trajectory variations in object and target placements, the number of fine-tuning samples is expected to correlate positively with position generalization. Consequently, the absolute number of episodes is not the sole definitive metric - since, in the extreme case, a model could simply memorize a single trajectory and be evaluated under an identical setup, akin to trajectory replay. In contrast, the number of training epochs primarily influences the in-domain success rate. Therefore, in our evaluations, all models are trained using the same number of samples until convergence (with the same number of training epochs), while evaluation conditions are strictly controlled to ensure identical setups across models in each trial.

\subsubsection{Task Specification}
\label{sub:task_spec}
\textbf{Organize stationery:} In this task, the robot is required to pick up pen-like objects and place them into a pen holder. Successful execution demands precise wrist rotation control and accurate relative positioning. A total of 400 episodes are collected for fine-tuning. In each episode, three pen-like objects are placed on the table. Evaluation is conducted on a 6-point scale: 1 point for successfully grasping each object and 1 point for correctly placing it into the pen holder.

\textbf{Play basketball:} In this task, the robot is required to pick up a small basketball and place it into a basketball hoop, which demands accurate perception of height and depth. A total of 400 episodes are collected for fine-tuning. Task performance is evaluated on a 2-point scale: 1 point for successfully grasping the basketball and 1 point for placing it into the hoop.

\textbf{Serve water:} In this task, the robot is required to pick up a glass and pour water into it. Successful execution requires precise positioning of both the glass and the water container. A total of 400 episodes are collected for fine-tuning. The task is scored on a 5-point scale: 1 point for grasping the glass, 1 point for picking up the water bottle, 1 point for positioning the water bottle for pouring, 1 point for placing the glass on the table,  and 1 point for positioning the water bottle back to the designated location.

\textbf{Pack a toy:} In this task, the robot is required to pack a toy into a designated box. The task involves not only placing the toy correctly but also arranging the box lid to facilitate proper closure. A total of 400 episodes are collected for fine-tuning. Evaluation is based on a 3-point scale: 1 point for successfully grasping the toy, 1 point for placing it inside the box, and 1 point for properly closing the box.

\textbf{Prepare food:} In this task, the robot is required to heat food using a microwave, representing a long-horizon manipulation task. This task is particularly challenging, especially as door opening and knob turning actions demand high-precision control.
A total of 2855 full-horizon episodes are collected for fine-tuning. Additionally, we collect mock ``intervention'' data for door opening and knob turning, which has proven effective for high-precision tasks~\citep{luo2025precise,amin2025pi}.
The task decomposes into the following subtasks: (1) open the microwave door, (2) pick up the food, (3) place the food into the microwave, (4) close the door, (5) turn the knob to start heating, (6) reopen the door, (7) remove the heated food, (8) place the food onto a plate, and (9) close the microwave door.
Because several subtasks pose significant difficulty, we evaluate performance at a per-subtask granularity. Each subtask begins from an in-domain initial pose, and success is measured by whether the commanded subtask is executed correctly. Each subtask is repeated 10 times, and we report the average completion success rate as the final score.

\textbf{Fold towel:} In this task, the robot is required to fold a towel, which presents the challenge of manipulating a deformable object. A total of 400 episodes are collected for fine-tuning. The task consists of four sequential steps: first, securely grasping the middle of the towel; second, laying it half-folded on the table; third, folding it again by flipping one edge; and finally, neatly arranging the folded towel. Evaluation is conducted on a binary scale (0–1), where 1 indicates a correctly folded towel in the intended configuration.

\begin{figure}[htb!]
  \centering
  \includegraphics[width=\linewidth]{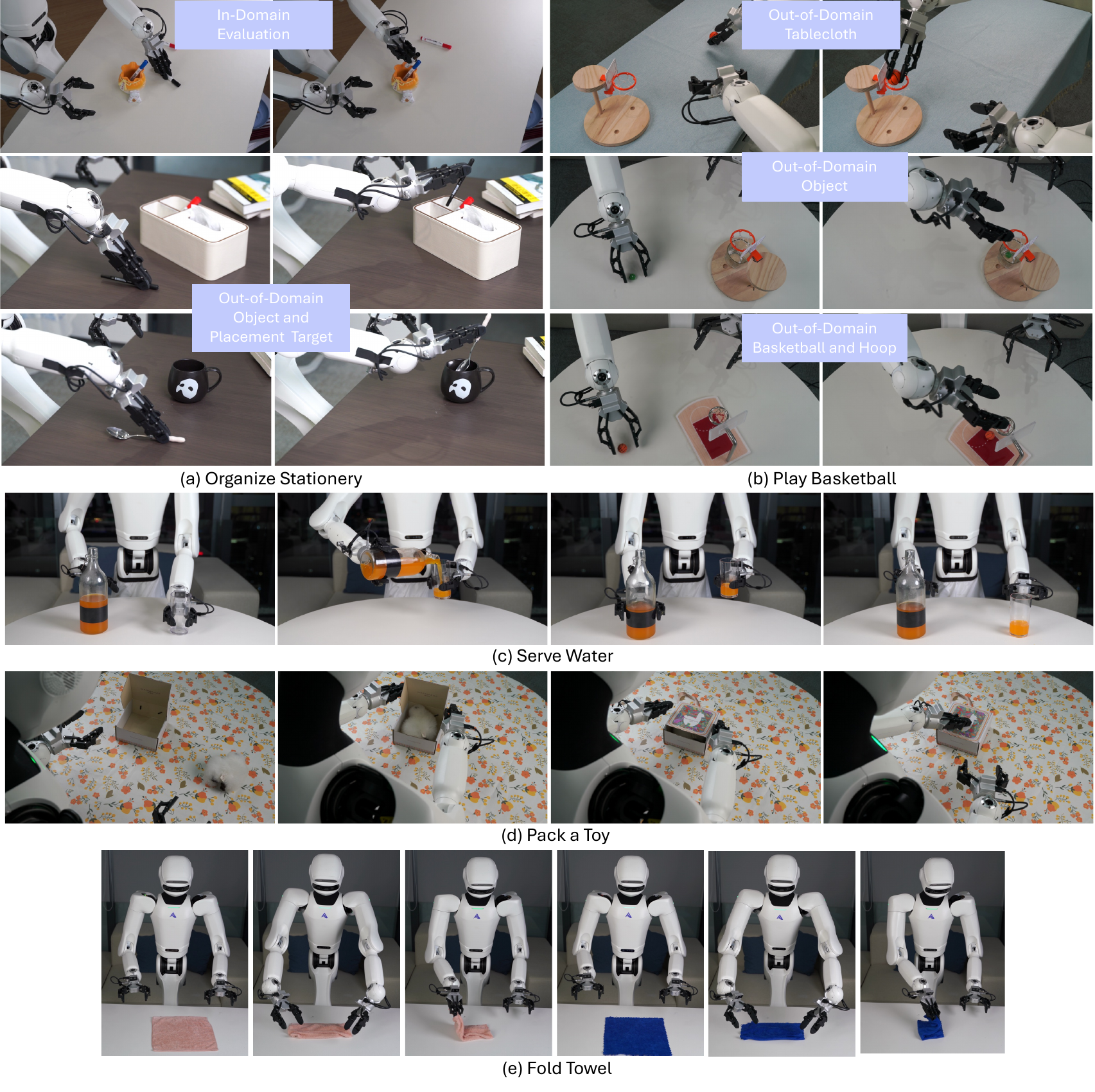}
   \caption{\textbf{Sample Model Rollouts of Fine-tuning Tasks.} \model demonstrates strong generalization to novel objects and environments.}
\label{fig:finetune_task_rollouts}
\end{figure}

\subsubsection{Evaluation Results}
We evaluate \model against two state-of-the-art baselines, $\pi_0$~\citep{black2024pi_0} and $\pi_{0.5}$~\citep{intelligence2025pi_}. The evaluation metric is a normalized score, averaged over 10 episodes per task and method. Scores are assigned according to Sec.~\ref{sub:task_spec}. For the \textit{Prepare Food} task, the overall score is computed as the average success rate across its subtasks. For all other tasks, each episode is assigned a score of 1.0 for full success, with fractional scores given for partial success to reflect the task completion progress. Evaluation conditions are strictly controlled to ensure identical setups across models. Each task layout is evaluated twice to account for inference randomness. Note that we deliberately evaluate on diverse layouts, including challenging setups, to assess the models' generalization capability. Sample rollouts are shown in Fig.~\ref{fig:finetune_task_rollouts}.

\begin{figure}[h!]
  \centering
  \includegraphics[width=0.8\linewidth]{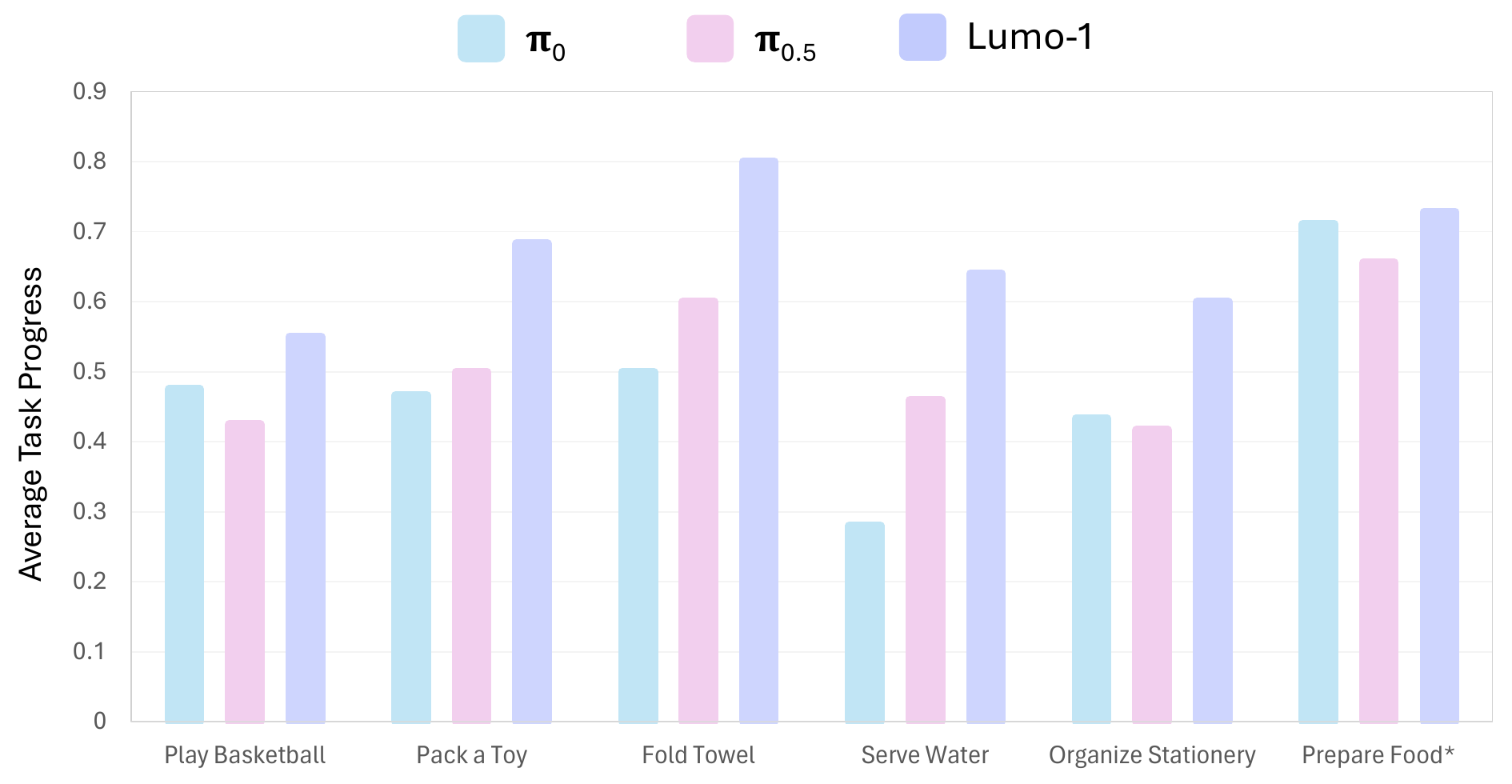}
   \caption{\textbf{Experimental Results on Fine-tuning Tasks.} \model outperforms the baseline models $\pi_0$ and $\pi_{0.5}$ across all tasks. The reported scores reflect task completion progress, except for the starred \textit{Prepare Food} task, whose score corresponds to the average success rate of its subtasks as detailed in Sec.~\ref{sub:task_spec}.}
\label{fig:finetune_task_results}
\end{figure}

\textbf{Results.} Evaluation results are shown in Fig.~\ref{fig:finetune_task_results}. \model outperforms the baseline models $\pi_0$ and $\pi_{0.5}$ across all six evaluation tasks, demonstrating strong performance on fine perception (e.g. Play Basketball), dexterous tasks (e.g. Pack a Toy, Fold Towel, Serve Water, Organize Stationery), and long-horizon task (e.g. Prepare Food), as well as robustness to positional variations and object generalization. 

\begin{figure}[h!]
  \centering
  \includegraphics[width=\linewidth]{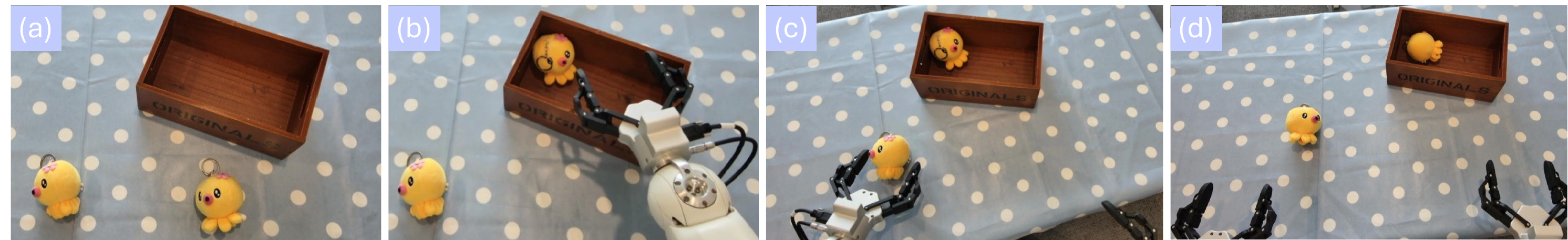}
   \caption{\textbf{Predicting Subtask Completeness Provides History Context.} The added context prevents ambiguity under similar observations, especially in long-horizon tasks with consistent execution pipelines. }
\label{fig:task_completeness_prediction}
\end{figure}

\model supports flexible fine-tuning configurations according to task characteristics and difficulty, including \textbf{action-only} and various \textbf{reasoning-augmented} settings. Optionally, the model can be further fine-tuned with a flow-matching action expert pre-trained on diverse robot trajectories.
For long-horizon tasks, previous approaches typically predict subtask prior to action generation. Here, we demonstrate the benefit of incorporating an additional subtask completeness prediction, as introduced in Sec.~\ref{sub:long_horizon_reasoning}, using a simple illustrative example as shown in Fig.~\ref{fig:task_completeness_prediction}. The task is to place \textbf{ONE} octopus into the wooden box, with the initial setup shown in Fig.~\ref{fig:task_completeness_prediction} (a). After successfully placing one octopus into the box, as shown in Fig.~\ref{fig:task_completeness_prediction} (b), the model relying solely on subtask prediction incorrectly proceeds with another ``pick up the octopus'' instruction (Fig.~\ref{fig:task_completeness_prediction} (c)). In contrast, the model with subtask completeness prediction correctly recognizes that the task has already been fulfilled and remains idle after executing the final ``back to home'' command, as shown in Fig.~\ref{fig:task_completeness_prediction} (d).

\subsection{Scaling Laws [$\mathcal{Q}6$]}

According to scaling law theories extensively studied in large language models~\citep{kaplan2020scaling, hoffmann2022training}, under a fixed computational budget, model loss follows a power-law relationship with both the number of parameters ($N$) and the dataset size ($D$). Achieving optimal performance further requires scaling these two factors in a balanced manner:
\begin{equation}
    L(N, D) = \frac{A}{N^{\alpha}} + \frac{B}{D^{\beta}} + E
\end{equation}
However, in robotic learning, acquiring large-scale, high-quality robot data is limited by physical constraints and annotation costs, while model size is typically determined by task requirements or upstream foundation models. Therefore, investigating the scaling laws of validation loss over the course of training under fixed model architectures and limited data is crucial for assessing data efficiency and guiding data collection strategies.

We adopt the \textbf{Data-Constrained Scaling Law}~\citep{muennighoff2023scaling}, which models the effective contributions of data and parameters using an exponential decay formulation, where the value of a data token diminishes by roughly $\left(1-e^{-1 / R_{D}^{*}}\right)$ per repetition.  Under the assumption of a fixed model size, the scaling law can be further simplified as:
\begin{equation}
    L(D) = \frac{B}{{D^{\prime}}^{\beta}} + E; \qquad
D^{\prime}=U_{D}+U_{D} R_{D}^{*}\left(1-e^{-R_{D} / R_{D}^{*}}\right),
\end{equation}
where $E$ is the asymptotic lower bound of the loss, $B$ controls the initial loss magnitude, and $\beta$ is the scaling exponent. $U_D$ denotes the amount of unique data, $R_D$ is the number of data repetitions, and $R_D^*$ is a learned decay constant that characterizes the diminishing marginal utility of repeated data.

\begin{figure}[t]
  \centering
  \includegraphics[width=\linewidth]{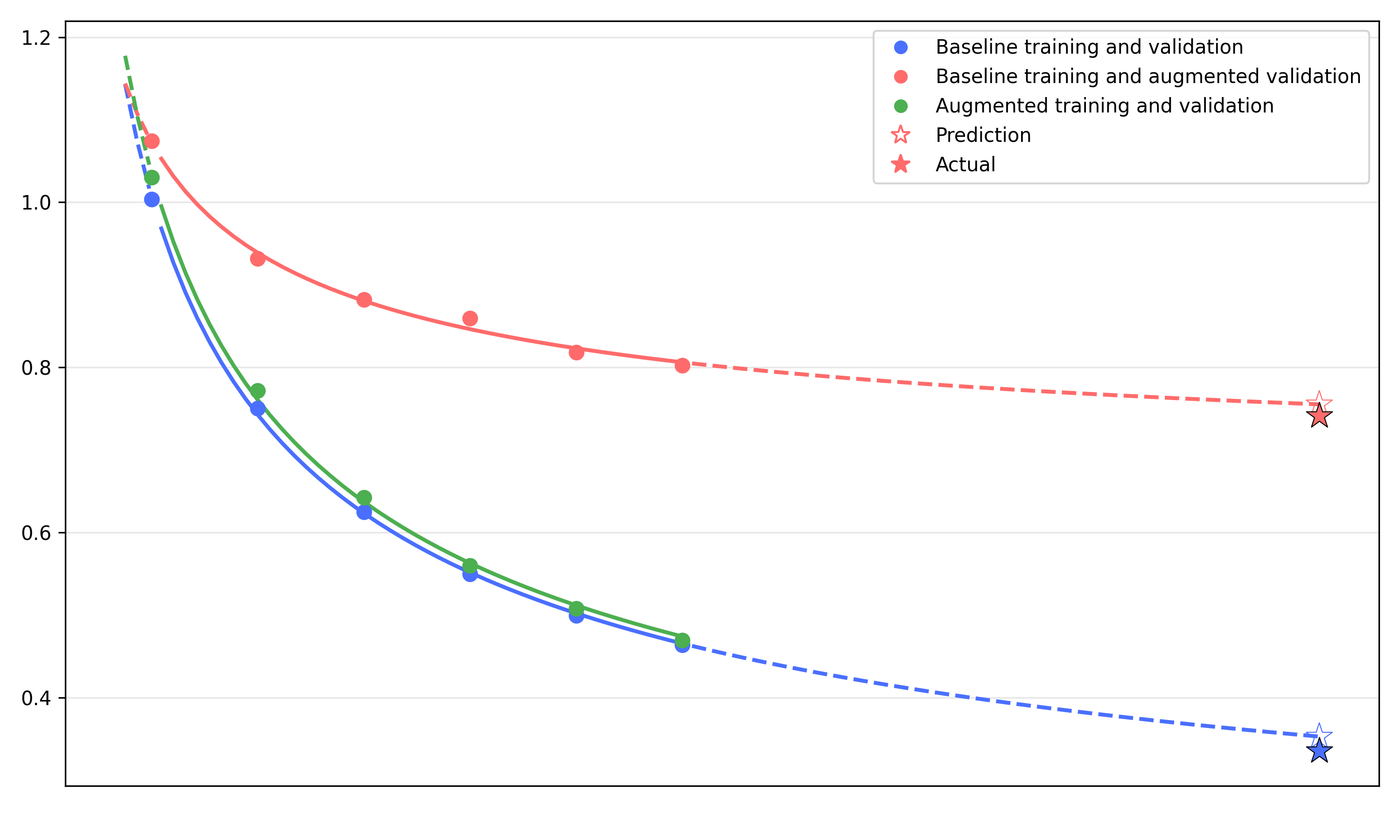}
   \caption{\textbf{Scaling Law Exploration.} Hollow stars show loss predictions from the Data-Constrained Scaling Law, which closely match the observed values (solid stars).}
   \label{fig:scaling_law}
\end{figure}

Fig.~\ref{fig:scaling_law} illustrates the validation loss curves under three different training-validation configurations, with the horizontal axis representing the number of repetitions of the original data $R_D$ (General Pick and Place as detailed in Sec.~\ref{sec:s1_data} robot data collection) and the vertical axis representing the validation loss:
\begin{itemize}
    \item \colorbox{white}{\textcolor{shadeblue}{\textbf{Blue:}}} trained on the original data and evaluated on in-domain validation set;
    \item \colorbox{white}{\textcolor{shadered}{\textbf{Red:}}} trained on the original data, evaluated on augmented validation set with image transformations (e.g., lighting and viewpoint variations) to simulate real-world conditions;
    \item \colorbox{white}{\textcolor{shadegreen}{\textbf{Green:}}} trained on augmented data and evaluated on the augmented validation set.
\end{itemize}
In the experiments, Warmup-Stable-Decay (WSD) learning rate scheduler~\citep{hu2024minicpm} was utilized to reduce training time, with learning rate decay applied at 20\% of the training steps for each checkpoint. Key observations include: (1) \textbf{Scaling Law Validity:} loss predictions from the Data-Constrained Scaling Law (hollow stars) closely match observed values (solid stars), confirming its applicability in data-constrained robotic learning. (2) \textbf{Data Diversity Necessity:} policies trained without augmentations (red curve) perform poorly under real-world variations, highlighting the importance of diverse  training data. Training with broad diversity (green curve), including prompt and image augmentations, improves resilience to validation perturbations and reduces loss on fully out-of-domain data (e.g., involving backgrounds, scenes, objects).

\subsection{Key Findings}
\note{Pre-training on embodied VLM data enhances embodied reasoning capabilities.}
\note{Co-training robot data alongside VLM data effectively preserves VLM capabilities.}
\note{Pre-training on cross-embodied data improves action learning for the target embodiment.}
\warning{Embodied reasoning enhances both action prediction and generalization.}
\warning{Reinforcement learning can be leveraged to improve reasoning–action consistency.}
\important{Subtask generation and completeness prediction enhance long-horizon task robustness.}
\important{Scaling laws should be studied and applied to robot data learning to guide the development of generalist robot policies.}

\section{Related Work}
\paragraph{Generalist Robot Manipulation Policies.}
Developing generalist manipulation policies that can interpret human instructions and interact effectively with the physical world has long been a fundamental challenge in robotics. Recent advances in vision-language-action (VLA) models have shown remarkable progress in both manipulation capability and policy generalization, typically following three key directions: (1) integrating pre-trained vision-language models into robotic policies to enable generalization to novel environments and tasks~\citep{brohan2022rt, zitkovich2023rt, driess2023palm, black2024pi_0, intelligence2025pi_}; (2) leveraging large-scale cross-embodiment datasets encompassing diverse robot platforms and task domains~\citep{o2024open, team2024octo, kim2024openvla, black2024pi_0, pertsch2025fast, intelligence2025pi_, liu2024rdt, li2024cogact}; and (3) jointly training on multi-modal data to strengthen perception–action grounding and reasoning, or leveraging external vision foundation models to provide action cues~\citep{brohan2022rt, intelligence2025pi_,li2025controlvla}.
In this work, we introduce \textbf{\model}, a vision-language-action (VLA) model trained through a three-stage pipeline that integrates large-scale robotic trajectories with curated vision-language data. The framework is designed for \textbf{purposeful robotic control} via structured reasoning. Comprehensive experiments show that \model (1) faithfully follows natural-language instructions and generalizes to unseen objects and environments; (2) efficiently adapts to novel long-horizon, dexterous manipulation tasks.

\paragraph{Reasoning Before Generation.}
Chain-of-thought (CoT) prompting~\citep{wei2022chain} has greatly enhanced multi-step reasoning in large language models across fields such as mathematics, programming, and question answering. This concept has been further extended to handle visual inputs through multimodal CoT~\citep{bigverdi2025perception, zhang2023multimodal}, where information from images is processed iteratively alongside textual reasoning. Inspired by these advances, recent robotics research has sought to integrate similar reasoning capabilities into vision-language-action (VLA) models to improve embodied task performance. ECoT~\citep{zawalski2024robotic} applies supervised fine-tuning to train VLAs to reason before executing actions. CoT-VLA~\citep{zhao2025cot} replaces textual CoT with visual subgoal frames generated before action prediction. MolmoAct~\citep{lee2025molmoact} and Emma-X \citep{sun2024emma} autoregressively generate reasoning data such as subtasks, depth perception and future gripper states. ThinkAct~\citep{huang2025thinkact} combines action-aligned reinforcement learning with visual latent planning to link embodied reasoning to real-world action prediction. In this work, we construct a structured reasoning trace that jointly captures subtask reasoning, planning, and trajectory prediction.

\paragraph{Whole-Body Mobile Manipulation}
While mobile manipulation has been extensively studied~\citep{gamma, acdit, hermes}, research on whole-body manipulation remains limited. This is primarily due to the lack of flexible robot hardware, the scarcity of data, and the inherent challenges of processing complex observations, achieving robust generalization, and generating coherent actions. Existing approaches~\citep{jiang2025behavior, gao2025towards, su2025dspv2} have explored various vision encoding strategies, but policy-level generalization has remained underexplored. In this work, our generalist policy produces coherent actions on a high-DoF robot with flexible head and torso motion, taking a further step towards humanoid robots performing human-like tasks in unmodified real-world environments.

\section{Conclusion}
\label{sec:conclusion}
We introduce \model, a generalist vision-language-action (VLA) model that unifies embodied reasoning with action generation. Our approach leverages the general multi-modal reasoning capabilities of pre-trained vision-language models (VLMs) and progressively extends them to embodied reasoning, action prediction, and ultimately to structured reasoning and reasoning–action alignment through a three-stage training pipeline followed by reinforcement learning fine-tuning. Comprehensive evaluations demonstrate that \model exhibits strong embodied reasoning and robust out-of-distribution generalization. We enable efficient adaptation to novel tasks with distinct reasoning structures, and propose a fine-tuning mode involving subtask completeness prediction and generation, which proves particularly robust for long-horizon tasks. Across a diverse suite of challenging robotic tasks, \model consistently outperforms strong baselines under both flat-instruction and fine-grained control settings.

\newpage
\section{Contributions}
Author contributions in the following areas are listed in alphabetical order.
\label{sec:contribution}
\begin{itemize}
    \item \textbf{Data:} Baifu Huang, Binyan Sun, Haotian Yang, Kuncheng Luo, Shangjin Xie, Weiqi Jin, Yanzhao Yu
    \item  \textbf{Model Architecture:} Binyan Sun, Jianan Wang, Peijun Tang
    \item  \textbf{Training:} Baifu Huang, Binyan Sun, Peijun Tang, Shangjin Xie,  Weiqi Jin
    \item  \textbf{Evaluation:} Baifu Huang, Binyan Sun, Haotian Yang, Kuncheng Luo, Peijun Tang, Shangjin Xie, Weiqi Jin 
    \item  \textbf{Writing:} Haotian Yang, Jianan Wang, Peijun Tang
    \item  \textbf{Project Lead:} Jianan Wang

\end{itemize}

\clearpage


\begin{thebibliography}{81}
\providecommand{\natexlab}[1]{#1}
\providecommand{\url}[1]{\texttt{#1}}
\expandafter\ifx\csname urlstyle\endcsname\relax
  \providecommand{\doi}[1]{doi: #1}\else
  \providecommand{\doi}{doi: \begingroup \urlstyle{rm}\Url}\fi

\bibitem[Amin et~al.(2025)Amin, Aniceto, Balakrishna, Black, Conley, Connors, Darpinian, Dhabalia, DiCarlo, Driess, et~al.]{amin2025pi}
A.~Amin, R.~Aniceto, A.~Balakrishna, K.~Black, K.~Conley, G.~Connors, J.~Darpinian, K.~Dhabalia, J.~DiCarlo, D.~Driess, et~al.
\newblock $\pi\_0.6$: a vla that learns from experience.
\newblock \emph{arXiv preprint arXiv:2511.14759}, 2025.

\bibitem[Bai et~al.(2025{\natexlab{a}})Bai, Cai, Chen, Chen, Chen, Cheng, Deng, Ding, Gao, Ge, Ge, Guo, et~al.]{qwen3vl}
S.~Bai, Y.~Cai, R.~Chen, K.~Chen, X.~Chen, Z.~Cheng, L.~Deng, W.~Ding, C.~Gao, C.~Ge, W.~Ge, Z.~Guo, et~al.
\newblock Qwen3-vl technical report.
\newblock \emph{arXiv preprint arXiv:2511.21631}, 2025{\natexlab{a}}.

\bibitem[Bai et~al.(2025{\natexlab{b}})Bai, Chen, Liu, Wang, Ge, Song, Dang, Wang, Wang, Tang, et~al.]{bai2025qwen2}
S.~Bai, K.~Chen, X.~Liu, J.~Wang, W.~Ge, S.~Song, K.~Dang, P.~Wang, S.~Wang, J.~Tang, et~al.
\newblock Qwen2. 5-vl technical report.
\newblock \emph{arXiv preprint arXiv:2502.13923}, 2025{\natexlab{b}}.

\bibitem[Bigverdi et~al.(2025)Bigverdi, Luo, Hsieh, Shen, Chen, Shapiro, and Krishna]{bigverdi2025perception}
M.~Bigverdi, Z.~Luo, C.-Y. Hsieh, E.~Shen, D.~Chen, L.~G. Shapiro, and R.~Krishna.
\newblock Perception tokens enhance visual reasoning in multimodal language models.
\newblock In \emph{Proceedings of the Computer Vision and Pattern Recognition Conference}, pages 3836--3845, 2025.

\bibitem[Bjorck et~al.(2025)Bjorck, Casta{\~n}eda, Cherniadev, Da, Ding, Fan, Fang, Fox, Hu, Huang, et~al.]{bjorck2025gr00t}
J.~Bjorck, F.~Casta{\~n}eda, N.~Cherniadev, X.~Da, R.~Ding, L.~Fan, Y.~Fang, D.~Fox, F.~Hu, S.~Huang, et~al.
\newblock Gr00t n1: An open foundation model for generalist humanoid robots.
\newblock \emph{arXiv preprint arXiv:2503.14734}, 2025.

\bibitem[Black et~al.(2024)Black, Brown, Driess, Esmail, Equi, Finn, Fusai, Groom, Hausman, Ichter, et~al.]{black2024pi_0}
K.~Black, N.~Brown, D.~Driess, A.~Esmail, M.~Equi, C.~Finn, N.~Fusai, L.~Groom, K.~Hausman, B.~Ichter, et~al.
\newblock $\pi$0: A vision-language-action flow model for general robot control.
\newblock \emph{arXiv preprint arXiv:2410.24164}, 2024.

\bibitem[Brohan et~al.(2022)Brohan, Brown, Carbajal, Chebotar, Dabis, Finn, Gopalakrishnan, Hausman, Herzog, Hsu, et~al.]{brohan2022rt}
A.~Brohan, N.~Brown, J.~Carbajal, Y.~Chebotar, J.~Dabis, C.~Finn, K.~Gopalakrishnan, K.~Hausman, A.~Herzog, J.~Hsu, et~al.
\newblock Rt-1: Robotics transformer for real-world control at scale.
\newblock \emph{arXiv preprint arXiv:2212.06817}, 2022.

\bibitem[Bu et~al.(2025)Bu, Cai, Chen, Cui, Ding, Feng, Gao, He, Hu, Huang, et~al.]{bu2025agibot}
Q.~Bu, J.~Cai, L.~Chen, X.~Cui, Y.~Ding, S.~Feng, S.~Gao, X.~He, X.~Hu, X.~Huang, et~al.
\newblock Agibot world colosseo: A large-scale manipulation platform for scalable and intelligent embodied systems.
\newblock \emph{arXiv preprint arXiv:2503.06669}, 2025.

\bibitem[Cheang et~al.(2025)Cheang, Chen, Cui, Hu, Huang, Kong, Li, Li, Liu, Ma, et~al.]{cheang2025gr3}
C.~Cheang, S.~Chen, Z.~Cui, Y.~Hu, L.~Huang, T.~Kong, H.~Li, Y.~Li, Y.~Liu, X.~Ma, et~al.
\newblock Gr-3 technical report.
\newblock \emph{arXiv:2507.15493}, 2025.

\bibitem[Chen et~al.(2025{\natexlab{a}})Chen, Xie, Ma, Sanketi, and Goldberg]{chen2025robo2vlm}
K.~Chen, S.~Xie, Z.~Ma, P.~R. Sanketi, and K.~Goldberg.
\newblock Robo2vlm: Visual question answering from large-scale in-the-wild robot manipulation datasets.
\newblock \emph{arXiv preprint arXiv:2505.15517}, 2025{\natexlab{a}}.

\bibitem[Chen et~al.(2025{\natexlab{b}})Chen, Liu, Qian, Jiang, Li, Zhang, Liu, Gu, Hou, Wang, Wang, and Zhang]{acdit}
S.~Chen, J.~Liu, S.~Qian, H.~Jiang, L.~Li, R.~Zhang, Z.~Liu, C.~Gu, C.~Hou, P.~Wang, Z.~Wang, and S.~Zhang.
\newblock Ac-dit: Adaptive coordination diffusion transformer for mobile manipulation, 2025{\natexlab{b}}.
\newblock URL \url{https://arxiv.org/abs/2507.01961}.

\bibitem[Chen et~al.(2023)Chen, Ge, Ge, Ding, Li, Wang, Xu, Shan, and Liu]{chen2023egoplan}
Y.~Chen, Y.~Ge, Y.~Ge, M.~Ding, B.~Li, R.~Wang, R.~Xu, Y.~Shan, and X.~Liu.
\newblock Egoplan-bench: Benchmarking multimodal large language models for human-level planning.
\newblock \emph{arXiv preprint arXiv:2312.06722}, 2023.

\bibitem[Chi et~al.(2023)Chi, Xu, Feng, Cousineau, Du, Burchfiel, Tedrake, and Song]{chi2023diffusion}
C.~Chi, Z.~Xu, S.~Feng, E.~Cousineau, Y.~Du, B.~Burchfiel, R.~Tedrake, and S.~Song.
\newblock Diffusion policy: Visuomotor policy learning via action diffusion.
\newblock \emph{The International Journal of Robotics Research}, page 02783649241273668, 2023.

\bibitem[Dai and Wang(2025)]{dai2025co}
Y.~Dai and J.~Wang.
\newblock Co-evolving embodied intelligence with design for artificial intelligence architecture.
\newblock \emph{Nature Reviews Electrical Engineering}, 2\penalty0 (3):\penalty0 149--150, 2025.

\bibitem[Deitke et~al.(2025)Deitke, Clark, Lee, Tripathi, Yang, Park, Salehi, Muennighoff, Lo, Soldaini, et~al.]{deitke2025molmo}
M.~Deitke, C.~Clark, S.~Lee, R.~Tripathi, Y.~Yang, J.~S. Park, M.~Salehi, N.~Muennighoff, K.~Lo, L.~Soldaini, et~al.
\newblock Molmo and pixmo: Open weights and open data for state-of-the-art vision-language models.
\newblock In \emph{Proceedings of the Computer Vision and Pattern Recognition Conference}, pages 91--104, 2025.

\bibitem[Driess et~al.(2023)Driess, Xia, Sajjadi, Lynch, Chowdhery, Wahid, Tompson, Vuong, Yu, Huang, et~al.]{driess2023palm}
D.~Driess, F.~Xia, M.~S. Sajjadi, C.~Lynch, A.~Chowdhery, A.~Wahid, J.~Tompson, Q.~Vuong, T.~Yu, W.~Huang, et~al.
\newblock Palm-e: An embodied multimodal language model.
\newblock 2023.

\bibitem[Driess et~al.(2025)Driess, Springenberg, Ichter, Yu, Li-Bell, Pertsch, Ren, Walke, Vuong, Shi, et~al.]{driess2025knowledge}
D.~Driess, J.~T. Springenberg, B.~Ichter, L.~Yu, A.~Li-Bell, K.~Pertsch, A.~Z. Ren, H.~Walke, Q.~Vuong, L.~X. Shi, et~al.
\newblock Knowledge insulating vision-language-action models: Train fast, run fast, generalize better.
\newblock \emph{arXiv preprint arXiv:2505.23705}, 2025.

\bibitem[Du et~al.(2024)Du, Wu, Li, Huang, and Wei]{duetal2024embspatial}
M.~Du, B.~Wu, Z.~Li, X.~Huang, and Z.~Wei.
\newblock {E}mb{S}patial-bench: Benchmarking spatial understanding for embodied tasks with large vision-language models.
\newblock In \emph{Proceedings of the Annual Meeting of the Association for Computational Linguistics}, volume~2, pages 346--355, 2024.

\bibitem[Fang et~al.(2025)Fang, Zhang, Dong, Li, Wang, Zhang, Tian, Hu, and Li]{fang2025robix}
H.~Fang, M.~Zhang, H.~Dong, W.~Li, Z.~Wang, Q.~Zhang, X.~Tian, Y.~Hu, and H.~Li.
\newblock Robix: A unified model for robot interaction, reasoning and planning.
\newblock \emph{arXiv preprint arXiv:2509.01106}, 2025.

\bibitem[Fu et~al.(2024)Fu, Hu, Li, Feng, Wang, Lin, Roth, Smith, Ma, and Krishna]{Fu2024BLINKML}
X.~Fu, Y.~Hu, B.~Li, Y.~Feng, H.~Wang, X.~Lin, D.~Roth, N.~A. Smith, W.-C. Ma, and R.~Krishna.
\newblock Blink: Multimodal large language models can see but not perceive.
\newblock In \emph{Proceedings of the European Conference on Computer Vision}, pages 148--166, 2024.

\bibitem[Gao et~al.(2025)Gao, Wang, Zuo, Jiang, Zhang, Zeng, Zhu, Ma, Chen, Sheng, et~al.]{gao2025towards}
G.~Gao, J.~Wang, J.~Zuo, J.~Jiang, J.~Zhang, X.~Zeng, Y.~Zhu, L.~Ma, K.~Chen, M.~Sheng, et~al.
\newblock Towards human-level intelligence via human-like whole-body manipulation.
\newblock \emph{arXiv preprint arXiv:2507.17141}, 2025.

\bibitem[Geist et~al.(2024)Geist, Frey, Zhobro, Levina, and Martius]{geist2024learning}
A.~R. Geist, J.~Frey, M.~Zhobro, A.~Levina, and G.~Martius.
\newblock Learning with 3d rotations, a hitchhiker's guide to so (3).
\newblock \emph{arXiv preprint arXiv:2404.11735}, 2024.

\bibitem[Gupta et~al.(2021)Gupta, Savarese, Ganguli, and Fei-Fei]{gupta2021embodied}
A.~Gupta, S.~Savarese, S.~Ganguli, and L.~Fei-Fei.
\newblock Embodied intelligence via learning and evolution.
\newblock \emph{Nature communications}, 12\penalty0 (1):\penalty0 5721, 2021.

\bibitem[Hoffmann et~al.(2022)Hoffmann, Borgeaud, Mensch, Buchatskaya, Cai, Rutherford, Casas, Hendricks, Welbl, Clark, et~al.]{hoffmann2022training}
J.~Hoffmann, S.~Borgeaud, A.~Mensch, E.~Buchatskaya, T.~Cai, E.~Rutherford, D.~d.~L. Casas, L.~A. Hendricks, J.~Welbl, A.~Clark, et~al.
\newblock Training compute-optimal large language models.
\newblock \emph{arXiv preprint arXiv:2203.15556}, 2022.

\bibitem[Hu et~al.(2024)Hu, Tu, Han, He, Cui, Long, Zheng, Fang, Huang, Zhao, et~al.]{hu2024minicpm}
S.~Hu, Y.~Tu, X.~Han, C.~He, G.~Cui, X.~Long, Z.~Zheng, Y.~Fang, Y.~Huang, W.~Zhao, et~al.
\newblock Minicpm: Unveiling the potential of small language models with scalable training strategies.
\newblock \emph{arXiv preprint arXiv:2404.06395}, 2024.

\bibitem[Huang et~al.(2025)Huang, Wu, Chen, Wang, and Yang]{huang2025thinkact}
C.-P. Huang, Y.-H. Wu, M.-H. Chen, Y.-C.~F. Wang, and F.-E. Yang.
\newblock Thinkact: Vision-language-action reasoning via reinforced visual latent planning.
\newblock \emph{arXiv preprint arXiv:2507.16815}, 2025.

\bibitem[Intelligence et~al.(2025)Intelligence, Black, Brown, Darpinian, Dhabalia, Driess, Esmail, Equi, Finn, Fusai, et~al.]{intelligence2025pi_}
P.~Intelligence, K.~Black, N.~Brown, J.~Darpinian, K.~Dhabalia, D.~Driess, A.~Esmail, M.~Equi, C.~Finn, N.~Fusai, et~al.
\newblock $\pi\_0.5$: a vision-language-action model with open-world generalization.
\newblock \emph{arXiv preprint arXiv:2504.16054}, 2025.

\bibitem[Ji et~al.(2025)Ji, Tan, Shi, Hao, Zhang, Zhang, Wang, Zhao, Mu, An, et~al.]{ji2025robobrain}
Y.~Ji, H.~Tan, J.~Shi, X.~Hao, Y.~Zhang, H.~Zhang, P.~Wang, M.~Zhao, Y.~Mu, P.~An, et~al.
\newblock Robobrain: A unified brain model for robotic manipulation from abstract to concrete.
\newblock In \emph{Proceedings of the Computer Vision and Pattern Recognition Conference}, pages 1724--1734, 2025.

\bibitem[Jiang et~al.(2025{\natexlab{a}})Jiang, Yuan, Liu, Lu, Cui, Liu, Cheng, Gao, Xu, and Zhao]{jiang2025galaxea}
T.~Jiang, T.~Yuan, Y.~Liu, C.~Lu, J.~Cui, X.~Liu, S.~Cheng, J.~Gao, H.~Xu, and H.~Zhao.
\newblock Galaxea open-world dataset and g0 dual-system vla model.
\newblock \emph{arXiv preprint arXiv:2509.00576}, 2025{\natexlab{a}}.

\bibitem[Jiang et~al.(2025{\natexlab{b}})Jiang, Zhang, Wong, Wang, Ze, Yin, Gokmen, Song, Wu, and Fei-Fei]{jiang2025behavior}
Y.~Jiang, R.~Zhang, J.~Wong, C.~Wang, Y.~Ze, H.~Yin, C.~Gokmen, S.~Song, J.~Wu, and L.~Fei-Fei.
\newblock Behavior robot suite: Streamlining real-world whole-body manipulation for everyday household activities.
\newblock \emph{arXiv preprint arXiv:2503.05652}, 2025{\natexlab{b}}.

\bibitem[Kamath et~al.(2023)Kamath, Hessel, and Chang]{kamath2023s}
A.~Kamath, J.~Hessel, and K.-W. Chang.
\newblock What's" up" with vision-language models? investigating their struggle with spatial reasoning.
\newblock \emph{arXiv preprint arXiv:2310.19785}, 2023.

\bibitem[Kaplan et~al.(2020)Kaplan, McCandlish, Henighan, Brown, Chess, Child, Gray, Radford, Wu, and Amodei]{kaplan2020scaling}
J.~Kaplan, S.~McCandlish, T.~Henighan, T.~B. Brown, B.~Chess, R.~Child, S.~Gray, A.~Radford, J.~Wu, and D.~Amodei.
\newblock Scaling laws for neural language models.
\newblock \emph{arXiv preprint arXiv:2001.08361}, 2020.

\bibitem[Kim et~al.(2024)Kim, Pertsch, Karamcheti, Xiao, Balakrishna, Nair, Rafailov, Foster, Lam, Sanketi, et~al.]{kim2024openvla}
M.~J. Kim, K.~Pertsch, S.~Karamcheti, T.~Xiao, A.~Balakrishna, S.~Nair, R.~Rafailov, E.~Foster, G.~Lam, P.~Sanketi, et~al.
\newblock Openvla: An open-source vision-language-action model.
\newblock \emph{arXiv preprint arXiv:2406.09246}, 2024.

\bibitem[Kirillov et~al.(2023)Kirillov, Mintun, Ravi, Mao, Rolland, Gustafson, Xiao, Whitehead, Berg, Lo, et~al.]{kirillov2023segment}
A.~Kirillov, E.~Mintun, N.~Ravi, H.~Mao, C.~Rolland, L.~Gustafson, T.~Xiao, S.~Whitehead, A.~C. Berg, W.-Y. Lo, et~al.
\newblock Segment anything.
\newblock In \emph{Proceedings of the IEEE/CVF international conference on computer vision}, pages 4015--4026, 2023.

\bibitem[Kwon et~al.(2023)Kwon, Li, Zhuang, Sheng, Zheng, Yu, Gonzalez, Zhang, and Stoica]{kwon2023efficient}
W.~Kwon, Z.~Li, S.~Zhuang, Y.~Sheng, L.~Zheng, C.~H. Yu, J.~Gonzalez, H.~Zhang, and I.~Stoica.
\newblock Efficient memory management for large language model serving with pagedattention.
\newblock In \emph{Proceedings of the 29th symposium on operating systems principles}, pages 611--626, 2023.

\bibitem[Lee et~al.(2025)Lee, Duan, Fang, Deng, Liu, Li, Fang, Zhang, Wang, Lee, et~al.]{lee2025molmoact}
J.~Lee, J.~Duan, H.~Fang, Y.~Deng, S.~Liu, B.~Li, B.~Fang, J.~Zhang, Y.~R. Wang, S.~Lee, et~al.
\newblock Molmoact: Action reasoning models that can reason in space.
\newblock \emph{arXiv preprint arXiv:2508.07917}, 2025.

\bibitem[Li et~al.(2025{\natexlab{a}})Li, Zuo, Yu, Zhang, Yang, Zhang, Zhu, Zhang, Chen, Cui, et~al.]{li2025simplevla}
H.~Li, Y.~Zuo, J.~Yu, Y.~Zhang, Z.~Yang, K.~Zhang, X.~Zhu, Y.~Zhang, T.~Chen, G.~Cui, et~al.
\newblock Simplevla-rl: Scaling vla training via reinforcement learning.
\newblock \emph{arXiv preprint arXiv:2509.09674}, 2025{\natexlab{a}}.

\bibitem[Li et~al.(2025{\natexlab{b}})Li, Wu, Xi, Li, Huang, Zhang, Chen, Wang, Zhu, Liu, et~al.]{li2025controlvla}
P.~Li, Y.~Wu, Z.~Xi, W.~Li, Y.~Huang, Z.~Zhang, Y.~Chen, J.~Wang, S.-C. Zhu, T.~Liu, et~al.
\newblock Controlvla: Few-shot object-centric adaptation for pre-trained vision-language-action models.
\newblock \emph{arXiv preprint arXiv:2506.16211}, 2025{\natexlab{b}}.

\bibitem[Li et~al.(2024)Li, Liang, Wang, Luo, Chen, Liao, Wei, Deng, Xu, Zhang, et~al.]{li2024cogact}
Q.~Li, Y.~Liang, Z.~Wang, L.~Luo, X.~Chen, M.~Liao, F.~Wei, Y.~Deng, S.~Xu, Y.~Zhang, et~al.
\newblock Cogact: A foundational vision-language-action model for synergizing cognition and action in robotic manipulation.
\newblock \emph{arXiv preprint arXiv:2411.19650}, 2024.

\bibitem[Lipman et~al.(2022)Lipman, Chen, Ben-Hamu, Nickel, and Le]{lipman2022flow}
Y.~Lipman, R.~T. Chen, H.~Ben-Hamu, M.~Nickel, and M.~Le.
\newblock Flow matching for generative modeling.
\newblock \emph{arXiv preprint arXiv:2210.02747}, 2022.

\bibitem[Liu et~al.(2023)Liu, Zhu, Gao, Feng, Liu, Zhu, and Stone]{liu2023libero}
B.~Liu, Y.~Zhu, C.~Gao, Y.~Feng, Q.~Liu, Y.~Zhu, and P.~Stone.
\newblock Libero: Benchmarking knowledge transfer for lifelong robot learning.
\newblock \emph{Advances in Neural Information Processing Systems}, 36:\penalty0 44776--44791, 2023.

\bibitem[Liu et~al.(2024{\natexlab{a}})Liu, Wu, Li, Tan, Chen, Wang, Xu, Su, and Zhu]{liu2024rdt}
S.~Liu, L.~Wu, B.~Li, H.~Tan, H.~Chen, Z.~Wang, K.~Xu, H.~Su, and J.~Zhu.
\newblock Rdt-1b: a diffusion foundation model for bimanual manipulation.
\newblock \emph{arXiv preprint arXiv:2410.07864}, 2024{\natexlab{a}}.

\bibitem[Liu et~al.(2024{\natexlab{b}})Liu, Zeng, Ren, Li, Zhang, Yang, Jiang, Li, Yang, Su, et~al.]{liu2024grounding}
S.~Liu, Z.~Zeng, T.~Ren, F.~Li, H.~Zhang, J.~Yang, Q.~Jiang, C.~Li, J.~Yang, H.~Su, et~al.
\newblock Grounding dino: Marrying dino with grounded pre-training for open-set object detection.
\newblock In \emph{European conference on computer vision}, pages 38--55. Springer, 2024{\natexlab{b}}.

\bibitem[Liu et~al.(2025)Liu, Zhu, Guo, Wei, and Liu]{liu2025llava}
W.~Liu, F.~Zhu, H.~Guo, L.~Wei, and C.-L. Liu.
\newblock Llava-c: Continual improved visual instruction tuning.
\newblock \emph{arXiv preprint arXiv:2506.08666}, 2025.

\bibitem[Luo et~al.(2025)Luo, Xu, Wu, and Levine]{luo2025precise}
J.~Luo, C.~Xu, J.~Wu, and S.~Levine.
\newblock Precise and dexterous robotic manipulation via human-in-the-loop reinforcement learning.
\newblock \emph{Science Robotics}, 10\penalty0 (105):\penalty0 eads5033, 2025.

\bibitem[Muennighoff et~al.(2023)Muennighoff, Rush, Barak, Le~Scao, Tazi, Piktus, Pyysalo, Wolf, and Raffel]{muennighoff2023scaling}
N.~Muennighoff, A.~Rush, B.~Barak, T.~Le~Scao, N.~Tazi, A.~Piktus, S.~Pyysalo, T.~Wolf, and C.~A. Raffel.
\newblock Scaling data-constrained language models.
\newblock \emph{Advances in Neural Information Processing Systems}, 36:\penalty0 50358--50376, 2023.

\bibitem[O’Neill et~al.(2024)O’Neill, Rehman, Maddukuri, Gupta, Padalkar, Lee, Pooley, Gupta, Mandlekar, Jain, et~al.]{o2024open}
A.~O’Neill, A.~Rehman, A.~Maddukuri, A.~Gupta, A.~Padalkar, A.~Lee, A.~Pooley, A.~Gupta, A.~Mandlekar, A.~Jain, et~al.
\newblock Open x-embodiment: Robotic learning datasets and rt-x models: Open x-embodiment collaboration 0.
\newblock In \emph{2024 IEEE International Conference on Robotics and Automation (ICRA)}, pages 6892--6903. IEEE, 2024.

\bibitem[Pertsch et~al.(2025)Pertsch, Stachowicz, Ichter, Driess, Nair, Vuong, Mees, Finn, and Levine]{pertsch2025fast}
K.~Pertsch, K.~Stachowicz, B.~Ichter, D.~Driess, S.~Nair, Q.~Vuong, O.~Mees, C.~Finn, and S.~Levine.
\newblock Fast: Efficient action tokenization for vision-language-action models.
\newblock \emph{arXiv preprint arXiv:2501.09747}, 2025.

\bibitem[Rajbhandari et~al.(2020)Rajbhandari, Rasley, Ruwase, and He]{rajbhandari2020zero}
S.~Rajbhandari, J.~Rasley, O.~Ruwase, and Y.~He.
\newblock Zero: Memory optimizations toward training trillion parameter models.
\newblock In \emph{SC20: International Conference for High Performance Computing, Networking, Storage and Analysis}, pages 1--16. IEEE, 2020.

\bibitem[Ramanathan et~al.(2023)Ramanathan, Kalia, Petrovic, Wen, Zheng, Guo, Wang, Marquez, Kovvuri, Kadian, et~al.]{ramanathan2023paco}
V.~Ramanathan, A.~Kalia, V.~Petrovic, Y.~Wen, B.~Zheng, B.~Guo, R.~Wang, A.~Marquez, R.~Kovvuri, A.~Kadian, et~al.
\newblock Paco: Parts and attributes of common objects.
\newblock In \emph{Proceedings of the IEEE/CVF Conference on Computer Vision and Pattern Recognition}, pages 7141--7151, 2023.

\bibitem[Ravi et~al.(2024)Ravi, Gabeur, Hu, Hu, Ryali, Ma, Khedr, R{\"a}dle, Rolland, Gustafson, et~al.]{ravi2024sam}
N.~Ravi, V.~Gabeur, Y.-T. Hu, R.~Hu, C.~Ryali, T.~Ma, H.~Khedr, R.~R{\"a}dle, C.~Rolland, L.~Gustafson, et~al.
\newblock Sam 2: Segment anything in images and videos.
\newblock \emph{arXiv preprint arXiv:2408.00714}, 2024.

\bibitem[Ray et~al.(2024)Ray, Duan, Brown, Tan, Bashkirova, Hendrix, Ehsani, Kembhavi, Plummer, Krishna, et~al.]{ray2024sat}
A.~Ray, J.~Duan, E.~Brown, R.~Tan, D.~Bashkirova, R.~Hendrix, K.~Ehsani, A.~Kembhavi, B.~A. Plummer, R.~Krishna, et~al.
\newblock Sat: Dynamic spatial aptitude training for multimodal language models.
\newblock \emph{arXiv preprint arXiv:2412.07755}, 2024.

\bibitem[Redmon et~al.(2016)Redmon, Divvala, Girshick, and Farhadi]{redmon2016you}
J.~Redmon, S.~Divvala, R.~Girshick, and A.~Farhadi.
\newblock You only look once: Unified, real-time object detection.
\newblock In \emph{Proceedings of the IEEE conference on computer vision and pattern recognition}, pages 779--788, 2016.

\bibitem[Ren et~al.(2024)Ren, Chen, Jiang, Zeng, Xiong, Liu, Ma, Shen, Gao, Jiang, et~al.]{ren2024dino}
T.~Ren, Y.~Chen, Q.~Jiang, Z.~Zeng, Y.~Xiong, W.~Liu, Z.~Ma, J.~Shen, Y.~Gao, X.~Jiang, et~al.
\newblock Dino-x: A unified vision model for open-world object detection and understanding.
\newblock \emph{arXiv preprint arXiv:2411.14347}, 2024.

\bibitem[Senin(2008)]{senin2008dynamic}
P.~Senin.
\newblock Dynamic time warping algorithm review.
\newblock \emph{Information and Computer Science Department University of Hawaii at Manoa Honolulu, USA}, 855\penalty0 (1-23):\penalty0 40, 2008.

\bibitem[Shao et~al.(2024)Shao, Wang, Zhu, Xu, Song, Bi, Zhang, Zhang, Li, Wu, et~al.]{shao2024deepseekmath}
Z.~Shao, P.~Wang, Q.~Zhu, R.~Xu, J.~Song, X.~Bi, H.~Zhang, M.~Zhang, Y.~Li, Y.~Wu, et~al.
\newblock Deepseekmath: Pushing the limits of mathematical reasoning in open language models.
\newblock \emph{arXiv preprint arXiv:2402.03300}, 2024.

\bibitem[Shi et~al.(2023)Shi, Sharma, Zhao, and Finn]{shi2023waypoint}
L.~X. Shi, A.~Sharma, T.~Z. Zhao, and C.~Finn.
\newblock Waypoint-based imitation learning for robotic manipulation.
\newblock \emph{arXiv preprint arXiv:2307.14326}, 2023.

\bibitem[Song et~al.(2025)Song, Blukis, Tremblay, Tyree, Su, and Birchfield]{song2025robospatial}
C.~H. Song, V.~Blukis, J.~Tremblay, S.~Tyree, Y.~Su, and S.~Birchfield.
\newblock {RoboSpatial}: Teaching spatial understanding to {2D} and {3D} vision-language models for robotics.
\newblock In \emph{Proceedings of the IEEE/CVF Conference on Computer Vision and Pattern Recognition}, pages 15768--15780, 2025.

\bibitem[Su et~al.(2025)Su, Zhang, Chen, Tan, Tang, Wang, and Liu]{su2025dspv2}
Y.~Su, C.~Zhang, S.~Chen, L.~Tan, Y.~Tang, J.~Wang, and X.~Liu.
\newblock Dspv2: Improved dense policy for effective and generalizable whole-body mobile manipulation.
\newblock \emph{arXiv preprint arXiv:2509.16063}, 2025.

\bibitem[Sun et~al.(2025)Sun, Lang, Wu, Ding, Feng, Liu, Ye, Liu, Liu, Wang, and Yue]{Sun2025SpaceVistaAV}
P.~Sun, S.~Lang, D.~Wu, Y.~Ding, K.~Feng, H.~Liu, Z.~Ye, R.~Liu, Y.-H. Liu, J.~Wang, and X.~Yue.
\newblock Spacevista: All-scale visual spatial reasoning from mm to km.
\newblock 2025.
\newblock URL \url{https://api.semanticscholar.org/CorpusID:282055659}.

\bibitem[Sun et~al.(2024)Sun, Hong, Pala, Toh, Tan, Ghosal, Poria, et~al.]{sun2024emma}
Q.~Sun, P.~Hong, T.~D. Pala, V.~Toh, U.~Tan, D.~Ghosal, S.~Poria, et~al.
\newblock Emma-x: An embodied multimodal action model with grounded chain of thought and look-ahead spatial reasoning.
\newblock \emph{arXiv preprint arXiv:2412.11974}, 2024.

\bibitem[Team et~al.(2025{\natexlab{a}})Team, Cao, Tan, Ji, Lin, Li, Cao, Wang, Zhou, Han, et~al.]{team2025robobrain}
B.~R. Team, M.~Cao, H.~Tan, Y.~Ji, M.~Lin, Z.~Li, Z.~Cao, P.~Wang, E.~Zhou, Y.~Han, et~al.
\newblock Robobrain 2.0 technical report.
\newblock \emph{arXiv preprint arXiv:2507.02029}, 2025{\natexlab{a}}.

\bibitem[Team et~al.(2021)Team, Abramson, Ahuja, Brussee, Carnevale, Cassin, Fischer, Georgiev, Goldin, Gupta, et~al.]{team2021creating}
D.~I.~A. Team, J.~Abramson, A.~Ahuja, A.~Brussee, F.~Carnevale, M.~Cassin, F.~Fischer, P.~Georgiev, A.~Goldin, M.~Gupta, et~al.
\newblock Creating multimodal interactive agents with imitation and self-supervised learning.
\newblock \emph{arXiv preprint arXiv:2112.03763}, 2021.

\bibitem[Team et~al.(2025{\natexlab{b}})Team, Abeyruwan, Ainslie, Alayrac, Arenas, Armstrong, Balakrishna, Baruch, Bauza, Blokzijl, et~al.]{team2025gemini}
G.~R. Team, S.~Abeyruwan, J.~Ainslie, J.-B. Alayrac, M.~G. Arenas, T.~Armstrong, A.~Balakrishna, R.~Baruch, M.~Bauza, M.~Blokzijl, et~al.
\newblock Gemini robotics: Bringing ai into the physical world.
\newblock \emph{arXiv preprint arXiv:2503.20020}, 2025{\natexlab{b}}.

\bibitem[Team et~al.(2024)Team, Ghosh, Walke, Pertsch, Black, Mees, Dasari, Hejna, Kreiman, Xu, et~al.]{team2024octo}
O.~M. Team, D.~Ghosh, H.~Walke, K.~Pertsch, K.~Black, O.~Mees, S.~Dasari, J.~Hejna, T.~Kreiman, C.~Xu, et~al.
\newblock Octo: An open-source generalist robot policy.
\newblock \emph{arXiv preprint arXiv:2405.12213}, 2024.

\bibitem[Tong et~al.(2024)Tong, Brown, Wu, Woo, IYER, Akula, Yang, Yang, Middepogu, Wang, et~al.]{tong2024cambrian}
P.~Tong, E.~Brown, P.~Wu, S.~Woo, A.~J.~V. IYER, S.~C. Akula, S.~Yang, J.~Yang, M.~Middepogu, Z.~Wang, et~al.
\newblock Cambrian-1: A fully open, vision-centric exploration of multimodal llms.
\newblock \emph{Advances in Neural Information Processing Systems}, 37:\penalty0 87310--87356, 2024.

\bibitem[Wang et~al.(2021)Wang, Liu, Su, Wang, Wang, Hsu, and Chen]{wang2021ocid}
K.-J. Wang, Y.-H. Liu, H.-T. Su, J.-W. Wang, Y.-S. Wang, W.~H. Hsu, and W.-C. Chen.
\newblock Ocid-ref: A 3d robotic dataset with embodied language for clutter scene grounding.
\newblock \emph{arXiv preprint arXiv:2103.07679}, 2021.

\bibitem[Wei et~al.(2022)Wei, Wang, Schuurmans, Bosma, Xia, Chi, Le, Zhou, et~al.]{wei2022chain}
J.~Wei, X.~Wang, D.~Schuurmans, M.~Bosma, F.~Xia, E.~Chi, Q.~V. Le, D.~Zhou, et~al.
\newblock Chain-of-thought prompting elicits reasoning in large language models.
\newblock \emph{Advances in neural information processing systems}, 35:\penalty0 24824--24837, 2022.

\bibitem[Yang et~al.(2025)Yang, Li, Yang, Zhang, Hui, Zheng, Yu, Gao, Huang, Lv, et~al.]{yang2025qwen3}
A.~Yang, A.~Li, B.~Yang, B.~Zhang, B.~Hui, B.~Zheng, B.~Yu, C.~Gao, C.~Huang, C.~Lv, et~al.
\newblock Qwen3 technical report.
\newblock \emph{arXiv preprint arXiv:2505.09388}, 2025.

\bibitem[Ye et~al.(2023)Ye, Zhang, Weng, Gu, Wang, Zhang, Wang, Abbeel, and Gao]{ye2023reinforcement}
W.~Ye, Y.~Zhang, H.~Weng, X.~Gu, S.~Wang, T.~Zhang, M.~Wang, P.~Abbeel, and Y.~Gao.
\newblock Reinforcement learning with foundation priors: Let the embodied agent efficiently learn on its own.
\newblock \emph{arXiv preprint arXiv:2310.02635}, 2023.

\bibitem[Yu et~al.(2025)Yu, Zhang, Zhu, Yuan, Zuo, Yue, Fan, Liu, Liu, Liu, Lin, Lin, Ma, Sheng, et~al.]{Yu2025DAPOAO}
Q.~Yu, Z.~Zhang, R.~Zhu, Y.~Yuan, X.~Zuo, Y.~Yue, T.~Fan, G.~Liu, L.~Liu, X.~Liu, H.~Lin, Z.~Lin, B.~Ma, G.~Sheng, et~al.
\newblock Dapo: An open-source llm reinforcement learning system at scale.
\newblock \emph{ArXiv}, abs/2503.14476, 2025.
\newblock URL \url{https://api.semanticscholar.org/CorpusID:277104124}.

\bibitem[Yuan et~al.(2024)Yuan, Duan, Blukis, Pumacay, Krishna, Murali, Mousavian, and Fox]{yuan2024robopoint}
W.~Yuan, J.~Duan, V.~Blukis, W.~Pumacay, R.~Krishna, A.~Murali, A.~Mousavian, and D.~Fox.
\newblock Robopoint: A vision-language model for spatial affordance prediction for robotics.
\newblock \emph{arXiv preprint arXiv:2406.10721}, 2024.

\bibitem[Yuan et~al.(2025)Yuan, Wei, Gu, Hua, Liang, Chen, and Xu]{hermes}
Z.~Yuan, T.~Wei, L.~Gu, P.~Hua, T.~Liang, Y.~Chen, and H.~Xu.
\newblock Hermes: Human-to-robot embodied learning from multi-source motion data for mobile dexterous manipulation, 2025.
\newblock URL \url{https://arxiv.org/abs/2508.20085}.

\bibitem[Zawalski et~al.(2024)Zawalski, Chen, Pertsch, Mees, Finn, and Levine]{zawalski2024robotic}
M.~Zawalski, W.~Chen, K.~Pertsch, O.~Mees, C.~Finn, and S.~Levine.
\newblock Robotic control via embodied chain-of-thought reasoning.
\newblock \emph{arXiv preprint arXiv:2407.08693}, 2024.

\bibitem[Ze et~al.(2024)Ze, Zhang, Zhang, Hu, Wang, and Xu]{ze20243d}
Y.~Ze, G.~Zhang, K.~Zhang, C.~Hu, M.~Wang, and H.~Xu.
\newblock 3d diffusion policy: Generalizable visuomotor policy learning via simple 3d representations.
\newblock \emph{arXiv preprint arXiv:2403.03954}, 2024.

\bibitem[Zhai et~al.(2025)Zhai, Zhang, Zhang, Huang, Zhang, Zhou, Zhang, Liu, Lin, and Pang]{zhai2025vision}
S.~Zhai, Q.~Zhang, T.~Zhang, F.~Huang, H.~Zhang, M.~Zhou, S.~Zhang, L.~Liu, S.~Lin, and J.~Pang.
\newblock A vision-language-action-critic model for robotic real-world reinforcement learning.
\newblock \emph{arXiv preprint arXiv:2509.15937}, 2025.

\bibitem[Zhang et~al.(2024)Zhang, Gireesh, Wang, Fang, Xu, Chen, Dai, and Wang]{gamma}
J.~Zhang, N.~Gireesh, J.~Wang, X.~Fang, C.~Xu, W.~Chen, L.~Dai, and H.~Wang.
\newblock Gamma: Graspability-aware mobile manipulation policy learning based on online grasping pose fusion, 2024.
\newblock URL \url{https://arxiv.org/abs/2309.15459}.

\bibitem[Zhang et~al.(2023)Zhang, Zhang, Li, Zhao, Karypis, and Smola]{zhang2023multimodal}
Z.~Zhang, A.~Zhang, M.~Li, H.~Zhao, G.~Karypis, and A.~Smola.
\newblock Multimodal chain-of-thought reasoning in language models.
\newblock \emph{arXiv preprint arXiv:2302.00923}, 2023.

\bibitem[Zhao et~al.(2025)Zhao, Lu, Kim, Fu, Zhang, Wu, Li, Ma, Han, Finn, et~al.]{zhao2025cot}
Q.~Zhao, Y.~Lu, M.~J. Kim, Z.~Fu, Z.~Zhang, Y.~Wu, Z.~Li, Q.~Ma, S.~Han, C.~Finn, et~al.
\newblock Cot-vla: Visual chain-of-thought reasoning for vision-language-action models.
\newblock In \emph{Proceedings of the Computer Vision and Pattern Recognition Conference}, pages 1702--1713, 2025.

\bibitem[Zhou et~al.(2025)Zhou, An, Chi, Han, Rong, Zhang, Wang, Wang, Huang, Sheng, et~al.]{zhou2025roborefer}
E.~Zhou, J.~An, C.~Chi, Y.~Han, S.~Rong, C.~Zhang, P.~Wang, Z.~Wang, T.~Huang, L.~Sheng, et~al.
\newblock Roborefer: Towards spatial referring with reasoning in vision-language models for robotics.
\newblock \emph{arXiv preprint arXiv:2506.04308}, 2025.

\bibitem[Zitkovich et~al.(2023)Zitkovich, Yu, Xu, Xu, Xiao, Xia, Wu, Wohlhart, Welker, Wahid, et~al.]{zitkovich2023rt}
B.~Zitkovich, T.~Yu, S.~Xu, P.~Xu, T.~Xiao, F.~Xia, J.~Wu, P.~Wohlhart, S.~Welker, A.~Wahid, et~al.
\newblock Rt-2: Vision-language-action models transfer web knowledge to robotic control.
\newblock In \emph{Conference on Robot Learning}, pages 2165--2183. PMLR, 2023.
\end{thebibliography}

\clearpage
\appendix
\section*{Appendix}
The appendix includes the following sections:
\begin{itemize} 
    \item \S\ref{supp:training} - Training Details
    \item \S\ref{supp:datasemantics} - Self-collected Robot Data Semantics
    \item \S\ref{supp:reasoningconstruction} - Reasoning Data Construction
    \item \S\ref{supp:dataexamples} - Reasoning Examples
    \item \S\ref{supp:rlevalprompt} - RL Evaluation Prompt
    \item \S\ref{supp:more_examples} - Example Model Rollouts with Reasoning
\end{itemize}

\section{Training Details}
\label{supp:training}

\model achieves proficient robotic manipulation capabilities through the following progressive three-stage training protocol: Continued VLM pre-training, Co-Training on Cross-Embodiment Robot and VLM Data, and Target-Embodiment Action Training with Reasoning Process. The detailed configurations for each training stage are summarized in Table~\ref{tab:training_configuration}. Throughout the training process, the model was full-parameter fine-tuned. The Zero Redundancy Optimizer (ZeRO)~\citep{rajbhandari2020zero} was employed to mitigate GPU memory pressure.

\begin{table}[htbp]
\centering
\setlength{\tabcolsep}{8pt}
\begin{tabular}{lccc}
\toprule
 & \textbf{Stage-1} & \textbf{Stage-2} & \textbf{Stage-3} \\
\midrule
\textbf{Dataset} & VLM & VLM + Cross Embodied & Target Embodied \\
\textbf{Samples per epoch} & 16.3M & 212.5M & 16.2M  \\
\textbf{Total Tokens seen} & 13.7B & 200B & 193B \\
\textbf{Trainable Part} & Full Model & Full Model & Full Model\\
\midrule
\textbf{Min Pixels} & 3136 & 3136 & 3136\\
\textbf{Max Pixels} & 230400 & 230400 & 230400\\
\textbf{Per-device Batch Size} & 4 (concat) & 4 (concat) & 20\\
\textbf{Peak LR} & $5\times 10^{-5}$ & $1\times 10^{-5}$ & $1\times 10^{-5}$ \\
\textbf{Training Steps} & 7000 & 100000 & 70000 \\
\textbf{Optimizer} & AdamW & AdamW & AdamW \\
\textbf{Weight Decay} & 0.1 & 0.1 & 0.1  \\
\textbf{Warmup Ratio} & 0.05 & 0.01 & 0.01 \\
\textbf{LR Schedule} & Cosine & Constant & Warmup-Stable-Decay \\
\textbf{Max Seq. Length} & 4096 & 4096 & 4096 \\
\textbf{GPU Nums} & $16\times 8$ & $16\times 8$ & $16\times 8$ \\
\bottomrule
\end{tabular}
\caption{\textbf{Detailed Training Configuration.} ``concat'' refers to pre-concatenating the data to equalize sequence lengths, preventing imbalance in GPU memory occupancy.}
\label{tab:training_configuration}
\end{table}

\clearpage

\section{Self-collected Robot Data Semantics}
\label{supp:datasemantics}
\begin{figure}[h]
    \centering
    \includegraphics[width=\textwidth]{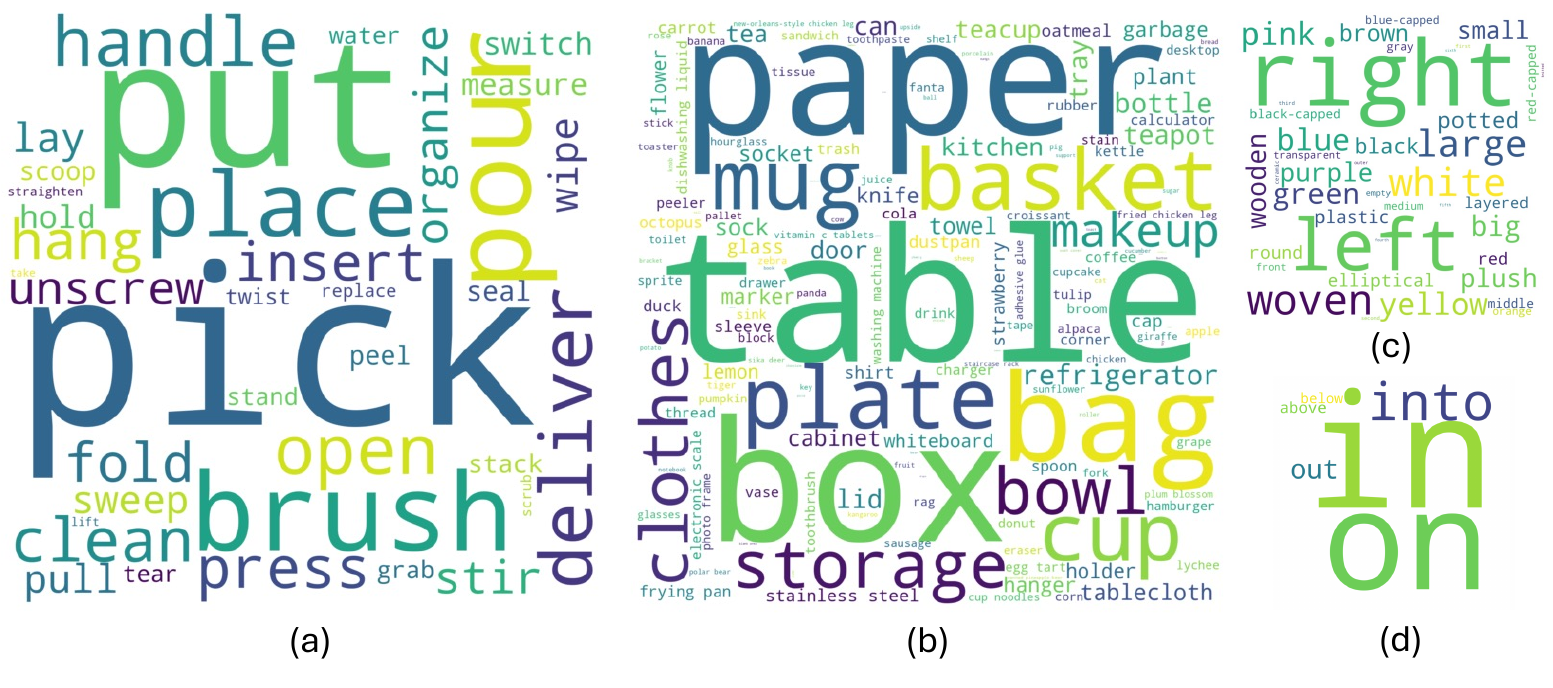}
    \caption{The word clouds provide a glimpse into the diversity of our self-collected robot data, with panels (a–d) corresponding to verbs, nouns, adjectives, and prepositions.}
    \label{fig:robot_data_semantics}
\end{figure}

We present in Fig.~\ref{fig:robot_data_semantics} word clouds for a subset of our annotated robot trajectory data, organized by part of speech (POS) as determined by NLTK, highlighting the diversity of verbs, nouns, adjectives, and prepositions.

\clearpage

\section{Reasoning Data Construction}
\label{supp:reasoningconstruction}
Our self-collected robot dataset comprises multi-view camera observations (including head and dual wrist cameras with corresponding intrinsic and extrinsic parameters), synchronized robot action data, and subtask annotations. We construct the \textbf{action reasoning data} as follows:
\begin{enumerate}
    \item \textbf{Waypoint Reasoning:} Waypoints represent key future actions projected onto the robot's head camera. Given the camera's intrinsic and extrinsic parameters, this projection can be performed directly. To ensure effective alignment between 2D visual prediction and downstream action generation, we adopt the same procedure described in Sec.~\ref{subsec:spatail_tokenizer} to select representative ``key'' action points based on the current observation.
    \label{itemone} 
    \item \textbf{Perception and Grounding:} Perception and grounding are expressed through bounding boxes and keypoint coordinates. The target manipulable object is localized using bounding box coordinates, while the corresponding placement or interaction region is indicated by keypoints - allowing flexible representation of vacant areas or multiple feasible target positions. Bounding boxes are obtained through three complementary approaches: (1) manual annotation of recurring objects followed by YOLO-based~\citep{redmon2016you} detector training; (2) open-vocabulary detection using models from the Grounding-DINO series~\citep{ren2024dino,liu2024grounding}; and (3) single-frame manual annotation combined with object tracking via SAM2~\citep{ravi2024sam}. Keypoints are derived either by sampling within container bounding boxes for ``place'' actions or by projecting the gripper position onto the head camera image plane. 
    \label{itemtwo} 
    \item \textbf{Subtask Reasoning.}  Subtask reasoning focuses on identifying the most plausible next step required to achieve the intended goal. Each subtask label is directly derived from the annotated subtask sequences in the collected dataset.
    \item \textbf{Visual Observation Description and Movement Reasoning.} 
    Visual observation description involves characterizing the scene and its constituent objects, with particular emphasis on the target object's attributes such as position, color, and material properties. Movement reasoning, in turn, infers the appropriate gripper motion based on its spatial relationship with the target object or location. We observe that general-purpose Vision-Language Models (VLMs) exhibit limited reliability in object localization and motion prediction. To mitigate this, we incorporate additional cues, including the target object's bounding box (derived from ~\ref{itemtwo}) and formatted action commands generated from the bounding box coordinates or waypoint guidance obtained from ~\ref{itemone}. The prompts are given in Fig.~\ref{fig:reasoning_generation_prompt} and Fig.~\ref{fig:reasoning_modification_prompt}, applied to Qwen2.5-VL-72B-Instruct and QwQ-32B, respectively. A representative illustration is shown in Fig.~\ref{fig:partial_reasoning_example}.

    \begin{figure}[t!]
        \centering
        \includegraphics[width=\textwidth]{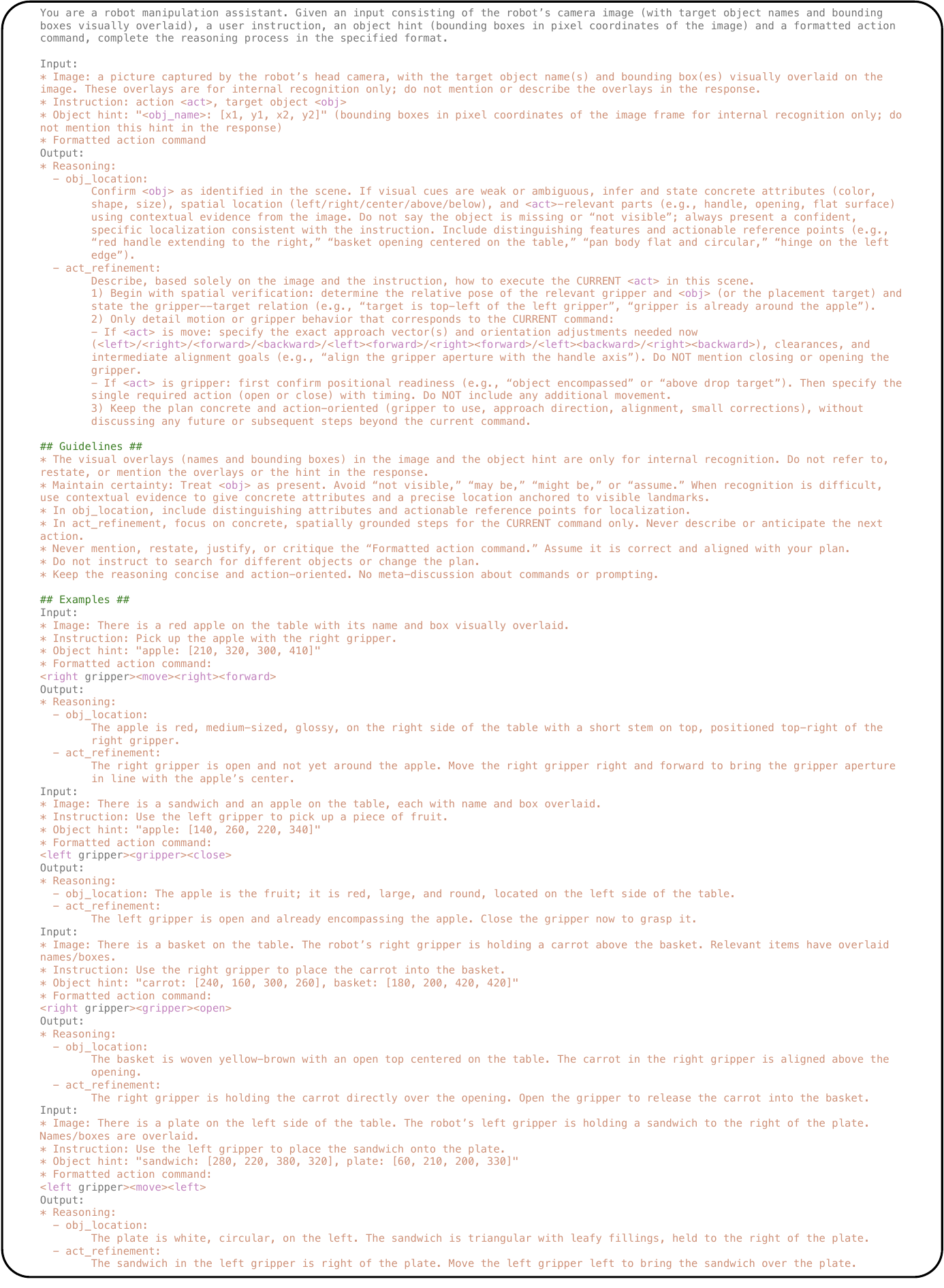}
        \caption{\textbf{Prompt for Generating Visual Observation Description and Movement Reasoning.}}
        \label{fig:reasoning_generation_prompt}
    \end{figure}

    \begin{figure}[t!]
        \centering
        \includegraphics[width=\textwidth]{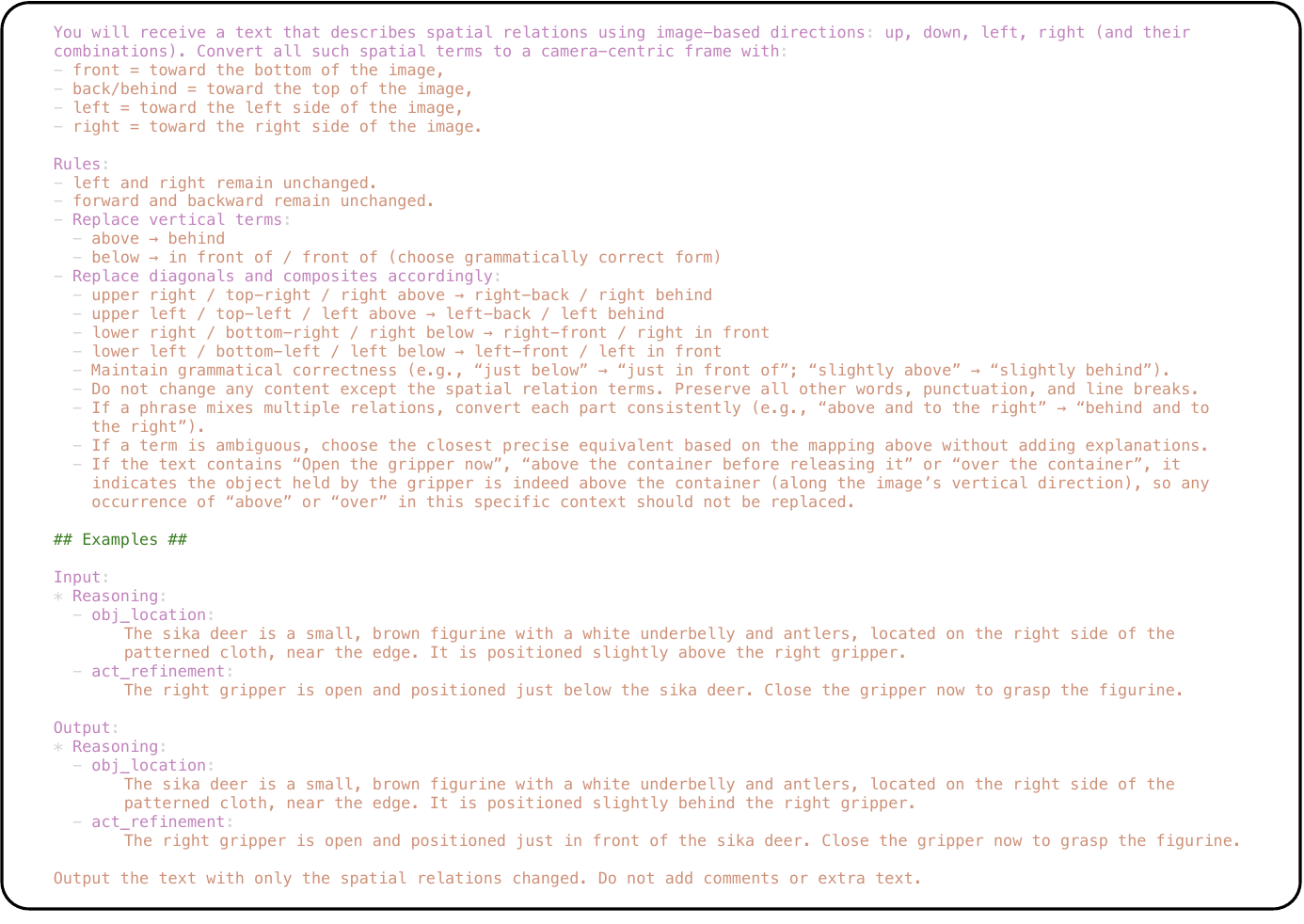}
        \caption{\textbf{Prompt for Transforming Direction Descriptions into Robot-centric Terms.}}
        \label{fig:reasoning_modification_prompt}
    \end{figure}

    \begin{figure}[h!]
        \centering
        \includegraphics[width=0.5\textwidth]{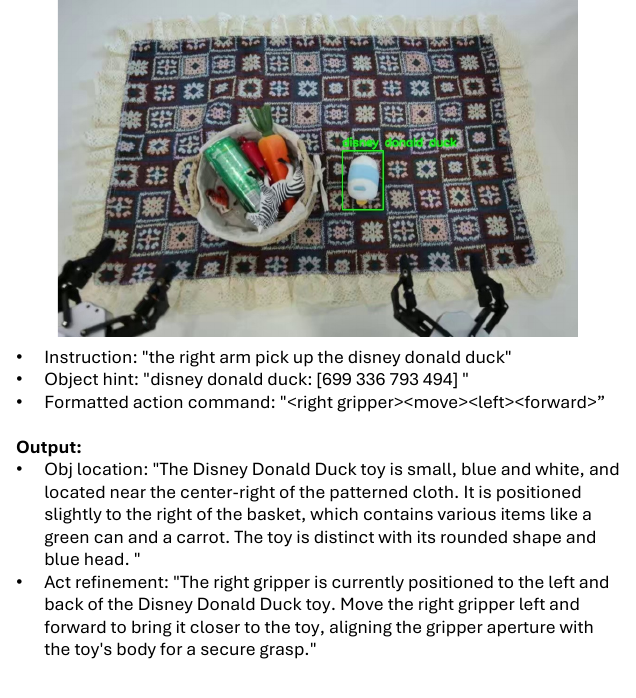}
        \caption{\textbf{Example of Visual Observation Description and Movement Reasoning.}}
        \label{fig:partial_reasoning_example}
    \end{figure}

    \item \textbf{Abstract Concept Reasoning.} We adopt two approaches for \textbf{abstract concept reasoning}. In the first approach, Qwen2.5-VL-72B-Instruct takes the image and instruction as input and outputs an enhanced instruction. In the second approach, the instruction is decomposed into verbs and nouns, which are individually enhanced with GPT-5 and subsequently recombined by Qwen3-30B-A3B-Instruct-2507. While the first approach focuses on visual attributes such as color, shape, and spatial position, the second emphasizes semantic attributes such as object functionality and intrinsic characteristics. 
\end{enumerate}

\newpage
\section{Reasoning Examples}
\label{supp:dataexamples}
\begin{figure}[h]
    \centering
    \includegraphics[width=\textwidth]{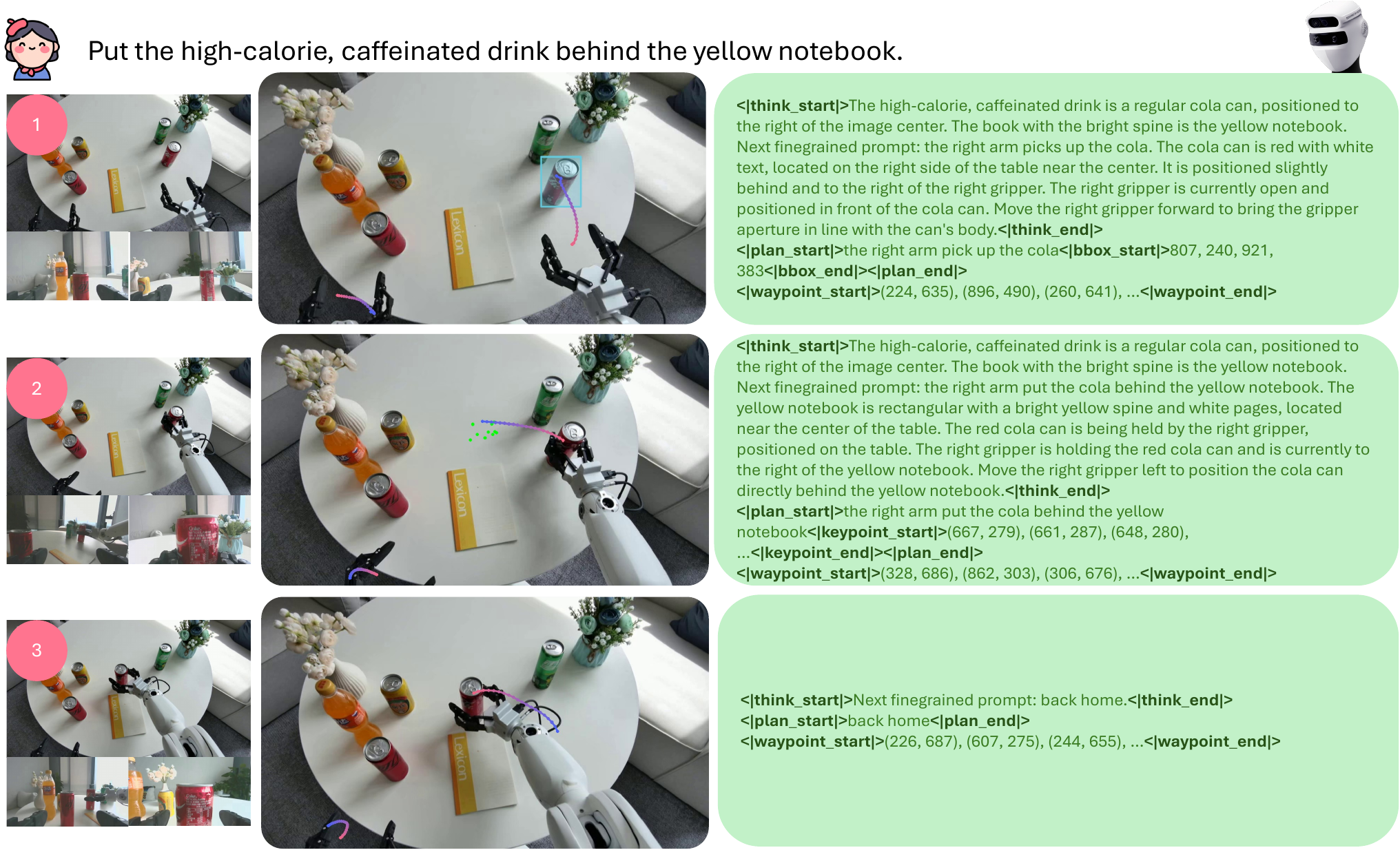}
    \caption{\textbf{Reasoning Example for a Task Requiring Conceptual Understanding}: ``Put the high-calorie, caffeinated drink behind the yellow notebook.''}
    \label{fig:reasoning_abstract}
\end{figure}

\begin{figure}[h]
    \centering
    \includegraphics[width=\textwidth]{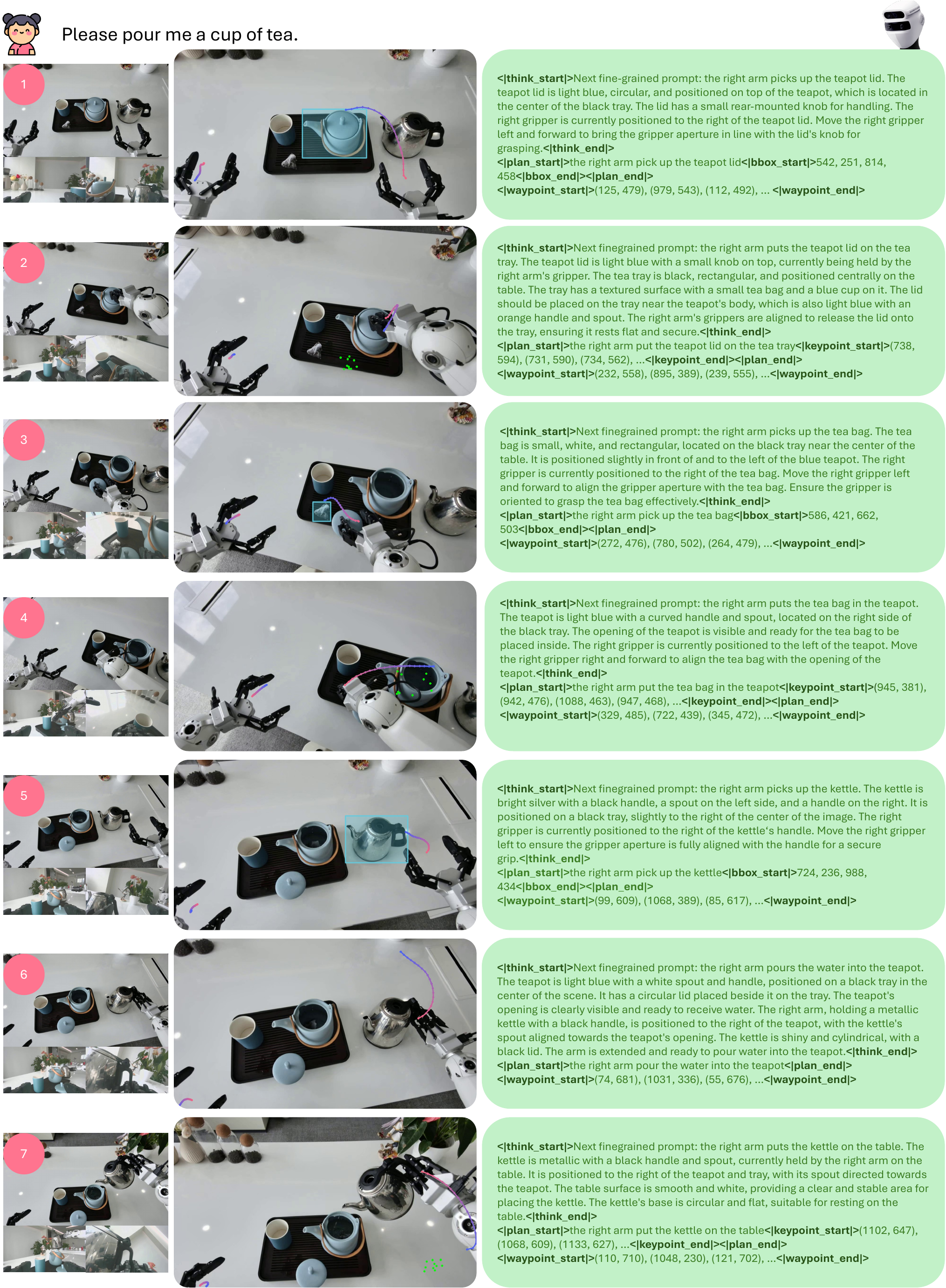}
    \caption{\textbf{Reasoning Example for a Long-horizon Task}: ``Please pour me a cup of tea.'' (Steps 1–7)}.
    \label{fig:reasoning_long_horizon_part1}
\end{figure}

\begin{figure}[h]
    \centering
    \includegraphics[width=\textwidth]{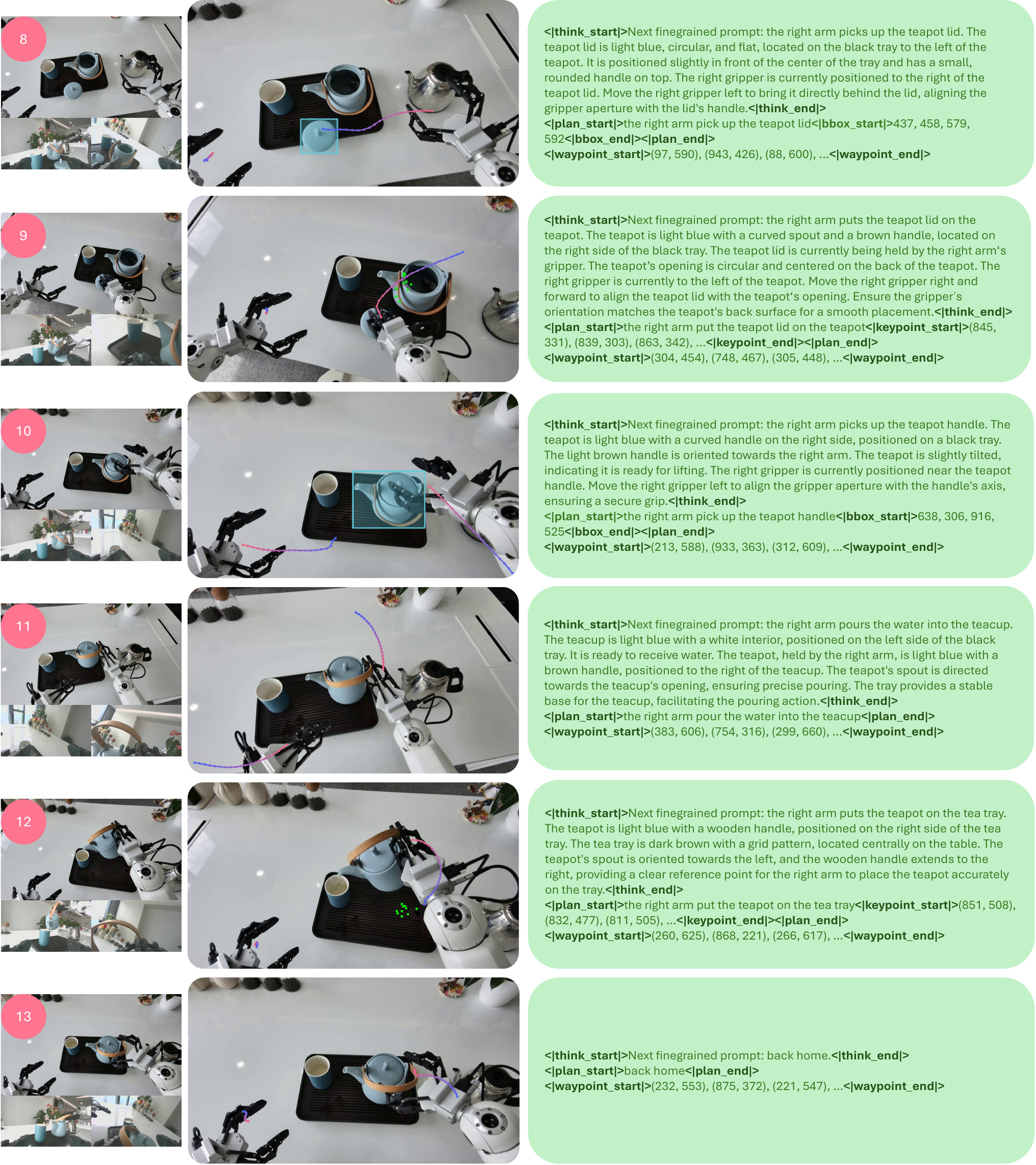}
    \caption{\textbf{Reasoning Example for a Long-horizon Task}: ``Please pour me a cup of tea.'' (Steps 8–13)}
    \label{fig:reasoning_long_horizon_part2}

\end{figure}

We present two examples from our reasoning–action dataset (showing only the reasoning component). Fig.~\ref{fig:reasoning_abstract} demonstrates abstract conceptual reasoning, while Fig.~\ref{fig:reasoning_long_horizon_part1} and Fig.~\ref{fig:reasoning_long_horizon_part2} illustrate a long-horizon task.

\clearpage

\section{RL Evaluation Prompt}
\label{supp:rlevalprompt}
\begin{figure}[h!]
    \centering
    \includegraphics[width=\textwidth]{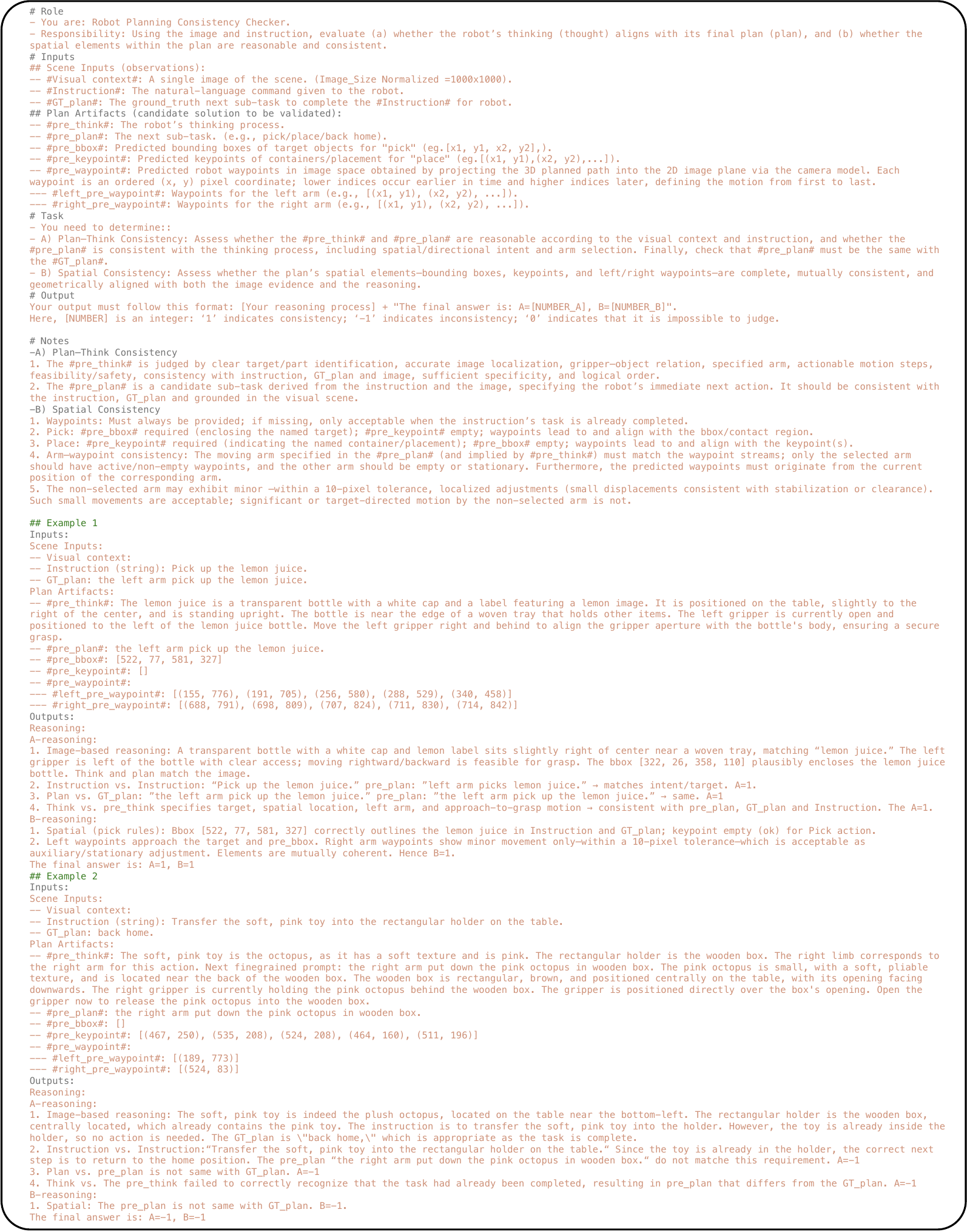}
    \caption{\textbf{Evaluation Prompt for Calculating the Consistency Reward in RL Stage.}}
    \label{fig:rl_evaluation_prompt}
\end{figure}

Fig.~\ref{fig:rl_evaluation_prompt} presents the evaluation prompt used in our RL stage. This prompt is designed to assess the correctness of textual content and the consistency between text and spatial content.

\clearpage

\section{Example Model Rollouts with Reasoning}
\label{supp:more_examples}
\begin{figure}[h!]
    \centering
    \includegraphics[width=\textwidth]{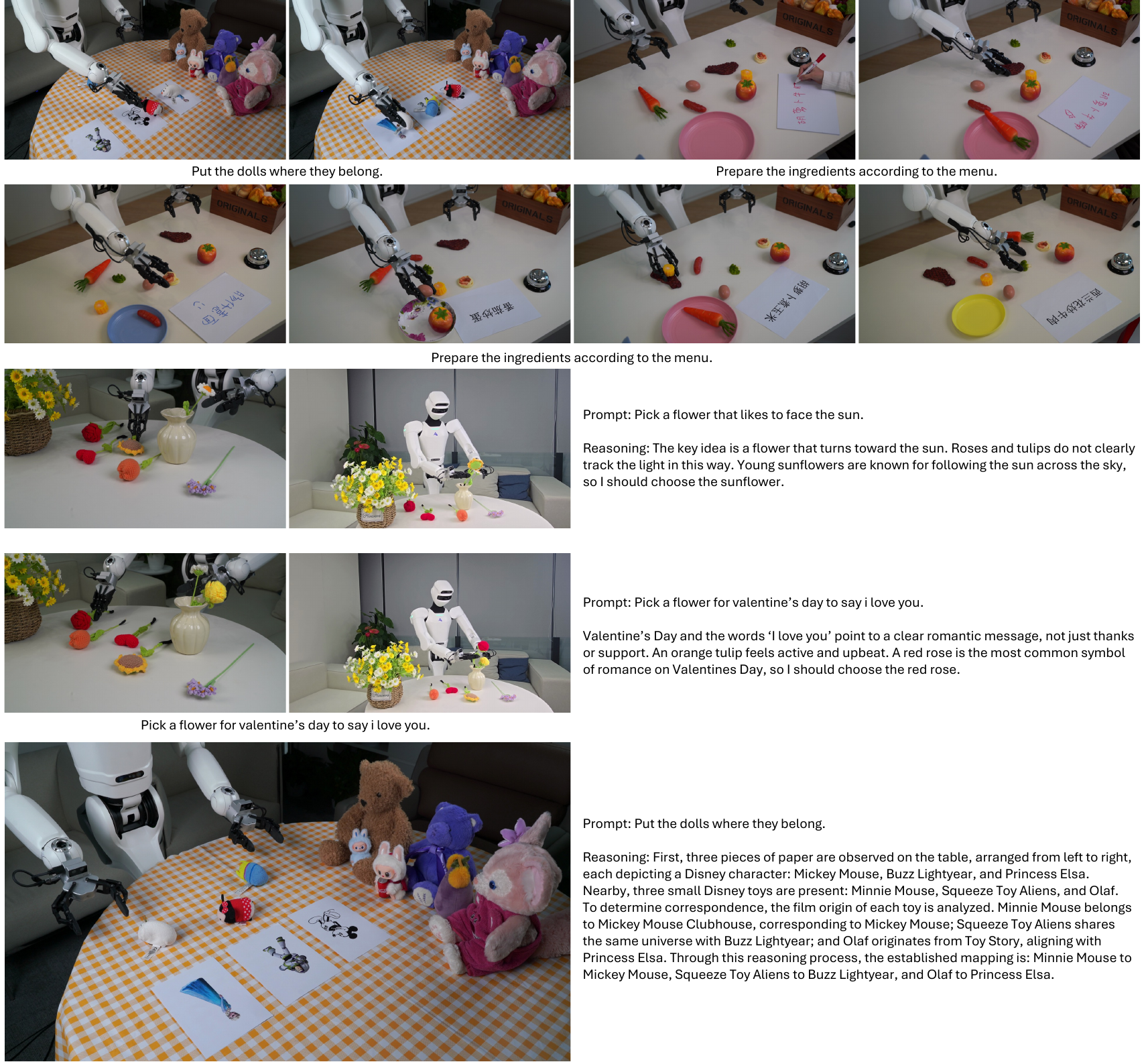}
    \caption{\textbf{Example Model Rollouts with Textural Reasoning.}}
    \label{fig:evluations_with_reasoning}
\end{figure}

Fig.~\ref{fig:evluations_with_reasoning} presents more representative model rollouts, some along with their corresponding reasoning traces.

\clearpage

\end{document}